\documentclass[11pt]{article}
\usepackage{geometry}
\input{header}

\newcommandx{\functionspace}[2][1=+]{\mathbb{F}_{#1}(#2)}

\newcommandx{\VarDeux}[3][3=]{\operatorname{Var}^{#3}_{#1}\left\{#2 \right\}}

\newcommand{\LeftEqNo}{\let\veqno\@@leqno}

\newcommand{\floor}[1]{\left\lfloor #1 \right\rfloor}
\newcommand{\ceil}[1]{\left\lceil #1 \right\rceil}

\newcommand{\N}{\ensuremath{\mathbb{N}}}

\newcommand{\R}{\ensuremath{\mathbb{R}}}

\newcommand{\PE}{\mathbb{E}}

\newcommand{\abs}[1]{\left\vert #1 \right\vert}

\newcommandx{\Vnorm}[2][1=V]{\| #2 \|_{#1}}
\newcommandx{\VnormEq}[2][1=V]{\left\| #2 \right\|_{#1}}
\newcommandx{\norm}[2][1=]{\ifthenelse{\equal{#1}{}}{\left\Vert #2 \right\Vert}{\left\Vert #2 \right\Vert^{#1}}}
\newcommandx{\normLigne}[2][1=]{\ifthenelse{\equal{#1}{}}{\Vert #2 \Vert}{\Vert #2\Vert^{#1}}}

\newcommand{\parenthese}[1]{\left(#1 \right)}

\newcommand{\parentheseDeux}[1]{\left[ #1 \right]}
\newcommand{\defEns}[1]{\left\lbrace #1 \right\rbrace }
\newcommand{\defEnsLigne}[1]{\lbrace #1 \rbrace }

\newcommand{\ps}[2]{\left\langle#1,#2 \right\rangle}
\newcommand{\psLigne}[2]{\langle#1,#2 \rangle}

\def\rset{\mathbb{R}}
\def\nset{\mathbb{N}}

\def\nsets{\mathbb{N}^*}

\newcommandx\probaMarkovTilde[2][2=]
{\ifthenelse{\equal{#2}{}}{{\widetilde{\mathbb{P}}_{#1}}}{\widetilde{\mathbb{P}}_{#1}\left[ #2\right]}}

\newcommand{\expe}[1]{\PE \left[ #1 \right]}
\newcommand{\expeExpo}[2]{\PE^{#1} \left[ #2 \right]}
\newcommand{\expeLigne}[1]{\PE [ #1 ]}
\newcommand{\expeLine}[1]{\PE [ #1 ]}
\newcommand{\expeMarkov}[2]{\PE_{#1} \left[ #2 \right]}

\newcommand{\expeMarkovExpo}[3]{\PE_{#1}^{#2} \left[ #3 \right]}

\newcommand{\Pens}{\mathcal{P}}

\newcommand{\plusinfty}{+\infty}

\newcounter{hypoconbis}
\newcounter{saveconbis}
\newcommand\debutH{\begin{list}
{\textbf{H\arabic{hypoconbis}}}{\usecounter{hypoconbis}}\setcounter{hypoconbis}{\value{saveconbis}}}
\newcommand\finH{\end{list}\setcounter{saveconbis}{\value{hypoconbis}}}

\def\ie{\textit{i.e.}}
\def\eqsp{\;}

\newcommand{\oointLigne}[1]{(#1)}

\newcommand{\ocint}[1]{\left(#1\right]}
\newcommand{\ooint}[1]{\left(#1\right)}
\newcommand{\ccint}[1]{\left[#1\right]}

\newcommandx{\weight}[2][2=n]{\omega_{#1,#2}^N}

\def\as{\ensuremath{\text{a.s.}}}

\def\rmd{\mathrm{d}}
\newcommandx\sequence[3][2=,3=]
{\ifthenelse{\equal{#3}{}}{\ensuremath{\{ #1_{#2}\}}}{\ensuremath{\{ #1_{#2}, \eqsp #2 \in #3 \}}}}
\newcommandx{\sequencen}[2][2=n\in\N]{\ensuremath{\{ #1, \eqsp #2 \}}}
\newcommandx\sequenceDouble[4][3=,4=]
{\ifthenelse{\equal{#3}{}}{\ensuremath{\{ (#1_{#3},#2_{#3}) \}}}{\ensuremath{\{  (#1_{#3},#2_{#3}), \eqsp #3 \in #4 \}}}}
\newcommandx{\sequencenDouble}[3][3=n\in\N]{\ensuremath{\{ (#1_{n},#2_{n}), \eqsp #3 \}}}

\def\iid{i.i.d.}
\def\rme{\mathrm{e}}

\def\eg{\textit{e.g.}}

\newcommand{\BEAS}{\begin{eqnarray*}}
\newcommand{\EEAS}{\end{eqnarray*}}
\newcommand{\BEA}{\begin{eqnarray}}
\newcommand{\EEA}{\end{eqnarray}}
\newcommand{\BEQ}{\begin{equation}}
\newcommand{\EEQ}{\end{equation}}
\newcommand{\BIT}{\begin{itemize}}
\newcommand{\EIT}{\end{itemize}}
\newcommand{\BNUM}{\begin{enumerate}}
\newcommand{\ENUM}{\end{enumerate}}
\newcommand{\BA}{\begin{array}}
\newcommand{\EA}{\end{array}}

\newcommand{\tr}{\mathop{ \rm tr}}

\newcommand{\idm}{I}
\newcommand{\rb}{\mathbb{R}}

\def \E{{\mathbb E}}
\def \R{{\mathbb R}}

\def \E{{\mathbb E}}

\def \N{{\mathbb N}}

\newcommand{\f}[1]{\mathcal{F}_{#1}}

\newcommand{\te}[2]{\theta_{#1}^{(#2)}}
\newcommand{\teD}[1]{\theta_{#1}}

\newcommand{\bte}[2]{\bar \theta_{#1}^{(#2)}}
\def\btheta{\bar{\theta}}
\newcommand{\ts}{\theta^*}

\def\cspace{\mathrm{C}}

\def\Rg{\ensuremath{R_{\gamma}}}
\newcommand{\pg}{\pi_{\gamma}}
\newcommand{\tav}{\bar \theta_{\gamma}}
\newcommandx{\tharg}[2][2=\gamma]{\theta_{#1}^{(#2)}}
\newcommand{\thargD}[2]{\theta_{#1,\gamma}^{#2}}
\newcommand{\thargDD}[2]{\theta_{#1}^{#2}}

\newcommand{\fquad}{f_{\Sigma}}

\newcommand{\tavd}[1]{\bar \theta_{2\gamma}}

\renewcommand{\epsilon}{\varepsilon}

\newcommand{\opnorm}[1]{{\left\vert\kern-0.25ex\left\vert\kern-0.25ex\left\vert #1 
    \right\vert\kern-0.25ex\right\vert\kern-0.25ex\right\vert}}

\def\generator{\mathcal{A}}

\def\momentNoise{\mathrm{\tau}}
\def\bfe{\mathbf{e}}
\def\bff{\mathbf{f}}

\def\Id{\operatorname{Id}}

\def\tildetheta{\tilde{\theta}}

\def\calC{\mathcal{C}}

\newcommandx{\CPE}[3][1=]{{\mathbb E}_{#1}\left[ \left. #2 \middle \vert #3 \right. \right]} 
\newcommandx{\CPVar}[3][1=]{\mathrm{Var}^{#3}_{#1}\left\{ #2 \right\}}
\newcommand{\CPP}[3][]
{\ifthenelse{\equal{#1}{}}{{\mathbb P}\left(\left. #2 \, \right| #3 \right)}{{\mathbb P}_{#1}\left(\left. #2 \, \right | #3 \right)}}

\def\barL{\bar{L}}

\def \rd {\rset^{d}}

\def \unk {\{ 1, \ldots, k\}}
\def \und {\{ 1, \ldots, d\}}

\renewcommand{\epsilon}{\varepsilon}

\def\intrd{\int_{\rset^d}}
\newcommand{\intrdarg}[1]{\int_{\rset^d} \defEns{#1} \pg (\rmd \theta)}
\newcommand{\intrdargD}[1]{\int_{\rset^d} #1 \pg (\rmd \theta)}
\def\pigrmd{\pi_{\gamma}(\rmd \theta)}

\newcommand{\mcb}[1]{\mathcal{B}(#1)}

\def\mcC{\mathcal{C}}

\def\ttau{\tilde{\tau}}
\def\tpsi{\tilde{\psi}}

\def\psig{\ensuremath{\psi_{\gamma}}}
\def\tpsig{\ensuremath{\tpsi_{\gamma}}}
\def\mcf{\mathcal{F}}
\def\varespilon{\varepsilon}
\def\mcr{\mathcal{R}}
\def\mbe{\mathbb{E}}
\def\bfA{\mathbf{A}}
\def\bfB{\mathbf{B}}
\def\bfT{\mathbf{T}}

\def\rborne{r}
\def\zerop{\{ 0,\ldots, p\}}
\def\zerod{\{ 0,\ldots, d\}}

\def\mrd{\mathrm{D}}
\def\mrc{\mathrm{C}}

\def\loss{\mathrm{L}}
\def\lreg{{\ell}}
\def\pmom{p}
\def\ptilde{\tilde{p}}
\def\tildep{\tilde{p}}
\title{Bridging the Gap between Constant Step Size Stochastic Gradient Descent and  Markov Chains}

\usepackage{authblk}

\author[1]{Aymeric Dieuleveut}
\author[2]{Alain Durmus}
\author[1]{Francis Bach}
\affil[1]{INRIA - Département d’informatique de l’ENS, École normale supérieure, CNRS, PSL Research University, 75005 Paris, France}
\affil[2]{CMLA - \'Ecole normale supérieure Paris-Saclay, CNRS, Université Paris-Saclay, 94235 Cachan, France.}

\begin{document}

\maketitle
\begin{abstract}: We consider the minimization of an objective
  function given access to unbiased estimates of its gradient through
  stochastic gradient descent (SGD) with constant step-size. While the
  detailed analysis was only performed for quadratic functions, we
  provide an explicit asymptotic expansion of the moments of the
  averaged SGD iterates that outlines the dependence on initial
  conditions, the effect of noise and the step-size, as well as the
  lack of convergence in the general (non-quadratic) case. For this
  analysis, we bring tools from Markov chain theory into the analysis
  of stochastic gradient.  We then show that Richardson-Romberg
  extrapolation may be used to get closer to the global optimum and we
  show empirical improvements of the new extrapolation scheme.
\end{abstract}

\newcommand{\tcsigne}[1]{\textcolor{red}{#1}}

\section{Introduction}
\label{sec:intro}

We consider the minimization of an objective function  given access to unbiased estimates of the function  gradients. This key methodological problem has raised interest in  different communities: in large-scale machine learning \cite{Bot_Bou_2008,Sha_Sha_Sre_2009,Sha_Sin_Sre_2007},   optimization \cite{Nem_Jud_Lan_2009,Nes_Via_2008}, and   stochastic approximation \cite{kushner:clark:1978, Pol_Jud_1992, Rup_1988}. The most widely used algorithms are stochastic gradient descent (SGD), a.k.a.~Robbins-Monro algorithm \cite{Rob_Mon_1951}, and some of its modifications based on averaging of the iterates~\cite{Pol_Jud_1992,Rak_Sha_Sri_2011,Sha_Zha_2013}.  

While the choice of the step-size may be done robustly in the  {deterministic}  case (see \eg~\cite{Ber_1995}), this remains a traditional theoretical and practical issue in the stochastic case. Indeed, early work suggested to use step-size decaying with the number $k$ of iterations as $O(1/k)$~\cite{Rob_Mon_1951}, but it appeared to be non-robust to ill-conditioning and slower decays such as $O(1/\sqrt{k})$ together with averaging lead to both good practical and theoretical performance~\cite{Bac_2014}.

We consider in this paper constant step-size SGD, which is  often used in practice. Although the algorithm is not converging in general to the global optimum of the objective function, constant step-sizes come with benefits: (a) there is a single parameter value to set as opposed to the several choices of parameters to deal with decaying step-sizes, \eg~as $1/( \square k+\triangle)^\circ$; the initial conditions are forgotten exponentially fast for well-conditioned (\eg~strongly convex) problems~\cite{Ned_Ber_2001,Nee_War_Sre_2014}, and the performance, although not optimal, is sufficient in practice (in a machine learning set-up, being only 0.1\% away from the optimal prediction often does not matter).

The main  goals of this paper are (a) to gain a complete understanding of the properties of constant-step-size SGD in the strongly convex case, and (b) to propose provable improvements to get closer to the optimum when precision matters or in high-dimensional settings. We consider the iterates of the SGD recursion on $\rb^d$ defined starting from $\theta_0 \in \rset^{d}$, for $k \geq 0$, and a step-size  $\gamma>0$  by

\begin{equation}
  \label{eq:def_grad_sto}
\te{k+1}{\gamma} = \te{k}{\gamma} - \gamma \big[  f'(\te{k}{\gamma}) + \varepsilon_{k+1}(\te{k}{\gamma}) \big] \eqsp,
  \end{equation}
where $f$ is the objective function to minimize (in machine learning the generalization performance), $\varepsilon_{k+1}(\te{k}{\gamma})$ the zero-mean statistically independent noise (in machine learning, obtained from a single observation).
Following~\cite{Bac_Mou_2013}, we leverage the property that the sequence of iterates $(\te{k}{\gamma})_{k \geq 0}$ is an  \emph{homogeneous Markov chain}.

\begin{figure}
\centering   \includegraphics[width=.48\linewidth]{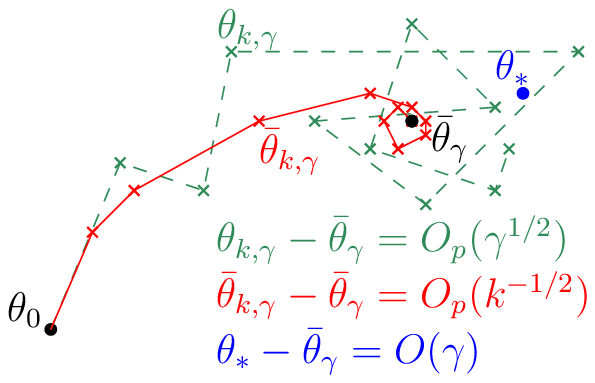} \hspace*{.02\linewidth}
   \includegraphics[width=.48\linewidth]{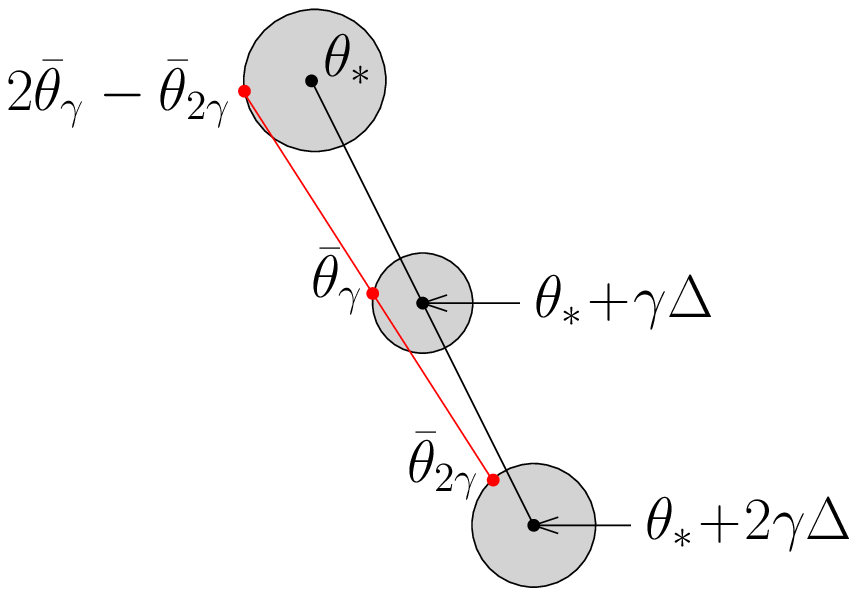}
  \caption{(Left) Convergence of iterates $\te{k}{\gamma}$ and averaged iterates $\bte{k}{\gamma}$ to the mean $\bar{\theta}_{\gamma}$ under the stationary distribution $\pi_\gamma$. (Right) Richardson-Romberg extrapolation, the disks are of radius $O(\gamma^2)$.}
\label{fig:joliplot}
\end{figure}

This interpretation allows us to capture the general behavior of the algorithm. In the strongly convex case, this Markov chain converges exponentially fast to a unique stationary distribution $\pi_\gamma$ (see \Cref{prop:existence_pi}) highlighting the facts that (a) initial conditions of the algorithms are forgotten quickly and (b) the algorithm does not converge to a point but  oscillates around the mean of $\pi_\gamma$. See an illustration in Figure~\ref{fig:joliplot} (left). It is known that the oscillations of the non-averaged iterates have an average magnitude of $\gamma^{1/2}$ \cite{Pfl_1986}.

Consider the process $(\bte{k}{\gamma})_{k \geq 0}$ given for
all $k \geq 0$ by 
\begin{equation}
  \label{eq:def_average}
 \bte{k}{\gamma} = \frac{1}{k+1} \sum_{j=0}^{k}
\te{j}{\gamma} \eqsp. 
\end{equation}
Then under appropriate conditions on the Markov chain
$(\te{k}{\gamma})_{k \geq 0}$, a central limit theorem on
$(\bte{k}{\gamma})_{k \geq 0}$ holds which implies that
$\bte{k}{\gamma} $ converges at rate $O(1/\sqrt{k})$ to
\begin{equation}
\label{eq:def_mean_pi_gamma}
 \bar{\theta}_{\gamma} = \int_{\rset^d} \vartheta \,\rmd \pi_\gamma
(\vartheta)  \eqsp.
\end{equation}
 The deviation between $\bte{k}{\gamma}$ and
 the global optimum $\ts$ is thus composed of a stochastic part
$\bte{k}{\gamma}- \bar{\theta}_{\gamma}$ and a deterministic part $
\bar{\theta}_{\gamma} - \ts$.

For quadratic functions, it turns out that the deterministic part vanishes~\cite{Bac_Mou_2013}, that is,  $\bar{\theta}_{\gamma} = \ts$ and thus averaged SGD with a constant step-size does converge. However, it is not true for general objective functions where we can only show that
$  \bar{\theta}_{\gamma} - \ts = O(\gamma)$, and this deviation is the reason why constant step-size SGD is not convergent.

The first main contribution of the paper is to provide an
explicit asymptotic expansion in the step-size $\gamma$ of
$\btheta_{\gamma}-\ts$. Second, a quantitative version of a central
limit theorem is established which gives a bound on
$\mathbb{E}[\normLigne{\btheta_{\gamma} - \bte{k}{\gamma}}^2]$ that
highlights all dependencies on initial conditions and noise variance,
as achieved for least-squares by~\cite{Def_Bac_2015}, with an
explicit decomposition into ``bias'' and ``variance'' terms: the bias
term characterizes how fast initial conditions are forgotten and is proportional to $\mathrm{N}(\theta_0 - \ts)$, for a suitable norm $\mathrm{N} : \rset^d \to \rset_+$; while the
variance term characterizes the effect of the noise in the gradient,
independently of the starting point, and increases with the covariance
of the noise.

Moreover, akin to weak error results for ergodic diffusions, we achieve a non-asymptotic weak error expansion in the step-size between $\pi_{\gamma}$ and the Dirac measure on $\rset^d$ concentrated at $\ts$. Namely, we prove that for all functions $g: \rset^d \to \rset$, regular enough, $\int_{\rset^d} g(\theta) \rmd  \pi_{\gamma}(\theta) = g(\ts) +   \gamma C_1^{g} + r^{g}_{\gamma}$, $r^{g}_{\gamma} \in \rset^d$, $\norm{r^g_{\gamma}} \leq C_2^g \gamma^2$, for some $C_1^g, C_2^g \geq 0$ independent of $\gamma$. Given this expansion, we can now use a very simple trick from numerical analysis, namely 
Richardson-Romberg extrapolation~\cite{Sto_Bul_2013}: if we run two SGD recursions $(\te{k}{\gamma})_{k \geq 0}$ and $(\te{k}{2\gamma})_{k \geq 0}$ with the two different step-sizes $\gamma$ and $2\gamma$, then the average processes $(\bte{k}{\gamma})_{k \geq 0}$ and $(\bte{k}{2\gamma})_{k \geq 0}$  will converge to  $\tav$ and $\tavd{f}$ respectively. Since $\tav= \ts + \gamma \Delta_1^{\Id}  + r_{\gamma}^{\Id}$ and $\tavd{f} =  \ts + 2 \gamma \Delta_1^{\Id}  + r_{2\gamma}^{\Id}$, for $r_{\gamma}^{\Id}, r_{2\gamma}^{\Id} \in \rset^d$, $\max(\norm{2 r_{\gamma}^{\Id}},\norm{ r_{2\gamma}^{\Id}})\leq 2C\gamma^2$, for $C \geq 0$ and $\Delta \in \rset^d$ independent of $\gamma$, the combined iterates $2\bte{k}{\gamma}-\bte{k}{2\gamma}$ will converge to $ \ts+ 2r_{\gamma}^{\Id}-r_{2\gamma}^{\Id}$ which is closer to $\ts$ by a factor $\gamma$. See illustration in Figure~\ref{fig:joliplot}(right).

In summary, we make the following contributions:

\begin{itemize}
\item  We provide in \Cref{sec:main} an asymptotic expansion in $\gamma$ of $\btheta_{\gamma}-\ts$ and an explicit version of a central limit theorem is given which bounds $\mathbb{E}[\normLigne{\btheta_{\gamma} - \bte{k}{\gamma}}^2]$. These two results outlines the dependence on initial conditions, the effect of noise and the step-size.
\item We show in \Cref{sec:main} that Richardson-Romberg extrapolation may be used to get closer to the global optimum.
\item We bring and adapt in \Cref{sec:MCprop} tools from analysis of discretization of diffusion processes into the one of SGD and create new ones. We believe that this analogy and the associated ideas  are interesting in their own right.
\item We show in \Cref{sec:exp} empirical improvements of the extrapolation schemes.
\end{itemize}

\paragraph{Notations}

We first introduce several notations.  We consider the finite
dimensional euclidean space $ \rset^{d} $ embedded with its canonical
inner product $ \langle\cdot, \cdot\rangle $.
Denote by $\{\bfe_1,\ldots,\bfe_d\}$ the canonical basis of $\rset^d$. Let
$E$ and $F$ be two real vector spaces, denote by $E \otimes F$ the tensor
product of $E$ and $F$.  For all $x \in E$ and $y \in F$ denote by $x
\otimes y \in E \otimes F$ the tensor product of $x$ and $y$.  Denote by $E^{\otimes k}$ the $k^{\text{th}}$ tensor
power of $E$ and $x^{\otimes k} \in E^{\otimes k}$ the
$k^{\text{th}}$ tensor power of $x$.
We let $\mathcal{L}((\rset^{d})^{\otimes k} , \rset^{\ell})$ stand for
the set of linear maps from $(\rset^{n})^{\otimes k} $ to
$\rset^{\ell}$ and for
$\mathrm{L} \in \mathcal{L}((\rset^{d})^{\otimes k} , \rset^{\ell})$,
we denote by $\norm{\mathrm{L}}$ the operator norm of
$\mathrm{L}$.

Let $n \in \nset^*$, denote by $\cspace^n(\rset^d,\rset^m)$ the set of $n$
times continuously differentiable functions from $\rset^d$ to
$\rset^m$. Let $F \in \cspace^n(\rset^d,\rset^m)$, denote by $F^{(n)}$ or $D^nF$, the
$n^{\text{th}}$ differential of $f$. Let $f \in \cspace^n(\rset^d , \rset)$. For any $ x \in \rd $,
$ f^{(n)}(x) $ is a tensor of order $ n $. For example, for all
$x \in \rset^d$, $ f^{(3)} (x) $ is a third order tensor. In addition,
for any $x \in \rset^d$ and any matrix, $ M\in \R^{d\times d} $, we
define $ f^{(3)}(x) M $ as the vector in $ \R^{d}$ given by: for any
$ l\in \und $, the $l^{\text{th}}$ coordinate is given by
$ (f^{(3)}(x) M )_l =\sum_{i,j=1}^{d} M_{i,j} \frac{\partial^{3}
  f}{\partial x_i \partial x_j\partial x_l}(x)$.  By abuse of notations, for
$f \in \cspace^1(\rset^d)$, we identify $f'$ with the gradient of $f$ and if $f \in \cspace^2(\rset^d)$, we identify $f''$ with the Hessian matrix of $f$.  A function $f : \rset^d \to \rset^q$ is said to be
locally Lipschitz if there exists $\alpha \geq 0$ such that for all
$x,y \in \rset^d$,
$\norm{f(x)-f(y)} \leq (1+\norm[\alpha]{x}+\norm[\alpha]{y})
\norm{x-y}$.  For ease of notations and depending on the context, we
consider $M \in \rset^{d \times d}$ either as a matrix or a second
order tensor. More generally, any
$M \in \mathrm{L}((\rset^d)^{\otimes k},\rset)$ will be also consider
as an element of $\mathrm{L}((\rset^d)^{\otimes (k-1)},\rset^d)$ by the canonical bijection. Besides, For any matrices $ M,N \in \R^{d\times d} $,
$ M \otimes N $ is defined as the endomorphism of $\R^{d\times d}$
such that $M \otimes N: P\mapsto M P N $. For any matrix
$ M \in \rset^{d\times d}$, $ \tr (M) $ is the trace of $ M $, \ie~the
sum of diagonal elements of the matrix $ M $.

For $a,b \in \rset$, denote
by $a \vee b$ and $a \wedge b$ the maximum and the minimum of $a$ and
$b$ respectively. Denote by $\floor{\cdot}$ and $\ceil{\cdot}$ the
floor and ceiling function respectively. 

Denote by   $\mathcal{B}(\rset^d)$  the Borel
$\sigma$-field of $\rset^d$.
For all $x \in \rset^d$, $\updelta_x$ stands for the Dirac measure at $x$.  

\section{Main results}

In this section, we describe the assumptions underlying our analysis, describe our main results and their implications.

\label{sec:main}
\subsection{Setting}
\label{sec:setting}
Let $f:\rset^d \to \rset$ be an objective function, satisfying the following assumptions:
\begin{assumption}
\label{hyp:strong_convex}
The function $f$ is strongly convex with convexity constant $\mu>0$, \ie~for all $\theta_1,\theta_2 \in \rset^d$ and $t \in \ccint{0,1}$,
\begin{equation*}
  f(t \theta_1 + (1-t) \theta_2) \leq t f(\theta_1) +(1-t) f(\theta_2) -(\mu/2)t(1-t) \norm[2]{\theta_1-\theta_2} \eqsp.
\end{equation*}
\end{assumption}

\begin{assumption}
\label{hyp:regularity}
The function $f$ is five times continuously differentiable with second to fifth uniformly bounded derivatives: for all $k \in \{2,\ldots,5\}$, $\sup_{\theta\in \rset^d} \norm{f^{(k)}(\theta)} < \plusinfty$. Especially  $f$ is $L$-smooth with $L \geq 0$: for all $ \theta_1,\theta_2\in \rset^{d}$
\begin{equation*}
\norm{  f'(\theta_1)-f'(\theta_2)} \leq L\norm{\theta_1-\theta_2} \eqsp.
\end{equation*}
\end{assumption}
If there exists a positive definite matrix $\Sigma \in \R^{d\times d}$, such that  the function $f$ is the quadratic function $ \theta \mapsto \normLigne{\Sigma^{1/2} (\theta -\ts)}^2/2$, then Assumptions~\Cref{hyp:strong_convex},~\Cref{hyp:regularity} are satisfied.


In the definition of SGD given by \eqref{eq:def_grad_sto}, $(\epsilon_{k})_{k \geq 1}$ is a sequence of random functions from $\rset^d$ to $\rset^d$ satisfying the following properties.
\begin{assumption}\label{ass:def_noise}
{There exists a filtration $(\f{k})_{k \geq 0}$ (\ie~for all $k \in \nset$, $\f{k} \subset \f{k+1}$) on some probability space $(\Omega, \mathcal{F}, \mathbb{P})$ such that  for any  $k\in \nset$ and $\theta \in \rset^{d}$, $\epsilon_{k+1}(\theta)$ is a  $\f{k+1}$-measurable random variable and $\expe{\epsilon_{k+1}(\theta)|\f{k}}=0$.
In addition, $(\epsilon_k) _{k\in \nset^*}  $ are independent and identically distributed (\iid) random fields. Moreover, we assume that  $ \theta_0 $ is $ \f{0} $-measurable.}
\end{assumption}
\Cref{ass:def_noise}  expresses that  we have access to an \iid~sequence $(f'_k)_{k \in \nsets}$ of unbiased estimator of $f'$, \ie~for all $k\in \nset$ and $\theta \in \rset^d$,
\begin{equation}
  \label{eq:def_f_prime_k}
 f'_{k+1}(\theta) = f'(\theta) +\epsilon_{k+1}(\theta) \eqsp. 
\end{equation}
Note that we do not assume random vectors~$ ( \epsilon_{k+1}(\te{k}{\gamma}) )_{k\in \nset}$ to be  \iid, a stronger assumption generally referred to as the semi-stochastic setting. Moreover, as~$ \theta_0 $ is $ \f{0} $-measurable, for any $ k\in \nset $,  $ \theta_k $ is $ \f{k} $-measurable. 

We also consider the following conditions on the noise, for $ p\geq 2 $:
\begin{assumption}[$p$]\label{ass:lip_noisy_gradient_AS} 
	For any $ k\in \nset^{*} $, $f'_k$ is almost surely $L$-co-coercive (with the same constant as in \Cref{hyp:regularity}): that is, for any $ \eta, \theta\in \R^{d} $, $ L \ps{f'_k(\theta)-f'_k(\eta)}{\theta-\eta} \geq \norm[2]{f'_k(\theta)-f'_k(\eta)} $.
	Moreover, there exists $ \tau_p\geq 0 $, such that  for any $ k\in \nset^{*} $, 
	$\PE^{1/p}[\norm{\epsilon_k(\ts)}^p] \le \tau_p$.
\end{assumption}

Almost sure $L$-co-coercivity \cite{Zhu_Mar_1995} is for example
satisfied if for any $ k\in \nset^{*} $, there exists a random function
$ f_k $ such that $ f'_k = (f_k)' $ and which is \as~convex and
$ L$-smooth. Weaker assumptions on the noise are discussed
in~\Cref{app:noisediscussion}. Finally we emphasize that under
\Cref{ass:def_noise} then to verify that
\Cref{ass:lip_noisy_gradient_AS}($p$)  holds, $p \geq 2$, it suffices to show that
$f'_1$ is almost surely $L$-co-coercive and 	$\PE^{1/p}[\norm{\epsilon_1(\ts)}^p] \le \tau_p$.
Under \Cref{ass:def_noise}-\Cref{ass:lip_noisy_gradient_AS}$(2)$, consider the function $\mcC : \rset^d \to \rset^{d \times d}$ defined for all $\theta \in \rset^d$ by
\begin{equation}
  \label{eq:def_cov_matrix}
\mcC(\theta) =  \expe{\epsilon_1(\theta)^{\otimes 2}} \eqsp.
\end{equation}
\begin{assumption}
	\label{assum:reguarity_noise}
	The function $\calC$
	is three time continuously differentiable and there exist $M_{\epsilon}, k_\epsilon \geq 0$ such that for all $\theta \in \rset^d$,
        \begin{equation*}
\max_{i \in \{1,2,3\}}  \norm{\mcC^{(i)}(\theta)  } \leq M_{\epsilon} \defEns{1+\norm[k_{\epsilon}]{\theta-\ts}} \eqsp.
        \end{equation*}
\end{assumption}
In other words, we assume that the covariance matrix $\theta \mapsto C(\theta) $ is a regular enough function, which is satisfied in natural settings.

\begin{example}[Learning from \iid~observations]
  \label{ex:iid_observation_learning}
  Our main motivation comes
from machine learning; consider two sets $\mathcal{X}, \mathcal{Y}$
and a convex loss function
$\loss : \mathcal{X}\times \mathcal{Y} \times \rset^{d} \to \rset$. The
objective function is the generalization error
$f_\loss (\theta ) = \E_{X,Y} [\loss(X, Y, \theta) ]$, where $(X,Y)$ are
some random variables. Given \iid~observations
$ (X_k, Y_k)_{k \in \nsets } $ with the same distribution as $(X,Y)$,
for any $k \in \nsets$, we define $f_k(\cdot) = \loss(X_k,Y_k, \cdot)$
the loss with respect to observation $ k $. SGD then corresponds to
following gradient of the loss on a single independent observation
$(X_k,Y_k)$ at each step; Assumption {\Cref{ass:def_noise}} is then
satisfied with $\f{k}=\sigma((X_j, Y_j)_{j \in \unk})$.

Two classical situations are worth mentioning. On the first hand,  in \emph{least-squares regression},  $\mathcal{X} =\rset^{d}$, $\mathcal{Y}=\R$, and the loss function is $\loss(X,Y, \theta) = (\langle X, \theta\rangle -Y)^{2}$. Then $f_\Sigma$ is the quadratic function
$\theta \mapsto \normLigne{\Sigma^{1/2} (\theta -\ts)}^2/2$, with  $\Sigma= \expeLine{XX^{\top}}$, which satisfies Assumption~{\Cref{hyp:regularity}}. For any $\theta \in \rset^d$,
\begin{equation}
\label{eq:eps_reg_line}
  \epsilon_k(\theta) = X_k X_k^{\top}\theta - X_k Y_k
\end{equation}
 Then, for any $ p\geq 2 $, Assumption~{\Cref{ass:lip_noisy_gradient_AS}}(p) and \Cref{assum:reguarity_noise}  is satisfied as soon as  observations are \as~bounded, while {\Cref{hyp:strong_convex}} is satisfied if the second moment matrix is invertible or additional regularization is added. In this setting, $\epsilon_k$ can be decomposed as $\epsilon_k =  \varrho_k + \xi_k$ where  $\varrho_k$ is the multiplicative part, $\xi_k$ the additive part,  given for $ \theta \in \rset^d$ by $\varrho_k(\theta)= (X_k X_k^{\top}-\Sigma) (\theta-\ts)$ and 
  \begin{equation}
    \label{eq:def_xi_k_addit_part}
 \xi_k =   (X_k^{\top} \ts - Y_k)X_k    \eqsp.
  \end{equation}
For all $k \geq 1$, $\xi_k$ does not depend on $\theta$. This two parts in the noise will appear in~\Cref{cor:quadconv}.  
Finally assume that there exists $ \rborne \geq 0 $ such that
\begin{equation}
  \label{eq:def_rborne}
\expeLigne{\norm[2]{X_k} X_k X_k ^{\top}} \preccurlyeq \rborne^{2} \Sigma \eqsp,  
\end{equation}
 then \Cref{ass:lip_noisy_gradient_AS}$(4)$ is satisfied. This assumption is satisfied, \eg, for \as~bounded data, or for data with bounded kurtosis, see~\cite{Die_Fla_Bac_2016} for details.

On the other hand, in \emph{logistic regression}, where
$\loss(X,Y, \theta) = \log (1+ \exp(-Y\langle X, \theta\rangle ))$. Assumptions {\Cref{ass:lip_noisy_gradient_AS}} or {\Cref{hyp:regularity}} are similarly satisfied, while 
\Cref{hyp:strong_convex} needs an additional restriction to a compact set.
\end{example}

\subsection{Summary and discussion of main results}

Under the stated assumptions, for all $\gamma \in \ooint{0,2/L}$
and $\theta_0 \in \rset^d$, the Markov chain $(\tharg{k})_{k \geq 0}$
converges in a certain sense specified below to a probability measure
on $(\rset^d, \mathcal{B}(\rset^d))$, $\pi_{\gamma}$ satisfying $\int_{\rset^d} \norm{\vartheta}^2  \pi_{\gamma}(\rmd \vartheta)< \plusinfty$, see \Cref{prop:existence_pi} in
\Cref{sec:MCprop}.  In the next section, by two different methods
(\Cref{theo:statio_general} and \Cref{THEO:BIAS1}), we show that under
suitable conditions on $f$ and the noise $(\epsilon_k)_{k \geq 1}$,
there exists $\Delta \in \rset^d$ such that for all $\gamma \geq 0$,
small enough
\begin{equation*}
\bar{\theta}_{\gamma} =  \int_{\rset^d} \vartheta \pi_{\gamma}(\rmd \vartheta) = \ts +  \gamma \Delta+r^{(1)}_{\gamma} \eqsp,
\end{equation*}
where $r_\gamma^{(1)} \in \rset^d$, $\normLigne{r_{\gamma}^{(1)}} \leq C\gamma^2 $ for some constant $C \geq 0$ independent of $\gamma$.
Using \Cref{prop:existence_pi}, we get that for all $k \geq 1$,
\begin{equation}
\label{eq:drift}
\PE[\bte{k}{\gamma}  - \ts]= \frac{A(\theta_0,\gamma)}{k} +  \gamma \Delta+ r_{\gamma}^{(2)}\eqsp, 
\end{equation}
where $r_\gamma^{(2)} \in \rset^d , \, \normLigne{r_{\gamma}^{(2)}} \leq C(\gamma^2 + \rme^{-k \mu \gamma}) $ for some constant $C \geq 0$ independent of $\gamma$.

This expansion in the step-size $\gamma$ shows that a
Richardson-Romberg extrapolation can be used to have better estimates
of $\ts$.  Consider the average iterates $(\bte{2\gamma}{k})_{k \geq
  0}$ and $(\bte{k}{\gamma})_{k \geq 0}$ associated with SGD with step
size $2\gamma$ and $\gamma$ respectively. Then \eqref{eq:drift} shows
that $(2 \bte{k}{\gamma}- \bte{k}{2\gamma})_{k \geq 0}$ satisfies
\begin{equation*}
  \PE[ 2 \bte{k}{\gamma}- \bte{k}{2\gamma}  - \ts] = \frac{2 A(\theta_0,\gamma)-A(\theta_0,2\gamma)}{k} + 2r^{(2)}_{\gamma} - r^{(2)}_{2\gamma} \eqsp,
\end{equation*}
and therefore is closer to the optimum  $\ts$. This very simple trick improves the convergence by a factor of $\gamma$ (at the expense of a slight increase of the variance). In practice, while the un-averaged gradient iterate $\te{k}{\gamma}$ saturates rapidly, $\bte{k}{\gamma}$ may already perform well enough to avoid saturation on real data-sets \cite{Bac_Mou_2013}. The Richardson-Romberg extrapolated iterate $ 2 \bte{k}{\gamma}- \bte{k}{2\gamma}$ very rarely reaches saturation in practice. This appears in synthetic experiments presented in \Cref{sec:exp}. Moreover, this procedure only requires to compute two parallel SGD recursions, either with the same inputs, or with different ones, and is naturally  parallelizable.

In \Cref{sec:convergenceres}, we give a quantitative version of a central limit theorem for $(\bte{k}{\gamma})_{k \geq 0}$, for a fixed $\gamma>0$ and $k$ going to $\plusinfty$ : under appropriate conditions, 
there exist constants $B_1(\gamma)$ and $B_2(\gamma)$ such that 
\begin{equation}
\label{eq:tcl_nonasympt}
  \expe{\norm[2]{\bte{k}{\gamma}-\tav}} = B_1(\gamma)/k + B_2(\gamma)/k^2 \eqsp.
\end{equation}

Combining \eqref{eq:drift} and \eqref{eq:tcl_nonasympt} characterizes the bias/variance trade-off of SGD used to estimate $\ts$.

\subsection{Related work}

The idea to study stochastic approximation algorithms using results
and techniques from the Markov chain literature is not new. It goes
back to \cite{freidlin:wentzell:1998}, which shows under appropriate
conditions that solutions of stochastic differential equations (SDE)
\begin{equation*}
  \rmd Y_t = - f'(Y_t) \rmd t +\gamma_t \rmd B_t\eqsp,
\end{equation*}
where $(B_t)_{t \geq 0}$ is a $d$-dimensional Brownian motion and
$(\gamma_t)_{t \geq 0}$ is a one-dimensional positive function,
$\lim_{t \to \plusinfty} \gamma_t = 0$, converge in probability to
some minima of $f$. An other example is
\cite{priouret:veretenikov:1998} which extends the classical Foster-Lyapunov
criterion from Markov chain theory (see \cite{Mey_Twe_2009}) to study the stability of the LMS
algorithm. In \cite{bouton:pages:1997}, the authors are interested in the
convergence of the multidimensional Kohonen algorithm. They show
that the Markov chain defined by this algorithm is  uniformly ergodic and
derive asymptotic properties on its limiting distribution.

The techniques we use in this paper to establish our results
share a lot of similarities with previous work. For example, our first
results in \Cref{sec:statprop} and \Cref{sec:convergenceres} regarding
an asymptotic expansion in $\gamma$ of $\btheta_{\gamma}-\ts$ and an
explicit version of a central limit theorem is given which bounds
$\mathbb{E}[\normLigne{\btheta_{\gamma} - \bte{k}{\gamma}}^2]$, can be
seen as complementary results of
\cite{aguech:moulines:priouret:2000}. Indeed, in
\cite{aguech:moulines:priouret:2000}, the authors decompose the
tracking error of a general algorithm in a linear regression model. To
prove their result, they develop the error using a perturbation
approach, which is quite similar to what we do.

Another and significant point of view to study stochastic
approximation relies on the gradient flow equation associated with the
vector field $ f'$: $\dot{x}_t = - f'(x_t)$. This approach
was introduced by \cite{ljung:1977} and \cite{kushner:clark:1978} and
have been applied in numerous papers since then, see
\cite{metivier:priouret:1984,metivier:priouret:1987,benveniste:metivier:priouret:1990,benaim:1996,tadic:doucet:2017}.
We use to establish our result in \Cref{sec:cont-interpr}, the strong
connection between SGD and the gradient flow equation as well.  The
combination of the relation between stochastic approximation
algorithms with the gradient flow equation and the Markov chain theory
have been developed in \cite{fort:pages:1995} and
\cite{fort:pages:1999}. In particular, \cite{fort:pages:1999}
establishes under certain conditions that there exists for all
$\gamma \in \ooint{0,\gamma_0}$, with $\gamma_0$ small enough, an
invariant distribution $\pi_{\gamma}$ for the Markov chain
$(\theta_k^{(\gamma)})_{k \in \nset}$, and
$(\pi_{\gamma})_{\gamma\in\ooint{0,\gamma_0}}$ is tight. In addition,
they show that any limiting distributions is invariant for the
gradient flow associated with $\nabla f$. Note that their conditions
and results are different from ours. In particular, we do not assume
that $(\theta_k^{(\gamma)})_{k \in \nset}$ is Feller but require that
$f$ is strongly convex contrary to \cite{fort:pages:1999}.

To the authors knowledge, the use of the
Richardson-Romberg method for stochastic approximation has only been
considered in \cite{moulines:priouret:roueff:2005} to recover the
minimax rate for recursive estimation of time varying autoregressive
process.

Several attempts have been made to improve convergence of
SGD. \cite{Bac_Mou_2013} proposed an online Newton algorithm which
converges in practice to the optimal point with constant step-size but
has no convergence guarantees.  The quadratic case was studied by
\cite{Bac_Mou_2013}, for the (uniform) average iterate: the variance
term is upper bounded by $ {\sigma^2 d}/{n}$ and the squared bias term
by $\|\ts\|^{2} / (\gamma n)$. This last term was improved to
$\|\Sigma^{-1/2} \ts\|^{2} / (\gamma n)^2$ by
\cite{Def_Bac_2015,Die_Bac_2015}, showing that asymptotically, the
bias term is negligible, see also~\cite{Lan_2012}.  Analysis has been
extended to ``tail averaging''~\cite{Jai_Kak_Kid_2016}, to improve the
dependence on the initial conditions. Note that this procedure can be
seen as a Richardson-Romberg trick with respect to $k$. Other
strategies were suggested to improve the speed at which initial
conditions were forgotten, for example using acceleration when the
noise is additive~\cite{Die_Fla_Bac_2016,Jai_Kak_Kid_2017}. A
criterion to check when SGD with constant step size is close to its
limit distribution was recently proposed in~\cite{Che_2017}.

In the context of discretization of ergodic diffusions, weak error estimates between the stationary distribution of the discretization and the invariant distribution of the associated diffusion have been first shown by \cite{Tal_Tub_1990} and \cite{Mat_Stu_Hig_2002} in the case of the Euler-Maruyama scheme. Then, \cite{Tal_Tub_1990} suggested the use of Richardson-Romberg interpolation to improve the accuracy of estimates of integrals with respect to the invariant distribution of the diffusion.  Extension of these results have been obtained for other types of discretization by   
\cite{Abd_Vil_Zyg_2014} and \cite{Che_Din_Car_2015}. We show in \Cref{sec:cont-interpr} that a weak error expansion in the step-size $\gamma$ also holds for SGD between $\pi_{\gamma}$ and $\updelta_{\ts}$. Interestingly as to the Euler-Maruyama discretization, SGD has a weak error of order $\gamma$. 
In addition, \cite{Dur_cSi_Mou_2016} proposed and analyzed the use of Richardson-Romberg extrapolation applied to the stochastic gradient Langevin dynamics (SGLD) algorithm. This method introduced by \cite{Wel_Teh_2011} combines SGD and the Euler-Maruyama discretization of the Langevin diffusion associated to a target probability measure~\cite{Dal_2017,durmus:moulines:2017}. Note that this method is however completely different from SGD, in part because Gaussian noise of order $\gamma^{1/2}$ (instead of $\gamma$) is injected in SGD  which changes the overall dynamics.

Finally, it is worth mentioning
\cite{mandt:hoffman:blei:2016,mandt:hoffman:blei:2017} which are
interested in showing that the invariant measure of constant step-size
SGD for an appropriate choice of the step-size $\gamma$, can be used
as a proxy to approximate the target distribution $\pi$ with density
with respect to the Lebesgue measure $\rme^{-f}$. Note that the
perspective and purpose of this paper is completely different since we
are interested in optimizing the function $f$ and not in sampling from
$\pi$.


\section{Detailed analysis} 
\label{sec:mainresults}
\label{sec:MCprop}
In this Section, we describe in detail our approach. A first step is
to describe the existence of a unique stationary distribution $\pg$
for the Markov chain $(\tharg{k})_{k \geq 0}$ and the convergence of
this Markov chain to $\pg$ in the Wasserstein distance of order
$2$.

\paragraph{Limit distribution}

We cast in this section SGD in the Markov chain framework and
introduce basic notion related to this theory, see \cite{Mey_Twe_2009}
for an introduction to this topic. Consider the Markov kernel
$R_{\gamma}$ on $(\rset^d, \mcb{\rset^d})$ associated with SGD
iterates $(\tharg{k})_{k \in \nset}$, \ie~for all $k \in \nset$ and
$\mathsf{A} \in \mcb{\rset^d}$, almost surely $R_{\gamma}(\theta_{k},
\mathsf{A}) = \mathbb{P}(\theta_{k+1} \in \mathsf{A} | \theta_k)$, for
all $\theta_0 \in \rset^d$ and $\mathsf{A} \in \mcb{\rset^d}$, $\theta
\mapsto R_{\gamma}(\theta,\mathsf{A})$ is Borel measurable and
$R_{\gamma}(\theta_0,\cdot)$ is a probability measure on
$(\rset^d,\mcb{\rset^d})$.  For all $k \in \nset^*$, we define the Markov kernel
$R_{\gamma}^k$ recursively by $R_{\gamma}^1 = R_{\gamma}$ and for $k
\geq 1$, for all $\theta_0 \in \rset^d$ and $\mathsf{A} \in \mcb{\rset^d}$
\begin{equation*}
  R_{\gamma}^{k+1}(\theta_0,\mathsf{A}) = \int_{\rset^d} R_{\gamma}^k(\theta_0, \rmd\theta) R_{\gamma}(\theta, \mathsf{A}) \eqsp.
\end{equation*}
For any probability measure $\lambda$ on
$(\rset^d,\mcb{\rset^d})$, we define the probability measure $\lambda R_{\gamma}$  for all $\mathsf{A} \in \mcb{\rset^d}$ by
\begin{equation*}
  \lambda R_{\gamma}^{k}  (\mathsf{A}) = \int_{\rset^d} \lambda(\rmd \theta) R_{\gamma}^k(\theta, \mathsf{A}) \eqsp.
\end{equation*}
By definition, for all probability measure $\lambda$ on
$\mcb{\rset^d}$ and $k \in \nset^*$, $\lambda R^{k}_{\gamma}$ is the
distribution of $\tharg{k}$ started from $\theta_0$ drawn from $\lambda$.
For any function $\phi : \rset^d \to \rset_+$ and $k \in \nset^*$, define the measurable
function $R_{\gamma}^k \phi : \rset^d \to \rset$ for all $\theta_0 \in \rset^d$,
\begin{equation*}
  R_{\gamma}^k \phi(\theta_0) = \int_{\rset^d} \phi(\theta)  R_{\gamma}^k (\theta_0,\rmd \theta) \eqsp.
\end{equation*}
For any measure $\lambda$ on $(\rset^d,\mathcal{B}(\rset^d))$ and any
measurable function $h: \rset^d \to \rset$, $\lambda (h)$ denotes
$\int_{\rset^d} h(\theta) d\lambda(\theta)$ when it exists.  Note that
with such notations, for any $ k\in \nset^* $, probability measure
$\lambda $ on $\mathcal{B}(\rset^d)$, measurable function
$h: \rset^d \to \rset_+$, we have
$ \lambda (\Rg^{k} h) = (\lambda \Rg^{k}) (h)$. A probability
measure $\pg$ on $(\rset^d, \mathcal{B}(\rset^d))$ is said to be a
invariant probability measure for $R_{\gamma}$, $\gamma >0$, if
$\pg R_{\gamma} = R_{\gamma}$.  A Markov chain $(\tharg{k})_{k \in \nset}$ satisfying the SGD recursion \eqref{eq:def_grad_sto} for $\gamma >0$ will be said at stationarity if it admits a invariant measure $\pg$ and $\tharg{k}$ is distributed according to $\pg$. Note that in this case for all $k \in \nset$, the distribution of $\tharg{k}$ is $\pg$.

To show that $(\tharg{k})_{k \geq 0}$ admits a unique stationary
distribution $\pi_{\gamma}$ and quantify the convergence of $(\nu_0
R_{\gamma}^k)_{k \geq 0}$ to $\pi_{\gamma}$, we use  the
Wasserstein distance.  
A probability measure $\lambda$ on $(\rset^d, \mathcal{B}(\rset^d))$
is said to have a finite second moment if
$\int_{\rset^d} \norm{\vartheta}^2 \lambda(\rmd \vartheta)<
\plusinfty$. The set of probability measure on
$(\rset^d, \mathcal{B}(\rset^d))$ having a finite second moment is
denoted by $\mathcal{P}_2(\rset^d)$.  For all probability measures
$\nu$ and $\lambda$ in $\mathcal{P}_2(\rset^d)$, 
define the \emph{Wasserstein distance} of order $2$ between $\lambda$
and $\nu$ by
\begin{equation*}
W_2(\lambda, \nu)= \inf
_{\xi \in \Pi(\lambda , \nu)} \Big(\int \| x-y\|^{2 } \xi(dx,
dy)\Big)^{1/2}  \eqsp, 
\end{equation*}  
 where $\Pi(\mu, \nu)$ is the set of
probability measure $\xi$ on $\mathcal{B}(\R^{d} \times \R^{d})$ satisfying for all $\mathsf{A} \in \mathcal{B}(\rset^d)$, $\xi(\mathsf{A} \times \rset^d) = \nu(\mathsf{A})$, $\xi( \rset^d \times \mathsf{A} ) = \lambda(\mathsf{A})$.

\begin{proposition}\label{prop:existence_pi}
Assume \Cref{hyp:strong_convex}-\Cref{hyp:regularity}-\Cref{ass:def_noise}-\Cref{ass:lip_noisy_gradient_AS}(2). For any step-size $\gamma \in\oointLigne{0, 2/L}$, the Markov chain $(\te{k}{\gamma})_{k \geq 0}$, defined by the recursion \eqref{eq:def_grad_sto}, admits a unique stationary distribution $\pi_{\gamma} \in \Pens_2(\rset^d)$.  In addition 
\begin{enumerate}[label=(\alph*)]
\item \label{prop:existence_pi_eq_1} for all $\theta \in \R^{d}$, $k\in \nsets$:  
\begin{equation*}  
W_2^{2}( R^k_\gamma(\theta, \cdot),  \pi_\gamma) \le (1 - 2\mu  \gamma (1-  \gamma L/2))^{k}  \int_{\rset^d} \norm[2]{\theta-\vartheta } \rmd \pi_\gamma(\vartheta)\eqsp;
\end{equation*}
\item \label{prop:existence_pi_item_2} for any Lipshitz function $\phi : \rset^d \to \rset$, with Lipschitz constant $L_\phi$, for all
  $\theta \in \R^{d}$, $k\in \nsets$:
  \begin{equation*}
  \abs{\Rg^{k} \phi(\theta)-\pg(\phi)} \leq   L_{\phi} (1 - 2\mu  \gamma (1-  \gamma L/2))^{k/2}  \left(\int \|\theta-\vartheta\|^{2}\rmd\pi_\gamma (\vartheta)\right)^{1/2} \eqsp.
  \end{equation*}
\end{enumerate}
\end{proposition}

\begin{proof}
  Let $\gamma \in \oointLigne{0,2/L}$ and  $\lambda_1,\lambda_2 \in \Pens_2(\rset^d)$. By \cite[Theorem
  4.1]{villani:2009}, there exists a couple of random variables
  $\theta^{(1)}_0,\theta^{(2)}_0$ such that
  $W_2^2(\lambda_1,\lambda_2) =
  \PE[\normLigne{\theta^{(1)}_0-\theta^{(2)}_0}^2]$ independent of $(\epsilon_k)_{k \in \nsets}$. Let
  $(\thargDD{k}{(1)})_{k \geq 0}$,$(\thargDD{k}{(2)})_{k \geq 0}$ be the
  SGD iterates associated with the step-size $\gamma$, starting from
  $\theta^{(1)}_0$ and $\theta^{(2)}_0$ respectively and sharing the
  same noise, \ie~for all $k \geq 0$,
  \begin{equation}
  \label{eq:def_coupling}
  \begin{cases}
  \thargDD{k+1}{(1)}&=  \thargDD{k}{(1)} - \gamma \big[  f'( \thargDD{k}{(1)}) + \varepsilon_{k+1}( \thargDD{k}{(1)}) \big] \\
  \thargDD{k+1}{(2)} &= \thargDD{k}{(2)} - \gamma \big[  f'( \thargDD{k}{(2)}) + \varepsilon_{k+1}( \thargDD{k}{(2)}) \big]\eqsp.
  \end{cases}
\end{equation}
Note that using that $\theta^{(1)}_0,\theta^{(2)}_0$ are independent
of $\varepsilon_1$, we have for $i,j\in\{1,2\}$ using
\Cref{ass:def_noise}, that
\begin{equation}
  \label{eq:indep_noise_initial_cond}
  \expeLigne{\psLigne{\theta^{(i)}_0}{\varepsilon(\theta^{(j)}_0)}} =
0 \eqsp. 
\end{equation}
Since for all $k \geq 0$, the distribution of
$(\thargDD{k}{(1)},\thargDD{k}{(2)})$ belongs to
$\Pi(\lambda_1 R_{\gamma}^{k},\lambda_2 R_{\gamma}^{k})$,
 by definition of the Wasserstein distance we get 
  \begin{align*}
  	&W_2^{2}( \lambda_1 \Rg, \lambda_2  \Rg) \le \E \left[ \|\thargDD{1}{(1)}-\thargDD{1}{(2)}\|^2\right] \\
  	&\le  \expe{ \|\theta^{(1)}_0-\gamma f_1'(\theta^{(1)}_0) -(\theta^{(2)}_0- \gamma f_1'(\theta^{(2)}_0))) \|^2 }\\
  	& \overset{i)}{\le} \expe{ \norm {\theta^{(1)}_0-\theta^{(2)}_0} ^{2} - 2 \gamma \Big\langle  f'(\theta^{(1)}_0)- f'(\theta^{(2)}_0), \theta^{(1)}_0-\theta^{(2)}_0\Big\rangle}  \\ &  \qquad \qquad\qquad \qquad\qquad \qquad + \gamma^{2}\expe{\norm[2] {f_1'(\theta^{(1)}_0)- f_1'(\theta^{(2)}_0)}}\\
  	& \overset{ii)}{\le}   \expe{\norm {\theta^{(1)}_0-\theta^{(2)}_0} ^{2}  - 2 \gamma (1-  \gamma L/2) \Big\langle  f'(\theta^{(1)}_0)- f'(\theta^{(2)}_0), \theta^{(1)}_0-\theta^{(2)}_0\Big\rangle} \\
  	& \overset{iii)}{\le}  (1-2\mu  \gamma (1-  \gamma L/2) )  \expe{ \norm {\theta^{(1)}_0-\theta^{(2)}_0} ^{2}} \eqsp,
  \end{align*} 
  using \eqref{eq:indep_noise_initial_cond} for $i)$, \Cref{ass:lip_noisy_gradient_AS}$(2)$  for $ii)$, and finally \Cref{hyp:strong_convex} for $iii)$.
  
  Thus by a straightforward induction, we get, setting $\rho =(1-2\mu  \gamma (1-  \gamma L/2) )$
  \begin{align}
  \nonumber
  W_2^{2}( \lambda_1 \Rg^k, \lambda_2  \Rg^k) &\le 
  \E \left[ \|\thargDD{k}{(1)}-\thargDD{k}{(2)}\|^2\right] \\
  \label{eq:coupling_2}
  & \leq  \rho \E \left[ \|\thargDD{k-1}{(1)}-\thargDD{k-1}{(2)}\|^2\right]
  \, {\le} \,  \rho^{k} W_2^2(\lambda_1,\lambda_2) \eqsp, 
  \end{align}
Since by \Cref{hyp:regularity}-\Cref{ass:def_noise}-\Cref{ass:lip_noisy_gradient_AS}$(2)$, $\lambda_1 R_\gamma \in \Pens_2(\rset^d)$, taking $ \lambda_2=\lambda_1 R_{\gamma}$ in \eqref{eq:coupling_2}, for any $N \in \nsets$, we have $    \sum_{k=1}^N   W_2^{2}( \lambda_1 \Rg^k, \lambda_2  \Rg^k) \leq     \sum_{k=1}^N  \rho^k  W_2^{2}( \lambda_1 , \lambda_1  R_{\gamma})$.
Therefore, we get $    \sum_{k=1}^{\plusinfty}   W_2^{2}( \lambda_1 \Rg^k, \lambda_1  \Rg^{k+1})< \plusinfty$. 
  By \cite[Theorem
  6.16]{villani:2009}, the space $\mathcal{P}_2(\rset^d)$ endowed with $W_2$ is a Polish space. Then,  $(\lambda_1 \Rg^{k})_{k\geq 0}$ is a Cauchy sequence and converges to a limit $\pi_\gamma^{\lambda_1} \in \mathcal{P}_2(\rset^d)$: 
  \begin{equation}
  \label{eq:convergence_wasser}
  \lim_{k \to \plusinfty} W_2(\lambda_1 \Rg^k,  \pi_\gamma^{\lambda_1})=  0  \eqsp.
\end{equation}
We show that the limit $\pi_\gamma^{\lambda_1}$ does not depend on
$\lambda_1$. Assume that there exists $\pi_{\gamma}^{\lambda_2}$ such
that
$ \lim_{k \to \plusinfty} W_2(\lambda_2 \Rg^k,
\pi_\gamma^{\lambda_2})= 0 $. By the triangle inequality
  $$W_2( \pi_\gamma^{\lambda_1}, \pi_\gamma^{\lambda_2} ) \le W_2(
  \pi_\gamma^{\lambda_1},\lambda_1 \Rg^k )+ W_2( \lambda_1
  \Rg^k, \lambda_2 \Rg^k)+ W_2(\pi_\gamma^{\lambda_2}, \lambda_2
  \Rg^k) \eqsp.$$ Thus by \eqref{eq:coupling_2} and \eqref{eq:convergence_wasser}, taking the limits as $k\to \plusinfty$, we
  get $W_2( \pi_\gamma^{\lambda_1}, \pi_\gamma^{\lambda_2} )=0$ and
  $\pi_\gamma^{\lambda_1}=\pi_\gamma^{\lambda_2}$. The limit is thus
  the same for all initial distributions and is denoted by
  $\pi_{\gamma}$.
  
  Moreover, $\pi_\gamma$ is invariant for $R_{\gamma}$. Indeed for all $k \in \nset^{*}$,
  \begin{equation*}
  W_2(\pi_{\gamma} R_{\gamma} , \pi_{\gamma}) \leq    W_2(\pi_{\gamma} R_{\gamma} , \pi_{\gamma} R_{\gamma}^k)+  W_2(\pi_{\gamma} R_{\gamma}^k,\pi_{\gamma}) \eqsp.
  \end{equation*}
  Using \eqref{eq:coupling_2} and \eqref{eq:convergence_wasser}, we get taking $k \to \plusinfty$, $   W_2(\pi_{\gamma} R_{\gamma} , \pi_{\gamma})=0$ and $\pi_{\gamma} R_{\gamma}  = \pi_{\gamma}$. The fact that $\pi_{\gamma}$ is the unique stationary distribution is straightforward by contradiction and using \eqref{eq:coupling_2}.
  
  Taking $\lambda_1 =\updelta_\theta$, $\lambda_2 =\pg$, using the invariance of $\pi_\gamma$  
  and \eqref{eq:coupling_2}, we get \ref{prop:existence_pi_eq_1}.

  Finally, if we take $\lambda_1= \updelta_\theta$ and $\lambda_2= \pi_{\gamma}$, using $\pg \Rg = \pg$, \eqref{eq:coupling_2}, and the Cauchy-Schwarz inequality, we have for  any $k\in \nset^{*}$:
\begin{align*}
\abs{\Rg^{k} \phi(\theta)-\pg(\phi)}& = \abs{\expe{\phi(\thargD{k}{(1)})-\phi(\thargD{k}{(2)}))}} {\le} L_\phi  \PE^{1/2}[\norm[2]{\thargD{k}{(1)}-\thargD{k}{(2)}}] \nonumber \\
&{\le} L_\phi  (1 - 2\mu  \gamma (1-  \gamma L/2))^{k/2} \left(\int \|\theta-\vartheta\|^{2}\rmd\pi_\gamma (\vartheta)\right)^{1/2} \eqsp, 
\end{align*}
which concludes the proof of \ref{prop:existence_pi_item_2}.

\end{proof}
A consequence of \Cref{prop:existence_pi} is that the expectation of $\bte{k}{\gamma}$ defined by \eqref{eq:def_average}   converges to $\int_{\rset^d} \vartheta \rmd \pi_{\gamma}(\vartheta)$ as $k$ goes to infinity at a rate of order $O(k^{-1})$, see \Cref{theo:convergence_loc_lip_wasserstein} in \Cref{sec:exist-poiss-solut}.

\subsection{Expansion of moments of $\pg$ when $\gamma$ is in a neighborhood of $ 0$}
\label{sec:statprop}
In this sub-section, we analyze the properties of the chain starting at $\theta_0$ distributed according to $\pi_{\gamma}$. As a result, we prove that the mean of the stationary distribution $\tav = \int_{\rset^d } \vartheta \pi_{\gamma} \, (\rmd \vartheta)$ is such that $\tav  = \ts + \gamma \Delta + O(\gamma^{2})$. Simple developments of Equation~\eqref{eq:def_grad_sto} at the equilibrium, result in  expansions of the first two moments of the chain.   It extends \cite{Pfl_1986, Lju_Pfl_Wal_1992} which showed  that $(\gamma ^{-1/2} (\pi_\gamma -\updelta_{\ts}))_{\gamma >0}$ converges in distribution to a normal law as $\gamma \to 0$. 
 

   \paragraph{Quadratic case} When $f$ is a quadratic function, \ie~$f'$ is affine, we have the following result.
   \begin{proposition}
     \label{lem:statio_quadractic}
  Assume $f = \fquad$, $\fquad: \theta \mapsto \norm{ \Sigma^{1/2} (\theta-\ts)}^{2}/2$, where $\Sigma$ is a positive definite matrix, and
\Cref{hyp:regularity}-\Cref{ass:def_noise}-\Cref{ass:lip_noisy_gradient_AS}(4). Let
  $\gamma \in \ooint{0, 2/L}$. Then, it holds $\tav = \ts$, $\Sigma \otimes I + I\otimes \Sigma - \gamma \Sigma \otimes \Sigma$ is invertible and 
 	\begin{equation*}		
   \intrd (\theta-\ts)^{\otimes 2}\pigrmd = \gamma  (\Sigma \otimes I + I\otimes \Sigma - \gamma \Sigma \otimes \Sigma ) ^{-1} \parentheseDeux{\intrd \mathcal{C}(\theta) \pigrmd} \eqsp,
\end{equation*}
    where $\tav$ and  $\mcC$ are given by \eqref{eq:def_mean_pi_gamma} and \eqref{eq:def_cov_matrix} respectively, and $\pi_{\gamma}$ is the invariant probability measure of $R_{\gamma}$ given by \Cref{prop:existence_pi}.
\end{proposition}

The first part of the result, which highlights the crucial fact that for a quadratic function, the mean under the limit distribution is the optimal point, is easy to prove. Indeed, since $\pi_{\gamma}$ is invariant for $(\tharg{k})_{ k \geq 0}$, if $\te{0}{\gamma}$ is distributed according to $ \pg$, then $\te{1}{\gamma}$ is distributed according to $ \pg$ as well. Thus as $
\te{1}{\gamma}= \te{0}{\gamma}- \gamma  f'(\te{0}{\gamma}) + \gamma \epsilon_{1}(\te{0}{\gamma}) $
taking expectations on both sides, we get $\int_{\rset^d}  f'(\vartheta) \rmd \pi_{\gamma}(\vartheta) =0$. For a quadratic function, whose gradient is linear: $ \int_{\rset^d}  f'(\vartheta) \rmd \pi_{\gamma}(\vartheta) = f'( \tav) = 0$ and thus $\tav = \ts$. This implies that the averaged iterate converges to $\ts$, see \eg~\cite{Bac_Mou_2013}. The proof for the second expression is given in~\Cref{sec:proof-crefl_quad}.



\paragraph{General case} While the quadratic case led to particularly simple  expressions, in general, we can only get a first order development of these expectations as $\gamma\to 0$. Note that it improved on \cite{Pfl_1986}, which shows a similar expansion but an error of order of $O(\gamma^{3/2})$.
\begin{theorem}
  \label{theo:statio_general}
Assume
  \Cref{hyp:strong_convex}-\Cref{hyp:regularity}-\Cref{ass:def_noise}-\Cref{ass:lip_noisy_gradient_AS}($6\vee[2(k_{\epsilon}+1)]$)-\Cref{assum:reguarity_noise}
  and let $\gamma \in \ooint{0,2/L}$. Then $f''(\ts)\otimes \idm + \idm \otimes  f''(\ts)$ is invertible and 
  \begin{align}
\label{theo:statio_general_eq_1}
&\tav - \ts = \gamma  f''(\ts)^{-1}
 f'''(\ts) 
  \bfA  \mcC(\ts)  + O(\gamma^{2}) \\
    \label{theo:statio_general_eq_2}
& \int_{\rset^d}(\theta - \ts)^{\otimes 2} \pi_{\gamma}(\rmd \theta) 
=  \gamma  \bfA \mcC(\ts) 
+ O(\gamma^{2}) \eqsp,
\end{align}
where
\begin{equation}
\label{eq:def_bfA}
\bfA   = \left( f''(\ts)\otimes \idm + \idm \otimes  f''(\ts)\right)^{-1}\eqsp,
\end{equation}
$\tav$ and  $\mcC$ are given by \eqref{eq:def_mean_pi_gamma} and \eqref{eq:def_cov_matrix} respectively, and $\pi_{\gamma}$ is the invariant probability measure of $R_{\gamma}$ given by \Cref{prop:existence_pi}.
\end{theorem}
\begin{proof}
  The proof is postponed to \Cref{sec:proof-crefth_statio_general}.
\end{proof}
This shows that $\gamma \mapsto \tav$ is a differentiable function at $\gamma=0$. The ``drift'' $\tav - \ts$ can be understood as an additional error occurring because the function is non quadratic ($f'''(\ts) \not =0$) and the step-sizes are not decaying to zero. The mean under the limit distribution is at distance $\gamma$ from $\ts$. In comparison, the final iterate oscillates in a sphere of radius proportional to $\sqrt{\gamma}$.
 
\subsection{Expansion for a given $\gamma>0$ when $k$ tends to $+\infty$}
\label{sec:convergenceres}
In this sub-section, we analyze the convergence of $\bte{k}{\gamma}$ to $\tav$, when $k\rightarrow \infty$, and the convergence of $\expeLigne{\normLigne[2]{\bte{k}{\gamma}-\tav}}$ to 0. 
Under suitable conditions \cite{glynn:meyn:1996}, $\bte{k}{\gamma}$ satisfies a central limit theorem: $\{\sqrt{k} (\bte{k}{\gamma} - \tav )\}_{k \in \nsets}$ converges in law to a $d$-dimensional Gaussian distribution with zero-mean. However, this result is purely asymptotic and we propose a new tighter development that describes how the initial conditions are forgotten. We show that the convergence behaves similarly to the convergence in the quadratic case, where the expected squared distance decomposes as a sum of a bias term, that scales as $k^{-2}$, and a variance term, that scales as $k^{-1}$, plus linearly decaying residual terms. We also describe how the asymptotic bias and variance can be easily expressed as moments of solutions associated to several \textit{Poisson equations}.

For any  Lipschitz function $\varphi: \rb^d \to \rb^{q}$, 
by \Cref{lem:poisson_exist} in \Cref{sec:exist-poiss-solut}, the function  $\psi_{\gamma} = \sum_{i=0}^{\plusinfty}
\{R_{\gamma}^i \varphi - \pi_{\gamma}(\varphi) \}$ is well-defined, Lipschitz and satisfies $\pi_{\gamma}(\psi_{\gamma}) = 0 $,
$(\Id- \Rg) \psi_{\gamma} = \varphi$.  $\psi_{\gamma}$  will be referred to as the \emph{Poisson
solution} associated with $\varphi$.
Consider the three following functions:
 \begin{itemize}
 	\item $\psi_{\gamma}$ the Poisson solution associated to $\varphi: \theta \mapsto \theta -\ts$, 
 	\item  $\varpi_{\gamma}$ the Poisson solution associated to $\theta \mapsto \psi_{\gamma}(\theta)$,
 	\item $\chi^{1}_{\gamma}$ the Poisson solution associated to $\theta \mapsto (\psi_{\gamma} (\theta))^{\otimes 2}$,
 	\item   $\chi^{2}_{\gamma}$ the Poisson solution associated to $\theta \mapsto ((\psi_{\gamma}-\varphi) (\theta))^{\otimes 2}$.
 \end{itemize}  
\begin{theorem}\label{th:tcl_non_asympt}
Assume \Cref{hyp:strong_convex}-\Cref{hyp:regularity}-\Cref{ass:def_noise}-\Cref{ass:lip_noisy_gradient_AS}(4) and let $\gamma \in \ooint{0, 1/(2L)}$.
Then setting $\rho=  (1 - \gamma \mu)^{1/2}$, for any starting point $\theta_0 \in \rset^d$, $k \in \nsets$,
\begin{align}
  \nonumber
 & \expe{\bte{k}{\gamma}-\tav}   =  k^{-1}( \psi_{\gamma}(\theta_0) + O(\rho^{k}))\eqsp,\\
    \nonumber
&\expe{ \left(\bte{k}{\gamma} - \tav \right)^{\otimes 2}}   = 
k^{-1}\,  \pi_\gamma \left(  \psi_{\gamma}^{\otimes 2} -(\psi_{\gamma} -\varphi)^{\otimes 2} \right)
  \\ 
  \nonumber
& \qquad  \qquad -k^{-2}\left[ \pi_\gamma\left (\varpi_\gamma \varphi^{\top}+\varphi \varpi_\gamma^{\top}\right  )+ \chi^{2}_{\gamma}(\theta_0) - \chi^{1}_{\gamma}(\theta_0) \right]+ O(k^{-3})\eqsp,
\end{align}
where $\bte{k}{\gamma}$, $\tav$ are given by \eqref{eq:def_average} and \eqref{eq:def_mean_pi_gamma} respectively, and $\pi_{\gamma}$ is the invariant probability measure of $R_{\gamma}$ given by \Cref{prop:existence_pi}.
\end{theorem}

Equation~\eqref{th:tcl_non_asympt} is a sum of three terms: (i) a variance term, that scales as $1/k$, and does not depend on the initial distribution (but only on the asymptotic distribution $\pg$), and (ii) a bias term, which scales as ${1/k^{2}}$, and depends on the initial point $\theta_0\in \rset^d$, (iii) a non-positive residual term, which scales as ${1/k^{2}}$.
\begin{proof}

 In order to give the intuition of the proof and to underline how the associated Poisson solutions are introduced,   
we here sketch the proof of the first result. By definition of $\varphi : \theta \mapsto \theta -\ts$ and since $\psi_{\gamma}$ satisfies $(\Id-R_{\gamma}) \psi_{\gamma} = \varphi$, we have 
\begin{equation*}
	\expe{\bte{k+1}{\gamma}}-\ts= (k+1)^{-1}\sum_{i=0}^{k} (\Rg^{i} \varphi)(\theta_0) = \pg(\varphi) +(k+1)^{-1}\psi_{\gamma}(\theta_0) + \Rg^{k+1} \psi_{\gamma}(\theta_0),
\end{equation*} 
where we have used that
\begin{equation*}
                                                            \sum_{i=0}^{\infty} \Rg^{i} (\varphi- \pg(\varphi)) - \Rg^{k+1}\sum_{i=0}^{\infty} \Rg^{i}(\varphi- \pg(\varphi))= \psi_\gamma- \Rg^{k+1}\psi_\gamma \eqsp.
\end{equation*}
Finally, we have that $\Rg^{k} \psi_{\gamma}(\theta_0)$ converges to 0 at linear speed, using 
\Cref{prop:existence_pi} and $\pg(\psi_{\gamma}) =0$.

The formal and complete proof of this result is postponed to \Cref{app:convergencetheorem}.
\end{proof}


This result gives an exact closed form for the asymptotic bias and variance, for a fixed $\gamma$, as $k\rightarrow \infty$.  Unfortunately, in the general case, it is neither possible to compute the Poisson solutions exactly, nor is it possible to prove a first order development of the limits as $\gamma\rightarrow 0$. 

When $\fquad$ is a quadratic function, it is possible, for any $\gamma>0$, to compute $\psi_{\gamma}$ and $\chi^{1,2}_{ \gamma}$ explicitly; we get the following decomposition of the error, which exactly recovers the result of \cite{aguech:moulines:priouret:2000} or \cite{Def_Bac_2015}. 
\begin{corollary}\label{cor:quadconv}
  Assume that $f$ is an objective function of a least-square
  regression problem, \ie~with the notations of
  \Cref{ex:iid_observation_learning}, $f = f_{\Sigma}$,
  $\Sigma = \expeLigne{X X^{\top}}$, $\epsilon_k$ are defined by
  \eqref{eq:eps_reg_line}, and step-size $\gamma \le 1/\rborne^{2}$, with $\rborne$ defined by \eqref{eq:def_rborne}.  Assume
  \Cref{hyp:strong_convex}-\Cref{hyp:regularity}-\Cref{ass:def_noise}-\Cref{ass:lip_noisy_gradient_AS}(4).
For any starting point $\theta_0 \in \rset^d$ :
	 \begin{align*}
		\E \bar{\theta}_k^{(\gamma)} - \ts & {=}   (1/(k\gamma)) \Sigma^{-1} ( \theta_0 - \ts) + O(\rho ^{k})\\
           \expe{\left (\bte{k}{\gamma} -\ts\right )^{\otimes 2}} & {=}  (1/k) \Sigma^{-1}\defEns{\int_{\rset^d} \calC(\theta) \rmd \pi_{\gamma}(\theta) } \Sigma^{-1} \\
&    \qquad       + (1/(k^2\gamma^2)) \Sigma^{-1}   \Omega  \left[\varphi(\theta_0)^{\otimes 2}   
		-   \pg (\varphi^{\otimes 2} ) \right] \Sigma^{-1} \\
		& \qquad	 -(1/(k^2\gamma^2)) (\Sigma^{-2}\otimes \Id+ \Id \otimes \Sigma^{-2}) \pi_\gamma(\varphi^{\otimes 2})+ O(k^{-3}) \eqsp .
	\end{align*}
	With $\Omega=(\Sigma \otimes I + I \otimes \Sigma - \gamma \Sigma \otimes \Sigma)  (\Sigma \otimes I + I \otimes \Sigma - \gamma \bfT)^{-1}$, and
        \begin{equation}
          \label{eq:def_operator_T}
         \bfT : \rset^{d \times d } \to \rset^{d \times d} \eqsp, \,  A \mapsto \expe{(X^{\top} A X) X X^{\top}} \eqsp.
        \end{equation}
\end{corollary}
\begin{proof}
  The proof is postponed  to the supplementary paper  \cite{dieuleveut:durmus:bach:sup:2018}, Section S3.
\end{proof}
The bound on the second order moment is composed of a variance term $  k^{-1} \Sigma^{-1} \pi_\gamma(\calC) \Sigma^{-1}  $, a bias term  which decays as $ k^{-2} $, and a non-positive residual term. Interestingly, the bias is $ 0 $ if we start under the limit distribution.


\subsection{Continuous interpretation of SGD and weak error expansion}
\label{sec:cont-interpr}
Under the stated assumptions on $f$ and  $(\epsilon_{k})_{k \in \nset^*}$, we have analyzed the convergence of the stochastic gradient recursion~\eqref{eq:def_grad_sto}. We here describe how this recursion can be seen as a noisy discretization of the following  \emph{gradient flow} equation, for $t\in \R_+$:
\begin{equation}
  \label{eq:gradientflow}
 \dot \theta_t =- f'(\theta_t) \eqsp.
\end{equation}
Note that since $ f'(\ts) = 0$ by definition of $\ts$ and \Cref{hyp:strong_convex}, then $\ts$ is an equilibrium point of \eqref{eq:gradientflow}, \ie~$\theta_t = \ts$ for all $t \geq 0$ if $\theta_0 = \ts$.
Under \Cref{hyp:regularity}, \eqref{eq:gradientflow} admits a unique
solution on $\rset_+$ for any starting point $\theta \in \rset^d$.  Denote by
$(\varphi_t)_{t \geq 0}$ the flow of \eqref{eq:gradientflow},  defined for
all $\theta \in \rset^d$ by $(\varphi_t(\theta))_{t \geq 0}$ as the solution of \eqref{eq:gradientflow} starting at $\theta$.

Denote by $(\generator,D(\generator))$, the \textit{infinitesimal generator} associated with the flow $(\varphi_t)_{t\geq 0}$ defined by
\begin{align}
\nonumber
\mrd(\generator) &= \defEns{h: \rset^d \to \rset \,: \, \text{ for all $\theta \in \rset^d$, } \lim_{t \to 0} \frac{h(\varphi_t(\theta)) - h(\theta)}{t} \text{ exists } } \\
\label{eq:def_generator}
\generator h(\theta) &= \lim_{t \to 0}\frac{\defEns{h(\varphi_t(\theta)) - h(\theta)}}{t} \text{ for all } h \in \mrd(\generator) \eqsp,  \ \ \theta \in \rset^d \eqsp.
\end{align}
Note that for any $h \in \mrc^1(\rset^d)$, $h \in D(\generator)$,  $
  \generator h = -\ps{ f'}{ h'} \eqsp.$
  
  Under \Cref{hyp:strong_convex} and \Cref{hyp:regularity},  for any locally
  Lipschitz function $g: \rset^{d} \to \rset$ (extension to a function  $g: \rset^{d} \to \rset^{q}$ can easily be done  considering all assumptions and results coordinatewise), denote by $h_g$  the solution of the continuous Poisson
equation defined for all $\theta \in \rset^d$ by $h_g(\theta)=\int_{0}^{\infty} \left (g(\varphi_s(\theta))- g(\ts)\right ) \rmd s $. Note that $h_g$ is well-defined by \Cref{lem:flow_properties}-\ref{item:lem:flow_properties_2} in \Cref{app:subsec:gradientflow},  since $g$ is assumed to be locally Lipschitz.
By \eqref{eq:def_generator}, we have for all $g: \rset^d \to \rset$, locally Lipschitz,
\begin{equation}
{  \generator h_g(\theta) =  g(\ts)-  g(\theta) \eqsp. }
\end{equation}
Under regularity assumptions on $g$ (see \Cref{theo:regularity_poisson}), $h_g$ is continuously differentiable and therefore satisfies $\ps{f'}{h_g'} = g - g(\ts)$. The idea is then to make a Taylor expansion of $h_g(\tharg{k+1})$ around $\tharg{k}$ to express $k^{-1}\sum_{i=1}^k g(\tharg{i}) - g(\ts)$ as convergent terms involving  the derivatives of $h_g$.
For $g: \rset^d \to \rset$ and $\lreg,\pmom \in \nset$, $\lreg \geq 1$ consider the following assumptions.
\begin{assumption}[$\lreg,\pmom$]  
  \label{assum:assum_f_test}
  There exist  $a_g,b_g \in \rset_+$ such that $g \in \mrc^{\lreg}(\rset^d)$ and for all $\theta \in \rset^d$ and $i \in \{1,\cdots,\lreg\}$, $ \norm{ g^{(i)}(\theta)} \leq a_g\defEns{\norm[\pmom]{\theta-\ts}+b_g}$.
\end{assumption}


\begin{theorem}
  \label{THEO:BIAS1}
        Let $g: \rset^d \to \rset$ satisfying
        \Cref{assum:assum_f_test}($5,\pmom$) for $\pmom \in
        \nset$. Assume
        \Cref{hyp:strong_convex}-\Cref{hyp:regularity}-\Cref{ass:def_noise}-\Cref{assum:reguarity_noise}.
        Furthermore, suppose that there exists $q \in \nset$ and $C \geq 0$ such that for all $\theta \in \rset^d$,
        \begin{equation*}
          \expe{\norm{\epsilon_1(\theta)}^{\pmom+k_{\varepsilon}+3}} \leq C(1+\norm[q]{\theta-\ts}) \eqsp,
        \end{equation*}
        and \Cref{ass:lip_noisy_gradient_AS}($2\tilde{p}$) holds for
        $\tilde{p} = \pmom+3+q \vee k_{\varepsilon}$.  Then there
        exists a constant $\varsigma >0$ only depending on $\ptilde$
        such that for all $\gamma \in \ooint{0,1/(\varsigma L)}$,
        $k \in \nset^*$ and any starting point  $\theta_0 \in \rset^d$ it holds that:
  \begin{multline}
\label{eq:expansion_sgd}
 {    \expe{k^{-1} \sum_{i=1}^k \defEns{g(\tharg{i})-g(\ts)}} = 
(1/(k \gamma))\defEns{h_g(\theta_0)-\expe{h_g(\tharg{k+1})}} }
\\ {+(\gamma/2)    \tr\left(h_g''(\ts)\, \mcC(\ts) \right)
- (\gamma/k) A_1(\theta_0) - \gamma^2 A_2(\theta_0,k)
  \eqsp,}
  \end{multline}
where $\tharg{k}$ is the Markov chain starting from $\theta_0$ and defined by the recursion \eqref{eq:def_grad_sto} and $\mcC$ is given by \eqref{eq:def_cov_matrix}.
In addition for some constant $C \geq 0$ independent of $\gamma$ and $k$, we have
\begin{equation*}
A_1(\theta_0)  \leq C \defEns{1+\norm[\tildep]{\theta_0-\ts}} \eqsp,\eqsp A_2(\theta_0,k)  \leq C \defEns{1+\norm[\tildep]{\theta_0-\ts}/k} \eqsp.
\end{equation*}
\end{theorem}
\begin{proof}
  The proof is postponed to \Cref{sec:proof-crefmc:th}.
\end{proof}

First in the case where $f'$ is linear, choosing for $g$ the identity
function, then $h_{\Id} = \int_0^{\plusinfty} \defEnsLigne{\varphi_s -
  \ts }\rmd s = \Sigma^{-1}$, and we get that the first term in
\eqref{eq:expansion_sgd} vanishes which is expected since in that case
$\bar{\theta}_{\gamma} = \ts$.  Second by
\Cref{lem:lip_g}-\ref{item:third_derivative_poisson}, we recover the
first expansion of \Cref{theo:statio_general} for arbitrary objective
functions $f$.  Finally note that for all $q \in \nset$, under
appropriate conditions, \Cref{THEO:BIAS1} implies that there exist constants 
$C_1,C_2(\theta_0) \geq 0$ such that $\expeMarkov{}{k^{-1}
  \sum_{i=1}^k \normLigne[2q] {\tharg{i}-\ts} } = C_1 \gamma +
C_2(\theta_0)/k + O(\gamma^2) $.

\subsection{Discussion}
Classical proofs of convergence rely on another decomposition, originally proposed by \cite{Nem_Yud_1983} and used in recent papers analyzing the averaged iterate \cite{Bac_Mou_2011} . We here sketch the arguments of these decompositions, in order to highlight the main difference, namely the fact that the residual term is not well controlled when $ \gamma $ goes to zero in the classical proof.

\paragraph{Classical decomposition}
 The starting point of this decomposition is to consider a Taylor expansion of $ f'(\te{k+1}{\gamma}) $ around $ \ts $.  For any $ k\in \nset $,
\begin{eqnarray*}
	f'(\te{k}{\gamma}) &=& f''(\ts) (\te{k}{\gamma}-\ts) + O\left(\norm[2]{\te{k}{\gamma}-\ts}\right) .
\end{eqnarray*}
As a consequence, using the definition of the SGD recursion \eqref{eq:def_grad_sto}, 
\begin{eqnarray*}
	\te{k+1}{\gamma}-\te{k}{\gamma} &=&  - \gamma   f'(\te{k}{\gamma})  - \gamma  \varepsilon_{k+1}(\te{k}{\gamma})\\
	&=& -	\gamma f''(\ts) (\te{k}{\gamma}-\ts)   - \gamma  \varepsilon_{k+1}(\te{k}{\gamma}) + 	\gamma  O\left(\norm[2]{\te{k}{\gamma}-\ts}\right) \eqsp.
\end{eqnarray*}
Thus
\begin{eqnarray*}
	f''(\ts) (\te{k}{\gamma}-\ts) 	&=& \gamma^{-1} (-\te{k+1}{\gamma}+\te{k}{\gamma} )  -   \varepsilon_{k+1}(\te{k}{\gamma}) + 	  O\left(\norm[2]{\te{k}{\gamma}-\ts}\right) \eqsp.
\end{eqnarray*}
Averaging over the first $ k$ iterates yields:
\begin{eqnarray}\label{eq:nemirjud}
(k+1)\left (\bte{k}{\gamma}-\ts \right )	&=&\gamma^{-1} f''(\ts) ^{-1} \left (\te{0}{\gamma}-\te{k+1}{\gamma}\right ) - \sum_{i=0}^{k} f''(\ts) ^{-1}\varepsilon_{i+1}\left (\te{i}{\gamma}\right ) \nonumber\\
&+& \sum_{i=0}^{k} O\left(\norm[2]{\te{i}{\gamma}-\ts}\right).
\end{eqnarray}
The term on the right-hand part of Equation \eqref{eq:nemirjud} is composed of a bias term (depending on the initial condition), a variance term, and a residual term.
This residual term  differentiates the general setting from the quadratic one (in which it  does not appear, as the first order Taylor expansion of $ f' $ is exact).
This decomposition has been used in \cite{Bac_Mou_2011} to prove upper bound on the error, but  does not allow for a tight decomposition in powers of $ \gamma $ when $ \gamma \to 0 $. Indeed, the residual $ \te{i}{\gamma}-\ts $ simply does not go to 0 when $ \gamma  \to 0 $: on the contrary, the chain becomes ill-conditioned when $ \gamma=0 $. 

\paragraph{New decomposition}
Here, we use the fact that for a function $ g: \rset^{d} \to \rset^{q}$ regular enough, there exists  $ h_g:  \rset^{d} \to \rset^{q} $ satisfying, for any $ \theta \in \rset^d $:
\begin{equation*}
	h_g'(\theta )f'(\theta ) = g(\theta) - g(\ts),
\end{equation*}
where  $h_g'(\theta ) \in \rset^{q\times d}$, and $ f'(\theta ) \in \rset^{d} $.
The starting point is then a first order Taylor development of $h_g(\tharg{k+1})$ around $\tharg{k}$. For any $ k\in \nsets$, we have
\begin{align*}
&	h_g(\tharg{k+1}) = h_g(\tharg{k}) +  {h_g'(\tharg{k})}{(\tharg{k+1}-\tharg{k})} + O\left(\norm[2]{\tharg{k+1}-\tharg{k}}\right)\\
& 	= h_g(\tharg{k}) - \gamma  {h_g'(\tharg{k})}{    f'(\te{k}{\gamma}) } - \gamma  {h_g'(\tharg{k})}{  \varepsilon_{k+1}(\te{k}{\gamma})} + O\left(\norm[2]{\tharg{k+1}-\tharg{k}}\right) \\
& 	= h_g(\tharg{k}) - \gamma (g(\tharg{k}) - g(\ts) )- \gamma  {h_g'(\tharg{k})}{  \varepsilon_{k+1}(\te{k}{\gamma})} + O\left(\norm[2]{\tharg{k+1}-\tharg{k}}\right).
\end{align*}
Thus reorganizing terms,
\begin{multline*}
  g(\tharg{k}) - g(\ts)  = \gamma^{-1} \defEns{ h_g(\tharg{k})-h_g(\tharg{k+1}) } \\
  +   {h_g'(\tharg{k})}{  \varepsilon_{k+1}(\te{k}{\gamma})} + \gamma^{-1} O\left(\norm[2]{\tharg{k+1}-\tharg{k}}\right).
\end{multline*}
Finally,  averaging over the first $ k $ iterations and taking $g = \Id$ give
\begin{align}
	(k+1) \left (\bte{k}{\gamma} - \ts \right ) =& \gamma^{-1} \left(h_{\Id}(\tharg{0})- h_{\Id}(\tharg{k+1}) \right) +   \sum_{i=0}^{k}{h_{\Id}'(\tharg{i})}{  \varepsilon_{i+1}\left (\te{i}{\gamma}\right )} \nonumber\\
	& + \gamma^{-1} \sum_{i=0}^{k} O\left(\norm[2]{\tharg{i+1}-\tharg{i}}\right)\eqsp. \label{mc:eq:decnew}
\end{align}
This expansion is the root of the proof of \Cref{THEO:BIAS1}, which formalizes the expansion as powers of $ \gamma $. The key difference between decomposition \eqref{eq:nemirjud} and \eqref{mc:eq:decnew} is that in the latter, when $ \gamma \to 0 $, the expectation of the residual term  tends to 0 and can naturally be controlled.


\section{Experiments}
\label{sec:exp}

\begin{figure}[h]
  	\begin{center}
	\begin{tabular}{p{0.01cm}cp{0.01cm}c}
		\begin{minipage}[t]{1cm}
			\rotatebox{90}{ \hspace{0.2cm}
				$\log_{10} \left[ f(\theta)-f(\theta _*) \right]  $ }
		\end{minipage}	
		&
			
		\includegraphics[width=4.2cm]{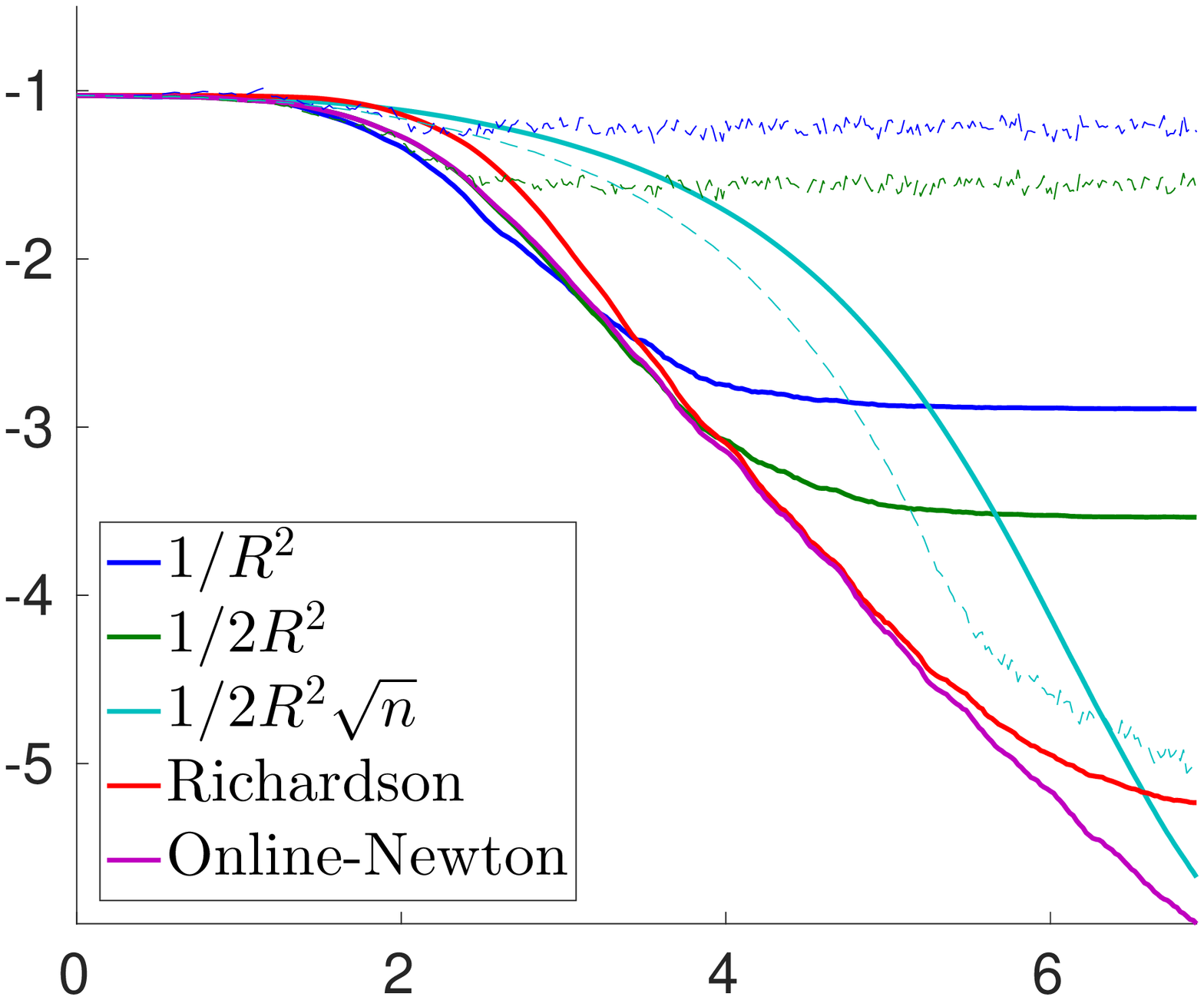}
		&
		\begin{minipage}[t]{2cm}
							\rotatebox{90}{ \hspace{0.2cm}
								$\log_{10} \left[ f(\theta)-f(\theta _*) \right]  $ }
						\end{minipage}	
						& 
                                                  \includegraphics[width=4.2cm]{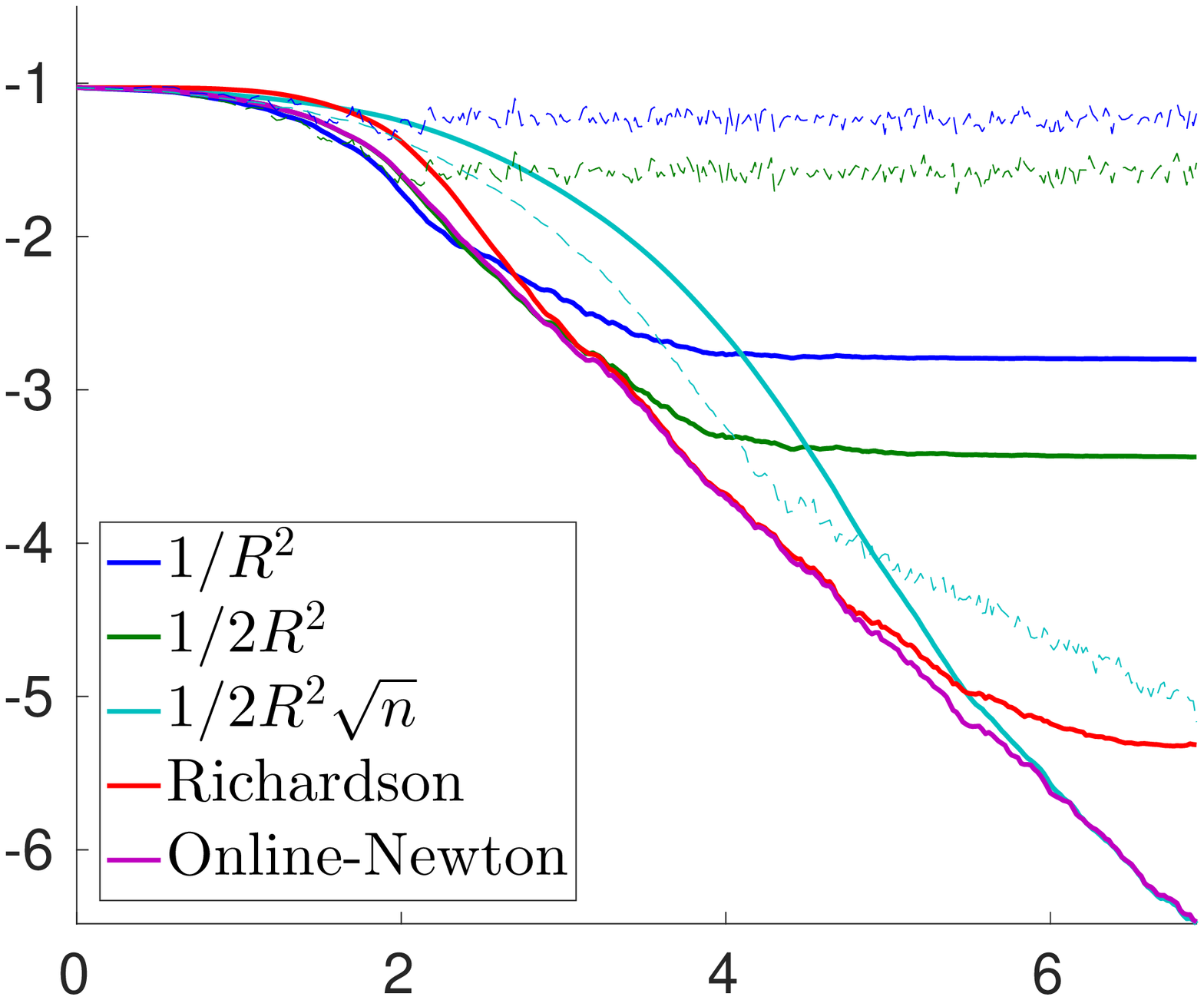}
                                                  \\
		& $\log_{10}(n)$&	& $\log_{10}(n)$	
        \end{tabular}
        \begin{tabular}{p{0.1cm}c}
\begin{minipage}[t]{2cm}
`			\rotatebox{90}{ \hspace{0.2cm}	$\log_{10} \left[ f(\theta)-f(\theta _*) \right]  $ } \, \, 
		\end{minipage}	
		&                                                                                                                      \includegraphics[width=4.2cm]{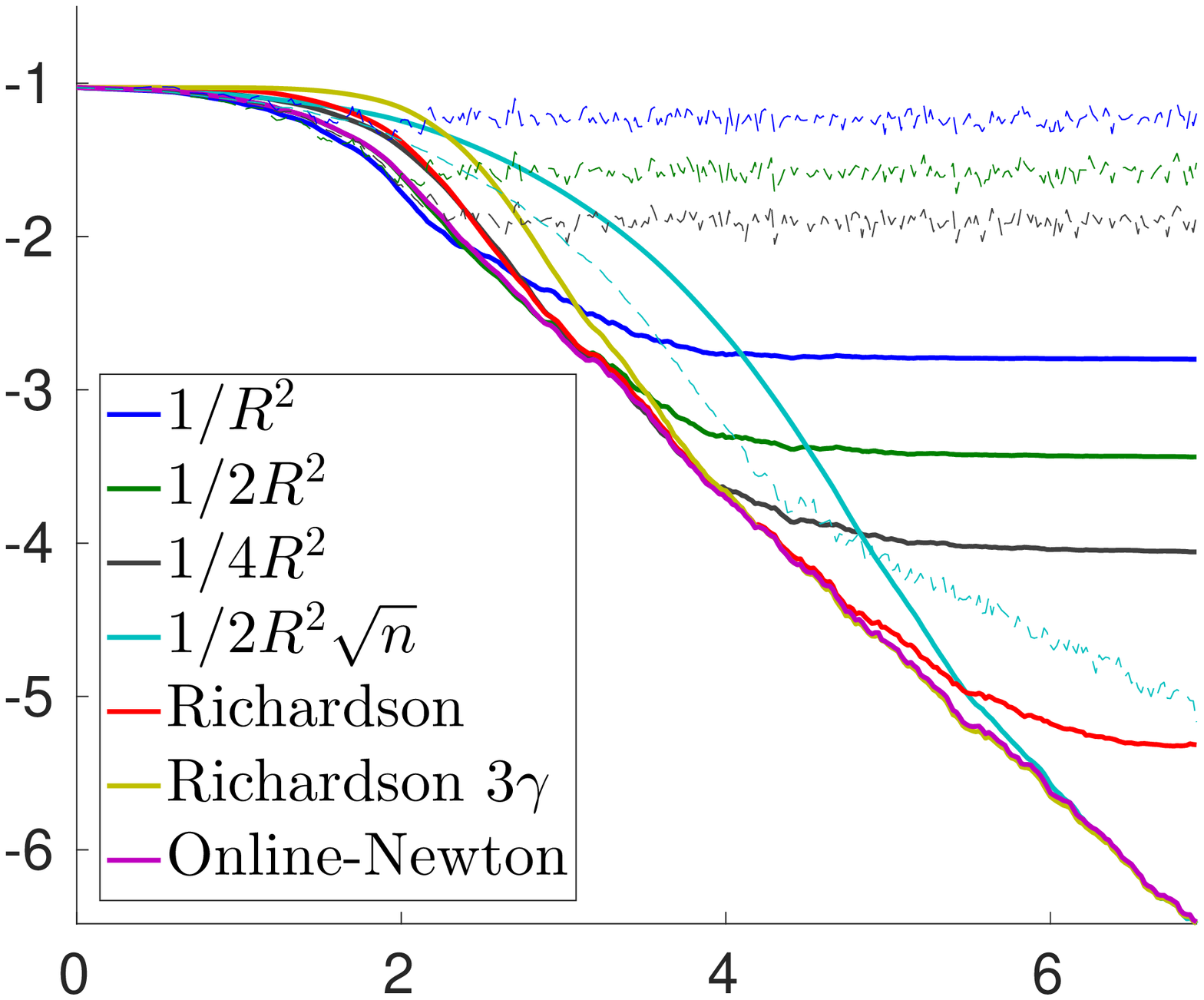}\\
          & $\log_{10}(n)$ \\
	\end{tabular}
\end{center}
			
					
	\caption{Synthetic data, logarithmic scales. Upper-left: logistic regression, $d=12$, with averaged SGD with step-size $1/R^2$, $1/2R^2$, decaying step-sizes ($\gamma_k=  1/(2R^2 \sqrt{k})$)  (averaged (plain) and non-averaged (dashed)), Richardson Romberg extrapolated iterates, and online Newton iterates. Upper-right: same in lower dimension ($d=4$). Bottom: same but with three different step-sizes and an estimator built using the Richardson estimator $\tilde{\theta_k^{3}} =\frac{8}{3}\bte{k}{ \gamma}-2\bte{k}{ 2\gamma}+\frac{1}{3}\bte{k}{ 4\gamma}$, with  3 different step-sizes $3 \gamma$, $2\gamma$ and $\gamma=1/4R^2$.}
	\label{fig:experiments}
\end{figure}

We performed experiments on simulated data, for logistic regression, with $n=10^7$ observations, for $d=12$ and $4$. Results are presented in \Cref{fig:experiments}. The data are \as~bounded by $R \geq 0$, therefore $R^2= L$. We consider SGD with  constant step-sizes  $1/R^2$, $1/2R^2$ (and $1/4R^2$) with or without averaging, with $R^{2}=L$. Without averaging, the chain saturates with an error proportional to $\gamma$ (since $\normLigne{\te{k}{\gamma} -\ts  }= O(\sqrt{\gamma})$ as $k \to \plusinfty$). Note that the ratio between the convergence limits of the two sequences is roughly $2$ in the un-averaged case, and 4 in the averaged case, which confirms the predicted limits. We consider Richardson Romberg iterates, which saturate at a much lower level, and performs much better than decaying step-sizes (as $ {1}/{\sqrt{n}}$) on the first iterations, as it forgets the initial conditions faster. Finally, we run  the online-Newton~\cite{Bac_Mou_2013}, which performs very well but has no convergence guarantee.  On the Right plot, we also propose an estimator that uses 3 different step-sizes to perform a higher order interpolation. More precisely, for all $k \in \nsets$, we compute $\tilde{\theta_k^{3}}=\frac{8}{3}\bte{k}{ \gamma}-2\bte{k}{ 2\gamma}+\frac{1}{3}\bte{k}{ 4\gamma}$. With such an estimator, the \emph{first 2} terms in the expansion, scaling as $\gamma$ and $\gamma^{2}$, should vanish, which explains that it does not saturate.


\section{Conclusion}
In this paper, we have used and developed Markov chain tools to
analyze the behavior of constant step-size SGD, with a complete
analysis of its convergence, outlining the effect of initial
conditions, noise and step-sizes. For machine learning problems, this
allows us to extend known results from least-squares to all loss
functions. This analysis leads naturally to using Romberg-Richardson
extrapolation, that provably improves the convergence behavior of the
averaged SGD iterates. Our work opens up several avenues for future
work: (a) show that Richardson-Romberg trick can be applied to the
decreasing step-sizes setting, (b) study the extension of our results under self-concordance condition \cite{Bac_2014}.

\section{Postponed proofs}
 \subsection{Discussion on assumptions on the noise}
 \label{app:noisediscussion}
 

Assumption~\Cref{ass:lip_noisy_gradient_AS}, made in the text, can be weakened in order to apply to settings where input observations are un-bounded (typically, Gaussian inputs would  not satisfy Assumption~\Cref{ass:lip_noisy_gradient_AS}). Especially, in many cases, we only need Assumption~\Cref{ass:lip_noisy_gradient_quadratic_mean} below. Let $p \geq 2$.
\begin{assumption}[p]\label{ass:lip_noisy_gradient_quadratic_mean}
\begin{enumerate}[label=(\roman*)]
\item There exists $\ttau_p \geq 0$ such that 
  $\defEnsLigne{\PE^{1/p}[\norm{\epsilon_1(\ts)}^p]} \leq \ttau_p \eqsp.$
\item For all $x,y \in \rset^d$, there exists $L \geq 0$ such that, for $q =2, \dots,  p$,  
  \begin{multline}
\label{eq:cocoercitity_lp}
\expe{\norm{f'_1(x)-f'_1(y)}^{q}} \\\leq L^{q-1} \norm{x-y}^{q-2}\ps{ x-y}{ f'(x) - f'(y)} \eqsp,
\end{multline}
where $L$ is the same constant appearing in \Cref{hyp:regularity} and $f'_1$ is defined by \eqref{eq:def_f_prime_k}.
 \end{enumerate}
\end{assumption}
On the other hand, we consider also the stronger assumption that the noise is independent of $\theta$ (referred to as the ``semi-stochastic'' setting, see \cite{Die_Fla_Bac_2016}), or more generally that the noise has a  uniformly bounded fourth order moment. 
\begin{assumption}\label{ass:semistochastic}
There exists $\tau \geq 0$ such that 
  $\sup_{\theta \in \rset^d} \defEnsLigne{\PE^{1/4}[\norm{\epsilon_1(\theta)}^4]} \leq \tau$.
\end{assumption}

Assumption~\Cref{ass:lip_noisy_gradient_quadratic_mean}($p$), $p \geq 2$, is the weakest, as it is satisfied for random design least mean squares and logistic regression with bounded fourth moment of the inputs.
Note that we do not assume that gradient or gradient estimates are \as ~bounded, to avoid the need for a constraint on the space where iterates live. It is straightforward to see that \Cref{ass:lip_noisy_gradient_quadratic_mean}($p$), $p \geq 2$, implies \Cref{ass:lip_noisy_gradient_AS}($p$) with $\tau_p=\ttau_p$, and \Cref{ass:semistochastic}-\Cref{hyp:regularity} implies \Cref{ass:lip_noisy_gradient_AS}($4$). 

It is important to note that assuming \Cref{ass:def_noise} --especially that  $ (\epsilon_{k})_{ k \in \nset^{\star} } $ are \iid~random  fields--  \emph{does not} imply \Cref{ass:semistochastic}. On the contrary, making \emph{the semi stochastic assumption}, \ie~that the noise functions $ (\epsilon_{k}(\theta_{k-1}))_{ k \in \nset^{\star} }  $ are \iid~vectors (\eg~satisfied if  $ \epsilon_{k}  $ is constant as a function of $ \theta $), is a very strong assumption, and implies \Cref{ass:semistochastic}.

\subsection{Preliminary results}
\label{sec:exist-poiss-solut}
We preface the proofs of the main results by some technical lemmas. 

\begin{lemma}
  \label{lem:poisson_exist}
  Assume  \Cref{hyp:strong_convex}-\Cref{hyp:regularity}-\Cref{ass:def_noise}-\Cref{ass:lip_noisy_gradient_AS}(2). Let
  $\phi: \rset^d \to \rset$ be a $ L_\phi $-Lipschitz function. For any step-size
  $\gamma \in\oointLigne{0, 2/L}$, the function  $\psi_{\gamma} : \rset^d \to \rset$ defined for all $\theta \in \rset^d$ by
  \begin{equation}
    \label{eq:def_poisson_sol_lip}
    \psi_{\gamma}(\theta) = \sum_{i=0}^{\plusinfty} R_\gamma^i\phi(\theta) \eqsp,
  \end{equation}
  is well-defined, Lipschitz and satisfies   $  (\Id-R_{\gamma}) \psi_{\gamma} = \phi $, $ \pi_{\gamma}(\psi_{\gamma}) = 0$.
  In addition, if $\tilde{\psi}_{\gamma}: \rset^d \to \rset$ is an other Lipchitz function satisfying  $  (\Id-R_{\gamma}) \tpsi_{\gamma} = \phi $, $ \pi_{\gamma}(\tpsi_{\gamma}) = 0$, then $\psi_{\gamma} = \tpsi_\gamma$.
\end{lemma}
\begin{proof}
  Let   $\gamma \in\oointLigne{0, 2/L}$. By \Cref{prop:existence_pi}-\ref{prop:existence_pi_item_2}, for any Lipschitz continuous function $\phi$,  $\{ \theta \mapsto \sum_{i=1}^{k} ( \Rg^{i} \phi (\theta) -\pi_\gamma(\phi)) \}_{k \geq 0}$ converges absolutely on all compact sets of $\rset^d$. 
 Therefore $\psi_{\gamma}$ given by \eqref{eq:def_poisson_sol_lip} is well-defined.
Let $(\theta,\vartheta)\in \rset^{d}\times \rset^{d}$.
Consider now the two processes $(\thargDD{k}{(1)})_{\geq 0}$,$(\thargDD{k}{(2)})_{k \geq 0}$ defined by \eqref{eq:def_coupling} with $\lambda_1= \updelta_\theta$ and $\lambda_2=\updelta_\vartheta$. 
Then, for any $k\in \nset^{*}$, using \eqref{eq:coupling_2}:
\begin{eqnarray}
\abs{\Rg^{k} \phi (\theta) - \Rg^{k} \phi (\vartheta)}  &\le& L_\phi\PE^{1/2}\parentheseDeux{\norm[2]{\thargD{k}{(1)}-\thargD{k}{(2)}}} \nonumber \\
&{\le}& L_\phi   (1 - 2\mu  \gamma (1-  \gamma L/2))^{k/2}  \|\theta-\vartheta\| \eqsp. \label{eq:Rg_lip}
\end{eqnarray}
Therefore by definition~\eqref{eq:def_poisson_sol_lip}, $\psi_{\gamma}$ is  Lipschitz continuous. Finally, it is straightforward to verify that $\psi_{\gamma}$ satisfies the stated properties.

If $\tilde{\psi}_{\gamma}: \rset^d \to \rset$ is an other Lipchitz function satisfying these properties, we have for all $\theta \in \rset^d$, $(\psig -\tpsig)(\theta) = \Rg(\psig -\tpsig)(\theta)$. Therefore for all $k \in \nsets$, $\theta \in \rset^d$, $(\psig -\tpsig)(\theta) = \Rg^k(\psig -\tpsig)(\theta)$. But by \Cref{prop:existence_pi}-\ref{prop:existence_pi_item_2}, $\lim_{k \to \plusinfty}  \Rg^k(\psig -\tpsig)(\theta)= \pg(\psig -\tpsig)=0$, which concludes the proof.
\end{proof}




\begin{lemma}
  \label{lem:gradthetagamma}
  Assume  \Cref{hyp:strong_convex}-\Cref{hyp:regularity}-\Cref{ass:def_noise}-\Cref{ass:lip_noisy_gradient_AS}(2). Then we have for any  $\gamma \in\oointLigne{0, 2/L}$.
  \begin{equation*}
     \label{eq:gradthetagamma}    
\intrd f'(\theta) \pigrmd  =0\eqsp.
  \end{equation*}
\end{lemma}
\begin{proof}
  Let $(\tharg{k})_{k \in \nset}$ be a Markov chain satisfying \eqref{eq:def_grad_sto}, with $\tharg{0}$ distributed according to $\pg$. Then the proof follows from taking the expectation in \eqref{eq:def_grad_sto} for $k=0$, using that the distribution of $\tharg{1}$ is $\pg$, $\expeLigne{\varepsilon_1(\theta)} =0$ for all $\theta \in \rset^d$ and $\epsilon_1$ is independent of $\tharg{0}$.
\end{proof}

\begin{lemma}
  \label{lem:majimpA7}
    Assume  \Cref{hyp:strong_convex}-\Cref{hyp:regularity}-\Cref{ass:def_noise}-\Cref{ass:lip_noisy_gradient_quadratic_mean}(2). Then for any initial condition $\tharg{0} \in \rset^d$, we have for any  $\gamma >0$,
    \begin{equation*}
\CPE{\norm[2]{\tharg{k+1}-\ts}}{\mcf_k} \leq  (1-2 \gamma \mu (1 -  \gamma L)) \norm{\te{k}{\gamma} -\ts}^{2} +2\gamma^{2}\ttau^{2}_2 \eqsp,
  \end{equation*}
  where  $(\te{k}{\gamma})_{k \geq 0}$ is given by \eqref{eq:def_grad_sto}.
  Moreover, if $\gamma \in \ooint{0,1/L}$, we have
  \begin{equation}
    \label{eq:moment_2_pi_gamma}
    \int_{\rset^d} \norm[2]{\theta-\ts}\pg(\rmd \theta) \leq \gamma\ttau^{2}_2/(\mu(1-\gamma L)) \eqsp.
  \end{equation}
\end{lemma}
\begin{proof}
The proof and result is very close to the ones from \cite{Nee_War_Sre_2014} but we  extend it without \as~Lipschitzness (\Cref{ass:lip_noisy_gradient_AS}) but with \Cref{ass:lip_noisy_gradient_quadratic_mean}.
Using \Cref{ass:def_noise}-\Cref{hyp:strong_convex} and $f'(\ts)=0$, we have
  \begin{align}
\nonumber
         \CPE{\norm[2]{\tharg{k+1}-\ts}}{\mcf_k} & \leq  \norm{\te{k}{\gamma} -\ts}^{2} +     \gamma^2 \CPE{\norm{f'_{k+1}(\te{k}{\gamma})}^{2}}{\mcf_k}  \\
    \label{eq:majimpA7_eq_0}
        &  - 2 \gamma \CPE{\ps{f'_{k+1}(\te{k}{\gamma})-f'_{k+1}(\ts) }{ \te{k}{\gamma} -\ts}}{\mcf_k}\\
& 
            \leq  (1-2\mu \gamma )\norm{\te{k}{\gamma} -\ts}^{2}+  \gamma^2 \CPE{\norm{f'_{k+1}(\te{k}{\gamma})}^{2}}{\mcf_k}  \eqsp.   \label{eq:majimpA7_eq_1}
  \end{align}

In addition, under \Cref{ass:def_noise}-\Cref{ass:lip_noisy_gradient_quadratic_mean}($2$) and using \eqref{eq:def_f_prime_k}, we have:
 \begin{align*}
& \CPE{\norm{f'_{k+1}(\te{k}{\gamma})}^{2}}{\mcf_k}  \\
& \qquad \qquad \le 2\left( \CPE{\norm{f'_{k+1}(\te{k}{\gamma})-f'_{k+1}(\ts)}^{2}}{\mcf_k} + \CPE{\norm{f'_{k+1}(\ts)}^{2}}{\mcf_k} \right)\\
& \qquad \qquad \le 2\left( \CPE{\norm{f'_{k+1}(\te{k}{\gamma})-f'_{k+1}(\ts)}^{2}}{\f{k}} +\tau^{2} \right)\\
& \qquad \qquad \le 2\left( L \CPE{\Big\langle f'_{k+1}(\te{k}{\gamma})-f'_{k+1}(\ts) , \te{k}{\gamma} -\ts\Big\rangle }{\f{k}} +\tau^{2} \right)\\
& \qquad \qquad \le 2\left( L {\Big\langle f'(\te{k}{\gamma})-f'(\ts) , \te{k}{\gamma} -\ts\Big\rangle} +\tau^{2} \right) \eqsp.
\end{align*}
Combining this result and \eqref{eq:majimpA7_eq_1} concludes the proof of the first inequality.

Regarding the second bound, let a fixed initial point $\tharg{0} \in \rset^d$. By Jensen inequality and the first result
we get for any $k \in \nset$ and $M \geq 0$,
\begin{multline*}
  \expe{\norm[2]{\tharg{k+1}-\ts} \wedge M} \leq (1-2 \gamma \mu (1 -  \gamma L))^{k+1} \norm{\tharg{0}-\ts}^2 \\+ 2\gamma^{2}\ttau^{2}_2 \sum_{i=0}^{k} (1-2 \gamma \mu (1 -  \gamma L))^i \eqsp.
\end{multline*}
Since by \Cref{prop:existence_pi}-\ref{prop:existence_pi_item_2}, $\lim_{k \to \plusinfty}   \expeLigne{\normLigne[2]{\tharg{k+1}-\ts} \wedge M} = \int_{\rset^d} \{\normLigne[2]{\theta-\ts} \wedge M \}\pg(\rmd \theta)$, we get for any $M \geq 0$, 
\begin{equation*}
  \int_{\rset^d} \{\normLigne[2]{\theta-\ts} \wedge M \} \pg(\rmd \theta) \leq \gamma\ttau^{2}_2/(\mu(1-\gamma L)) \eqsp.
\end{equation*}
Taking $M \to \plusinfty$ and applying the monotone convergence theorem concludes the proof. 
%
\end{proof}

Using \Cref{lem:majimpA7}, we can extend \Cref{lem:poisson_exist} to functions $\phi$ which are locally Lipschitz.
\begin{lemma}
  \label{lem:poisson_exist_2}
  Assume  \Cref{hyp:strong_convex}-\Cref{hyp:regularity}-\Cref{ass:def_noise}-\Cref{ass:lip_noisy_gradient_AS}(4). Let
  $\phi: \rset^d \to \rset$ be a function satisfying there exists
  $L_{\phi} \geq 0$ such that for any $x,y \in \rset^d$,
  \begin{equation}
    \label{eq:loc_lip_varphi}
\abs{    \phi(x)-\phi(y)} \leq L_{\phi}\norm{x-y}\defEns{1+\norm{x} + \norm{y}} \eqsp.
  \end{equation}
  For any step-size
  $\gamma \in\oointLigne{0, 1/L}$, it holds:
  \begin{enumerate}[label=(\alph*)]
  \item   \label{lem:poisson_exist_2_a} there exists $C \geq 0$ such that for all
  $\theta \in \R^{d}$, $k\in \nsets$:
  \begin{equation*}
  \abs{\Rg^{k} \phi(\theta)-\pg(\phi)} \leq   C L_\phi (1 - 2\mu  \gamma (1-  \gamma L))^{k/2}\defEns{1+\norm[2]{\theta-\ts}} \eqsp;
  \end{equation*}
  \item   the function
  $\psi_{\gamma} : \rset^d \to \rset$ defined for all
  $\theta \in \rset^d$ by     \eqref{eq:def_poisson_sol_lip}
  is well-defined satisfies   $  (\Id-R_{\gamma}) \psi_{\gamma} = \phi $, $ \pi_{\gamma}(\psi_{\gamma}) = 0$ and there exists $L_{\psi} \geq 0$ such that
  such that for any $x,y \in \rset^d$,
  \begin{equation}
\label{eq:poisson_loc_lip_2}
    \abs{    \psi(x)-\psi(y)} \leq L_{\psi}\norm{x-y}\defEns{1+\norm{x} + \norm{y}} \eqsp.
  \end{equation}
  \end{enumerate}
\end{lemma}
\begin{proof}
  In this proof, $C \geq 0$ is a constant which can change from line to line.
  \begin{enumerate}[wide, labelwidth=!, labelindent=0pt,label=(\alph*)]
  \item Let $\gamma \in \ooint{0,1/L}$. Consider the two processes $(\thargDD{k}{(1)})_{\geq 0}$,$(\thargDD{k}{(2)})_{k \geq 0}$ defined by \eqref{eq:def_coupling} with $\lambda_1= \updelta_\theta$ and $\lambda_2= \pi_{\gamma}$. Using \eqref{eq:loc_lip_varphi}, the Cauchy-Schwarz inequality, $\pg \Rg = \pg$ and \eqref{eq:coupling_2} we have for  any $k\in \nset^{*}$:
    \begin{align}
      \nonumber
      \abs{\Rg^{k} \phi(\theta)-\pg(\phi)}^2 & \leq \abs{ \expe{{\phi(\theta^{(1)}_k) - \phi(\theta^{(2)}_k)}}}^2 \\
      &\leq   L_\phi^{2} \expe{\norm[2]{\theta^{(1)}_k - \theta^{(2)}_k}} \expe{1+\norm[2]{\theta^{(1)}_k} + \norm[2]{\theta^{(2)}_k}} \nonumber \\
      \nonumber
                                             &\leq C  L_\phi^{2} (1 - 2\mu  \gamma (1-  \gamma L/2))^{k} \int \|\theta-\vartheta\|^{2}\rmd\pi_\gamma (\vartheta) \\
\nonumber
  & \qquad \qquad \times \parenthese{1+(1 - 2\mu  \gamma (1-  \gamma L))^{k}\norm[2]{\theta-\ts}} \eqsp, 
\end{align}
where we have \Cref{lem:majimpA7} for the last inequality. Then the proof is concluded using for all $x,y \in \rset^d$, $\norm[2]{x+y} \leq 2(\norm[2]{x} + \norm[2]{y})$ and \Cref{lem:majimpA7} again.
\item   Let   $\gamma \in\oointLigne{0, 1/L}$. By \ref{lem:poisson_exist_2_a}, $\{ \theta \mapsto \sum_{i=1}^{k} ( \Rg^{i} \phi (\theta) -\pi_\gamma(\phi)) \}_{k \geq 0}$ converges absolutely on all compact sets of $\rset^d$. 
Therefore $\psi_{\gamma}$ given by \eqref{eq:def_poisson_sol_lip} is well-defined.
Let $(\theta,\vartheta)\in \rset^{d}\times \rset^{d}$.
Consider now the two processes $(\thargD{k}{(1)})_{\geq 0}$,$(\thargD{k}{(2)})_{k \geq 0}$ defined by \eqref{eq:def_coupling} with $\lambda_1= \updelta_\theta$ and $\lambda_2=\updelta_\vartheta$. 
Then \eqref{eq:loc_lip_varphi}, the Cauchy-Schwarz inequality and \eqref{eq:coupling_2}, for any $k\in \nset^{*}$, we get:
\begin{align*}
\abs{\Rg^{k} \phi (\theta) - \Rg^{k} \phi (\vartheta)}^2  &\leq \abs{ \expe{{\phi(\theta^{(1)}_k) - \phi(\theta^{(2)}_k)}}}^2 \nonumber \\
& \leq C L_\phi^{2}   (1 - 2\mu  \gamma (1-  \gamma L))^{k/2}  \|\theta-\vartheta\| \defEns{1 + \norm[2]{\theta} + \norm[2]{\vartheta}} \eqsp.
\end{align*}
By definition \eqref{eq:def_poisson_sol_lip}, $\psi_{\gamma}$ satisfies \eqref{eq:poisson_loc_lip_2}. Finally, it is straightforward to verify that $\psi_{\gamma}$ satisfies the stated properties.
\end{enumerate}

\end{proof}

It is worth pointing out that under
Assumption~\Cref{ass:semistochastic} (the ``semi-stochastic''
assumption), a slightly different result holds. The following result
underlines the difference between a stochastic noise and a
semi-stochastic noise, especially the fact that the maximal step-size
differs depending on this assumption
made.

\begin{lemma}
    Assume  \Cref{hyp:strong_convex}-\Cref{hyp:regularity}-\Cref{ass:def_noise}-\Cref{ass:semistochastic}. Then for any initial condition $\tharg{0} \in \rset^d$, we have for any  $\gamma \in \ocint{0,2/(m+L)}$,
    \begin{equation*}
\CPE{\norm[2]{\tharg{k+1}-\ts}}{\mcf_k} \leq  (1-2 \gamma \mu L /(\mu + L)) \norm{\te{k}{\gamma} -\ts}^{2} +\gamma^{2}\tau^{2} \eqsp,
  \end{equation*}
where  $(\te{k}{\gamma})_{k \geq 0}$ is given by \eqref{eq:def_grad_sto}.
\end{lemma}

\begin{proof}
  First, note that since $f$ satisfies \Cref{hyp:strong_convex} and \Cref{hyp:regularity}, by \cite[Chapter 2,  (2.1.24)]{Nes_2004}, for all $x,y \in \rset^d$,
          \begin{equation}
          \label{eq:strongconvopt}
\Big\langle f'(x) - f'(y), x-y\Big\rangle \geq \frac{L \mu}{L+\mu} \|x-y\|^{2} + \frac{1}{L+\mu} \|f'(x) -f'(y)\|^{2}  \eqsp.
\end{equation}

Besides,  under \Cref{ass:semistochastic}, we have:
  \begin{eqnarray*}
		\expe{\norm{f'_{k+1}(\te{k}{\gamma})}^{2} |\f{k}}  &=&  {\norm{f'(\te{k}{\gamma})}^{2}}+\expe{\norm{f'_{k+1}(\te{k}{\gamma})-f'(\te{k}{\gamma})}^{2}}  \\
		&\le &  {\norm{f'(\te{k}{\gamma})}^{2}} + \tau^{2} \eqsp.
	\end{eqnarray*}
	So that finally, using \eqref{eq:majimpA7_eq_0}, \Cref{ass:def_noise}, \eqref{eq:strongconvopt}, \Cref{hyp:regularity} and rearranging terms we get
	\begin{align*}
   \CPE{\norm[2]{\tharg{k+1}-\ts}}{\mcf_k}
          &\le   (1-2 \gamma \mu L /(\mu + L)) \norm{\te{k}{\gamma} -\ts}^{2} + \gamma^{2 }\tau^{2}\\& \qquad  - 2 \frac{\gamma}{L+\mu} \norm{ f'(\te{k}{\gamma}) }^2 + \gamma^{2}  {\norm{f'(\te{k}{\gamma})}^{2}}  \eqsp.
	\end{align*}
        Using that $\gamma \leq 2/(m+L)$ concludes the proof.
\end{proof}

We give uniform bound on the moments of the chain $(\theta_{k}^{(\gamma)})_{k \geq 0}$ for $\gamma >0$.
For $p \geq 1$, recall that under \Cref{ass:lip_noisy_gradient_AS}(2p), the noise at optimal point has a moment of order $ 2p $ and we denote
\begin{equation}
\label{eq:definition_momentNoise}
\momentNoise_{2p} = \expeExpo{1/2p}{\norm[{2p}]{\epsilon_1(\ts)}} \eqsp.
\end{equation}
We give a bound on the $p$-order moment of the chain, under the assumption that the noise has a moment of order $ 2p $.

For moment of order larger than $2$, we have the following result.
\begin{lemma}
  \label{lem:moment_p}
  Assume \Cref{hyp:strong_convex}-\Cref{hyp:regularity}-\Cref{ass:def_noise}-\Cref{ass:lip_noisy_gradient_AS}(2p), for $ p\geq 1$.
	There exist numerical constants $C_p, D_p \geq 2$ that only depend on $p$, such that, if $\gamma \in \ooint{0,1/(LC_p)}$, for all $k \in \nsets$ and $\theta_0 \in \rset^d$
	\begin{equation*}
	\expeMarkovExpo{}{1/p}{\norm[2p]{\tharg{k}-\ts}} \leq (1-2 \gamma \mu(1-C_p \gamma L/2 ))^k  \expeMarkovExpo{}{1/p}{\norm[2p]{\theta_0-\ts}} +  \frac{D_p \gamma \momentNoise_{2p}^2}{\mu} \eqsp,
      \end{equation*}
      where $(\tharg{k})_{k \in \nset}$ is defined by \eqref{eq:def_grad_sto} with initial condition $\tharg{0}= \theta_0$.
	Moreover, the following bound holds
	\begin{equation}
          \label{eq:momentp_statio}
          \intrdargD{\norm[2p]{\theta-\ts}} \leq \parenthese{D_p \gamma \momentNoise_{2p}^2/\mu}^{p} \eqsp. 
	\end{equation}
	
\end{lemma}

\begin{remark}
\begin{itemize}[leftmargin=*]
\item Notably, \Cref{lem:moment_p} implies  that $ \intrdargD{\norm[4]{\theta-\ts} }=O(\gamma^{2}) $, and thus $   \intrdargD{\norm[3]{\theta-\ts} }=O(\gamma^{3/2}) $.
	We also note that  $  \intrdargD{ \| \theta - \ts\|^2} = O(\gamma)$,  also implies by Jensen's inequality that $\| \tav - \ts\|^2 = O(\gamma)$.
	\item 	Note that there is no contradiction between \eqref{eq:momentp_statio} and \Cref{THEO:BIAS1}, as for any $ p\geq 2 $, one has for $ g(\theta)=\norm [2]{\theta-\ts}  $ and $ h_g $ the solution to the Poisson equation, that $ h''_g(\ts) =0 $, so that the first term in the development (of order $ \gamma $) is indeed 0.
\end{itemize}
\end{remark}

\begin{proof}
Let $\gamma \in \oointLigne{0,(1/2L)}$.  Set for any $k \in \nsets$, $\delta_k = \normLigne{\tharg{k}-\ts}$. The proof is by induction on $p \in \nsets$.
For conciseness, in the rest of the proof, we skip the explicit dependence in $ \gamma $ in $ \te{i}{\gamma} $: we only denote it $ \theta_i $. 
  For $p=2$, the result holds by \Cref{lem:majimpA7}. Assume that the result holds for $p-1$, $p \in \nsets$, $p \geq 2$. By definition, we have
\begin{eqnarray}
\delta_{k+1}^{2p}&=&  \left(\delta_k^{2}  -2 \gamma  \langle f_{k+1}'(\theta_k), \theta_{k} -\ts\rangle+\gamma^{2 }  \| f_{k+1}'(\theta_k)\|^{2} \right)^{p} \nonumber\\
&=&  
\sum_{\substack{i,j, l \in \zerop^{3}\\ {i+j+l=p}}} \frac{p!}{i! j! l!} 
\delta_k^{2i }\ \  (2 \gamma)^{j}  \langle f_{k+1}'(\theta_k), \theta_{k} -\ts\rangle^{j} \ \ 
\gamma^{2l }  \| f_{k+1}'(\theta_k)\|^{2l} \eqsp . \label{eq:devensomme}
\end{eqnarray}
We upper bounds each term for $i,j,l \in \zerop$, as follows:
\begin{enumerate}[wide, labelwidth=!, labelindent=0pt]
	\item For $ i=p, j=l=0 $, we have $  \delta_k^{2p }$.
	\item For $ i=p-1, j=1, l=0 $, we have
          $p 2 \gamma \langle f_{k+1}'(\theta_k), \theta_{k} -\ts\rangle
          \delta_k^{2(p-1) }$, for which it holds by
          \Cref{ass:def_noise}
          \begin{equation}
            \label{eq:moment_p_proof_1}
           \CPE{p 2 \gamma \langle f_{k+1}'(\theta_k), \theta_{k} -\ts\rangle
          \delta_k^{2(p-1) }}{\mcf_k} =  p 2 \gamma \langle f'(\theta_k), \theta_{k}
          -\ts\rangle\delta_k^{2(p-1) } \eqsp.
          \end{equation}
	\item Else, either $ l\geq 1 $ or $ j\geq 2 $, thus $ 2l+j \geq 2 $.  We first upper bound, by the  Cauchy–Schwarz inequality: 
          \begin{equation}
            \label{eq:moment_p_proof_1_5}
            \E[   \langle f_{k+1}'(\theta_k), \theta_{k} -\ts\rangle^{j}  |\f{k}] \le \delta_k^{j} \norm[j]{f_{k+1}'(\theta_k)}\eqsp .
	\end{equation}
	Second, we have
	\begin{align}
          \nonumber
          \E[ \| f_{k+1}'(\theta_k)\|^{2l+j} |\f{k}] & \le 2^{2l+j-1} \bigg( \E[ \| f_{k+1}'(\theta_k) - f_{k+1}'(\ts) \|^{2l+j} |\f{k}] \\
                    \nonumber
                                                     & \qquad\qquad\qquad\qquad\qquad\quad+\E[ \| f_{k+1}'(\ts) \|^{2l+j} |\f{k}] \bigg) \\
\label{eq:moment_p_proof_2}
	&\le 2^{2l+j-1} \bigg( \E[ \| f_{k+1}'(\theta_k) - f_{k+1}'(\ts) \|^{2l+j} |\f{k}] + \tau_{2p}^{2l+j} \bigg)\eqsp ,
	\end{align}
	using for any $x,y \in \rset$,
        $(x+y)^{2l+j}\le 2^{2l+j-1} (x^{2l+j}+ y^{2l+j})$,
        \Cref{ass:lip_noisy_gradient_AS}($2p$), $ 2l+j\le 2p $ and
        Hölder inequality. In addition, using
        \Cref{ass:lip_noisy_gradient_AS}(2p), we get
	\begin{align*}   \E[ \| f_{k+1}'(\theta_k) - f_{k+1}'(\ts) \|^{2l+j} |\f{k}] &\le  L ^{2l+j-2}\delta_k^{2l+j-2} \E[ \| f_{k+1}'(\theta_k) - f_{k+1}'(\ts) \|^{2} |\f{k}] \\
&	\le  L ^{2l+j-1}\delta_k^{2l+j-2} \langle f'(\theta_k) - f'(\ts) , \theta_k- \ts\rangle \eqsp .
	\end{align*}
\end{enumerate}
Combining this result, \eqref{eq:moment_p_proof_1_5} and \eqref{eq:moment_p_proof_2} implies using $i+j+l \leq p$,
\begin{align}
  \nonumber
& \E[ \delta_k^{2i }\ \  (2 \gamma)^{j}  \langle f_{k+1}'(\theta_k), \theta_{k} -\ts\rangle^{j} \ \ 
\gamma^{2l }  \| f_{k+1}'(\theta_k)\|^{2l} |\f{k}] \\
  \nonumber
&\le  \delta_k^{2i +j} \  2^{2j+2l-1}    \gamma^{2l+j}   \bigg( \E[ \| f_{k+1}'(\theta_k) - f_{k+1}'(\ts) \|^{2l+j} |\f{k}] + \tau_{2p}^{2l+j} \bigg)\\
    \nonumber
&\le \gamma^{2l+j}  2^{2l+2j-1}  \delta_k^{2i + 2j+2l-2}   L ^{2l+j-1} \langle f'(\theta_k) - f'(\ts) , \theta_k- \ts \rangle \\
  \label{eq:moment_p_proof_3}
& \qquad\qquad\qquad\qquad\qquad\qquad
   +  \gamma^{2l+j}  2^{2l+2j-1}  \delta_k^{2i + j} \tau_{2p}^{2l+j} \eqsp.
\end{align}
Define then
\begin{equation*}
C_p = \max\parentheseDeux{2, (1/p) \sum_{\substack{i,j, l \in \zerop^{3}\\ {i+j+l=p}\\ j+2l\geq 2}} \frac{p!}{i! j! l!} 2^{2l+2j-1}} \eqsp.
\end{equation*}
Note that using $j+2l \geq 2$,  for $\gamma$ such that $\gamma L < 1/C_p$, it holds
\begin{equation}
  \label{eq:prop_cp}
  \frac{1}{p} \sum_{\substack{i,j, l \in \zerop^{3}\\ {i+j+l=p}\\ j+2l\geq 2}} \frac{p!}{i! j! l!} (\gamma L)^{2l+j-1}  2^{2l+2j-1} \leq  \gamma L C_p < 1 \eqsp .
\end{equation}
Therefore, we have combining this inequality, \eqref{eq:moment_p_proof_1}-\eqref{eq:moment_p_proof_3} in \eqref{eq:devensomme},
\begin{align*}
\E[	\delta_{k+1}^{2p}|\f{k}]&\le  \delta_k^{2p} - 2 \gamma p (1-\gamma L C_p/2) \delta_k^{2(p-1)}    \langle f'(\theta_k) - f'(\ts) , \theta_k- \ts \rangle \\& +  \sum_{\substack{i,j, l \in \zerop^{3}\\ {i+j+l=p}\\ j+2l\geq 2}} \frac{p!}{i! j! l!}   \gamma^{2l+j}  2^{2l+2j-1}  \delta_k^{2i + j} \tau_{2p}^{2l+j} \eqsp .
\end{align*}
Using \Cref{hyp:strong_convex}, for $j \in \zerop$,  $ (\gamma \tau_{2p} \delta_k)^{j} \le 2 (\gamma \tau_{2p})^{2j}+ 2 (\delta_k )^{2j}$, we get 
\begin{align*}
\E[	\delta_{k+1}^{2p}] 
                           &\le  (1-2\gamma \mu p (1-\gamma L C_p/2)) \E[\delta_k^{2p} ] \\
                           & +  \sum_{\substack{i,j, l \in \zerop^{3}\\ {i+j+l=p}\\ j+2l\geq 2}} \frac{p!}{i! j! l!}    4^{l+j}  (\gamma^{2} \tau_{2p}^{2})^{l+j} \E[\delta_k^{2i} ] + \frac{p!}{i! j! l!}    4^{l+j}   (\gamma^{2} \tau_{2p}^{2})^{l} \E[\delta_k^{2i +2 j} ]\eqsp .
\end{align*}
Finally, denoting $ c_k=\E^{1/p}[\delta_k^{2p}] $, using that by
Hölder inequality  $\PE[\delta_k^{2i}] \leq c_k^i$, for all $i \in \zerop$, we have:
\begin{align}
  \nonumber
  c_{k+1}^{p}&\le  (1-2\gamma \mu p (1-\gamma L C_p/2))  c_{k}^{p} + \sum_{\substack{i,j, l \in \zerop^{3}\\ {i+j+l=p}\\ j+2l\geq 2}} \frac{p!}{i! j! l!}    4^{l+j}  (\gamma^{2} \tau_{2p}^{2})^{l+j}  c_{k}^{i}\\
  \label{eq:ineq_c_k_1}
& \qquad +  \sum_{\substack{i,j, l \in \zerop^{3}\\ {i+j+l=p}\\ j+2l\geq 2}} \frac{p!}{i! j! l!}    4^{l+j}   (\gamma^{2} \tau_{2p}^{2})^{l}  c_{k}^{i+j} \eqsp.
\end{align}
Define
\begin{equation*}
  D_p  = \max_{ u \in \zerop} \parentheseDeux{2^{p-u} \binom{p}{u}^{-1}\defEns{\sum_{\substack{i,j, l \in \zerop^{3}\\ {i+j+l=p}\\ j+2l\geq 2\\l+j=u}} \frac{p!}{i! j! l!}    4^{l+j}  +  \sum_{\substack{i,j, l \in \zerop^{3}\\ {i+j+l=p}\\ j+2l\geq 2\\l=u}} \frac{p!}{i! j! l!}    4^{l+j}   }}
\end{equation*}
Note that using \eqref{eq:prop_cp}, $C_p \geq 2$ and $\mu \leq L$, $   (1-2\gamma \mu (1-\gamma L C_p/2)) \geq (1-\gamma L C_p (1-\gamma L C_p/2)) \geq 1/2 $. Using this inequality and  $ 1-p t \le (1-t)^{p} $ for $ t\geq 0 $ we get by \eqref{eq:ineq_c_k_1} setting $\rho = (1-2\gamma \mu(1-\gamma L C_p/2))$,
\begin{align*}
  & \parenthese{\rho c_k + D_p \gamma^2 \tau_p}^p = \sum_{u=0}^p \binom{p}{u} (\rho c_k)^{p-u} (D_p \gamma^2 \tau_p)^{u}\\
  & \geq  (1-2\gamma \mu p (1-\gamma L C_p/2))  c_{k}^{p} + \sum_{u=0}^p2^{u-p} c_k^{p-u}  \binom{p}{u} (\rho c_k)^{p-u} (D_p \gamma^2 \tau_p)^{u} \geq
c_{k+1}^p \eqsp.    
\end{align*}
A straightforward induction implies the first statement. The proof of \eqref{eq:momentp_statio} is similar to the one of \eqref{eq:moment_2_pi_gamma} and is omitted.
\end{proof}
\begin{lemma}
\label{lem:lip_g_0}
  Let $g: \rset^d \to \rset$ satisfying
  \Cref{assum:assum_f_test}($1,\pmom$) for $\pmom \in \nset$.
 Then for all $\theta_1,\theta_2 \in \rset^d$,
		\begin{equation*}
		\abs{g(\theta_1)-g(\theta_2)} \leq a_g \norm{\theta_1-\theta_2}\defEns{b_g+\norm[\pmom]{\theta_1-\ts}+\norm[\pmom]{\theta_2-\ts}} \eqsp.
		\end{equation*}
\end{lemma}
\begin{proof}
Let $\theta_1,\theta_2 \in \rset^d$.
		By the mean value theorem, there exists $s \in \ccint{0,1}$ such that if $\eta_s =s\theta_1 +(1-s) \theta_2$ then
		\begin{equation*}
		\abs{g(\theta_1)-g(\theta_2)} = Dg(\eta_s)\defEns{\theta_1-\theta_2} \eqsp.
		\end{equation*}
		The proof is then concluded using \Cref{assum:assum_f_test}($\lreg,\pmom$) and  
		\begin{equation*}
		\norm{\eta_s-\ts} \leq \max\parenthese{\norm{\theta_1-\ts},\norm{\theta_2-\ts}} \eqsp.
		\end{equation*}  
\end{proof}

\begin{proposition}
  \label{theo:convergence_loc_lip_wasserstein}
  Let $g: \rset^d \to \rset$ satisfying
  \Cref{assum:assum_f_test}($1,\pmom$) for $\pmom \in \nset$.  Assume
  \Cref{hyp:strong_convex}-\Cref{hyp:regularity}-\Cref{ass:def_noise}-\Cref{ass:lip_noisy_gradient_AS}($2
  \pmom$).  Let $C_{\pmom} \geq 2$ be given by \Cref{lem:moment_p} and  only depending on
  $\pmom$. For all $\gamma \in \oointLigne{0,1/(LC_{\pmom})}$, for all
  initial point $\theta_0 \in \rset^d$, there exists $C_{g} $ independent of $\theta_0$ such that
  for all $ k\geq 1 $:
	\begin{equation*}
	\abs{ \expeMarkov{}{k^{-1} \sum_{i=1}^k \defEns{g(\tharg{i})}} - \int_{\rset^d} g(\theta) \pi_{\gamma}(\rmd \theta)} \leq C_g(1+\norm[p]{\theta_0-\ts})/k \eqsp. 
	\end{equation*}
\end{proposition}

\begin{proof}
Let $\gamma \in \oointLigne{0,1/(LC_{\pmom})}$.  Consider the two processes
  $(\thargDD{k}{(1)})_{\geq 0}$,$(\thargDD{k}{(2)})_{k \geq 0}$ defined
  by \eqref{eq:def_coupling} with $\lambda_1= \updelta_\theta$ and
  $\lambda_2=\updelta_\vartheta$. By Jensen inequality, we have
	\begin{equation}
          \label{eq:theo:convergence_loc_lip_wasserstein_1}
          \abs{\sum_{i=1}^k \left( \expeMarkov{}{g(\thargDD{i}{(1)})} - \int_{\rset^d} g(\vartheta) \pi_{\gamma}(\rmd \vartheta)\right) }
	\leq   \sum_{i=1}^k \expe{\abs{g( \thargDD{i}{(1)}) - g(\thargDD{i}{(2)})} } \eqsp.
	\end{equation}
	Using~\Cref{lem:lip_g_0}, the Cauchy Schwarz  and Minkowski inequalities and \eqref{eq:coupling_2}  we get
	\begin{align*}
&          \expe{  \abs{g(\thargDD{i}{(1)}) - g(\thargDD{i}{(2)})}} \\
                                                                     &\qquad  \leq a_g \expeExpo{1/2}{\norm[2]{\thargDD{i}{(1)}-\thargDD{i}{(2)}}} \expeExpo{1/2}{\left (b_g + \norm[\pmom]{\thargDD{i}{(1)}-\ts}+ \norm[\pmom]{\thargDD{i}{(2)}-\ts}\right )^{2}}\\
	&\qquad  \leq  a_g \left ( \rho ^{i} \intrd \norm{\theta-\vartheta} \rmd\pg (\vartheta) \right  )^{1/2}\\
	& \qquad \qquad \qquad \qquad\times  \left( b_g  + \expeExpo{1/2}{\norm[2 \pmom]{\thargDD{i}{(1)}-\ts}}+ \expeExpo{1/2}{\norm[2\pmom]{\thargDD{i}{(2)}-\ts}}\right) \eqsp,
	\end{align*}
	with $\rho =(1-2\mu \gamma (1- \gamma L/2) )$. Moreover,
        \Cref{lem:moment_p} and Hölder inequality imply that there exists $D_{\pmom} \geq 2$ such that  for all
        $\gamma \in \ooint{0,1/(LC_{\pmom})}$ and $i \in \nset$:
        \begin{multline*}
 \expeExpo{1/2}{\norm[2\pmom]{\thargDD{i}{(1)}-\ts}} \leq
        2^{(\pmom/2-1)_+} \expeExpo{1/2}{\norm[2\pmom]{\thargDD{0}{(1)}-\ts}} \\+
        2^{(\pmom/2-1)_+} \left(\frac{D_{\pmom} \gamma
            \momentNoise_{2\pmom}^2}{\mu}\right)^{\pmom/2} \eqsp.          
        \end{multline*}
        Thus, using that for all $i \in \nset$, $\thargDD{i}{(2)}$ has for distribution $\pg$ and \Cref{lem:moment_p} again,  we obtain for all $i \in \nset$,
        \begin{equation*}
\expeMarkov{}{  \abs{g(\thargDD{i}{(1)}) - g(\thargDD{i}{(2)})}} 
		\leq   \tilde{C}_g  \rho ^{i/2}\eqsp,
	\end{equation*}
        where
        \begin{align*}
\tilde{C}_g &=          a_g  \left (\intrd \norm[2]{\theta-\vartheta} \rmd\pg (\vartheta) \right  )^{1/2}   \Bigg[ b_g  \\  
		& \qquad + 2^{(\pmom/2-1)_+} \norm[\pmom]{\thargDD{0}{(1)}-\ts}  +  (2^{(\pmom/2-1)_+}+1) \left(\frac{D_{\pmom} \gamma \momentNoise_{2\pmom}^2}{\mu}\right)^{\pmom/2} \Bigg] \eqsp.
        \end{align*}
Combining this result and \eqref{eq:theo:convergence_loc_lip_wasserstein_1} concludes the proof.

\end{proof}

\subsection{Proof of  \Cref{lem:statio_quadractic}}
\label{sec:proof-crefl_quad}
\begin{proof}[Proof of \Cref{lem:statio_quadractic}]
 By \Cref{lem:gradthetagamma}, we have
  $\int_{\rset^d} f'(\theta) \pg(\rmd \theta) = 0$. Since $f'$ is
  linear, we get $f'(\tav) = 0$, which implies by
  \Cref{hyp:strong_convex} that $\tav = \ts$.
  
 Let $\gamma \in \ooint{0,2/L}$  and $(\tharg{k})_{k \in \nset}$ given
  by \eqref{eq:def_grad_sto} with $\tharg{0}$ distributed according to
  $\pg$ independent of $(\epsilon_k)_{k \in \nsets}$. 
  Note that if $f = f_{\Sigma}$, \eqref{eq:def_grad_sto} implies for $k =1$:
  \begin{equation*}
(\te{1}{\gamma} -\ts)^{\otimes 2}= \Big( (\Id- \gamma  \Sigma ) \left( \te{0}{\gamma} - \ts \right) + \gamma \varepsilon_1(\te{0}{\gamma}) \Big)^{\otimes 2}    
\end{equation*}
Taking the expectation, using \Cref{ass:def_noise}, $\tharg{0}$ is independent of $\varespilon_1$ and $\pg \Rg = \pg$, we get
\begin{multline*}
  \intrdargD{(\theta-\ts)^{\otimes 2}} =\ (\Id- \gamma  \Sigma ) \parentheseDeux{\intrdargD{(\theta-\ts)^{\otimes 2}}}(\Id- \gamma  \Sigma ) \\
  + \gamma^{2} \int_{\rset^d} \mcC(\theta) \pg(\rmd \theta) \eqsp.
\end{multline*}
\begin{equation}
  \label{eq:rec_moment2quadratic}
(\Sigma \otimes \Id + \Id\otimes \Sigma - \gamma \Sigma \otimes \Sigma )\parentheseDeux{   \intrd (\theta-\ts)^{\otimes 2}\pigrmd } = \gamma        \intrd \mcC(\theta) \pigrmd\eqsp. 
\end{equation}

It remains to show that
$(\Sigma \otimes \Id + \Id\otimes \Sigma - \gamma \Sigma \otimes
\Sigma ) $ is invertible. To show this result, we just claim that it is a
symmetric definite positive operator. Indeed, since $\gamma <2L^{-1}$,
$\Id-(\gamma/2) \Sigma$ is symmetric positive definite and is diagonalizable
with the same orthogonal vectors $(\bff_i)_{i \in \zerod}$ as
$\Sigma$. If we denote by $(\lambda_i)_{i \in \zerod}$, then we get
that $(\Sigma \otimes \Id + \Id\otimes \Sigma - \gamma \Sigma \otimes
\Sigma )  = \Sigma \otimes (\Id-\gamma/2\Sigma) +  (\Id-\gamma/2\Sigma)\otimes \Sigma$ is also diagonalizable in the orthogonal basis of $\rset^d \otimes \rset^d$, $(\bff_i \otimes \bff_j)_{i,j \in \zerod}$  and  $(\lambda_i(1-\gamma\lambda_j)+\lambda_j(1-\gamma \lambda_i))_{i,j \in \zerod}$ are its eigenvalues.
\end{proof}

Note that in the case of the regression setting described in \Cref{ex:iid_observation_learning}, we can specify \Cref{lem:statio_quadractic} as follows.
\begin{proposition}\label{prop:quadwithT}
  Assume that $f$ is an objective function of a least-square
  regression problem, \ie~with the notations of
  \Cref{ex:iid_observation_learning}, $f = f_{\Sigma}$,
  $\Sigma = \expeLigne{X X^{\top}}$ and $\epsilon_k$ are defined by
  \eqref{eq:eps_reg_line}.  Assume
  \Cref{hyp:strong_convex}-\Cref{hyp:regularity}-\Cref{ass:def_noise}-\Cref{ass:lip_noisy_gradient_AS}(4)
   and let $\rborne$ defined by \eqref{eq:def_rborne}.
We have for all $\gamma \in \ooint{0,1/\rborne^{2}}$,
\begin{equation*}
 (\Sigma \otimes \Id + \Id\otimes \Sigma - \gamma \bfT ) \parentheseDeux{ \int_{\rset^d} (\theta-\ts)^{\otimes 2} \pigrmd } = \gamma  \E [\xi_1^{\otimes 2} ]\eqsp,
\end{equation*}
where  $\bfT$ and $\xi_1$ are defined by \eqref{eq:def_operator_T} and \eqref{eq:def_xi_k_addit_part} respectively.
\end{proposition}

\begin{proof}
The proof follows the same line as the proof of \Cref{lem:statio_quadractic} and is omitted.
\end{proof}

\subsection{Proof of \Cref{theo:statio_general}}
\label{sec:proof-crefth_statio_general}

We preface the proof by a couple of preliminaries lemmas.
\begin{lemma}
    \label{preli:theo:statio_general}
Assume
  \Cref{hyp:strong_convex}-\Cref{hyp:regularity}-\Cref{ass:def_noise}-\Cref{ass:lip_noisy_gradient_AS}($6\vee 2k_{\varepsilon}$)-\Cref{assum:reguarity_noise}
  and let $\gamma \in \ooint{0,2/L}$. Then
  \begin{equation}
      \label{eq:firstdev}
\tav - \ts = \gamma  f''(\ts)^{-1}
 f'''(\ts) 
  \bfA  \parentheseDeux{\intrdarg{\mcC(\theta)}}  + O(\gamma^{3/2})  \eqsp,
\end{equation}
where $\bfA$ is defined by \eqref{eq:def_bfA},
$\tav$ and  $\mcC$ are given by \eqref{eq:def_mean_pi_gamma} and \eqref{eq:def_cov_matrix} respectively.
\end{lemma}

\begin{proof}
   Let $\gamma \in \ooint{0,2/L}$  and $(\tharg{k})_{k \in \nset}$ given
  by \eqref{eq:def_grad_sto} with $\tharg{0}$ distributed according to
  $\pg$ independent of $(\epsilon_k)_{k \in \nsets}$. 
For conciseness, in the rest of the proof, we skip the explicit dependence in $ \gamma $ in $ \te{i}{\gamma} $: we only denote it $ \theta_i $. 

First by a third Taylor expansion with integral remainder of $f'$ around $\ts$, we have that for all $x \in \rset^d$,
\begin{equation}
   \label{eq:theo:statio_general_1_0}
f'(\theta) = f''(\ts) (\theta - \ts) + (1/2) f'''(\ts) (\theta - \ts)^{\otimes 2} + \mcr_1(\theta) \eqsp,
\end{equation}
where $\mcr_1 : \rset^d\to \rset^d$ satisfies
\begin{equation}
  \label{eq:theo:statio_general_1}
 \sup_{\theta \in \rset^d} \{\norm{\mcr_1(\theta)} / \norm{\theta-\ts}^3\} < \plusinfty  \eqsp.
\end{equation}
It follows from \Cref{lem:gradthetagamma}, taking the integral with respect to $\pg$,
\begin{equation*}
0  =  \intrdarg{  f''(\ts) (\theta - \ts) + (1/2) f'''(\ts) (\theta - \ts)^{\otimes 2} + \mcr_1(\theta)} \eqsp.  
\end{equation*}
Using \eqref{eq:theo:statio_general_1}, \Cref{lem:moment_p} and Hölder inequality, we get 
\begin{equation}
 f''(\ts)  ( \tav - \ts) +  (1/2) f'''(\ts) \parentheseDeux{ \intrdargD{  (\theta - \ts)^{\otimes 2} } } =  O(\gamma^{3/2})\eqsp.  \label{eq:dev_tav_ordre1}
\end{equation}
Moreover, we have by a second order Taylor expansion with integral remainder of $f'$ around $\ts$,
\begin{equation*}
  \teD{1} -\ts   =   \teD{0} - \ts - \gamma \big[ f''(\ts) ( \teD{0} - \ts)  + \varepsilon_1(\teD{0}) + \mcr_2(\theta_0) \big] \eqsp,
\end{equation*}
where $\mcr_2 : \rset^d \to \rset^d$ satisfies
\begin{equation}
  \label{eq:theo:statio_general_2}
   \sup_{\theta \in \rset^d} \{\norm{\mcr_2(\theta)} / \norm{\theta-\ts}^2\} < \plusinfty \eqsp.
\end{equation}
Taking the second order moment of this equation, and using \Cref{ass:def_noise}, $\theta_0$ is independent of $\epsilon_1$, \eqref{eq:theo:statio_general_2}, \Cref{lem:moment_p} and Hölder inequality, we get  
\begin{multline*}
 \intrdargD{  (\theta - \ts)^{\otimes 2} }
 = (\Id - \gamma f''(\ts) )  \parentheseDeux{ \intrdargD{  (\theta - \ts)^{\otimes 2} }}  (\Id - \gamma f''(\ts) ) \\
 + \gamma^2 \intrdargD{\mcC(\theta)}  + O(\gamma^{5/2}). 
\end{multline*}
This leads to:
\begin{equation*}
 \intrdargD{  (\theta - \ts)^{\otimes 2} }
=  \gamma \bfA \parentheseDeux{ \intrdargD{\mcC(\theta)}}
+ O(\gamma^{3/2}) \eqsp.
\end{equation*}
Combining   this result and \eqref{eq:dev_tav_ordre1}, we have that \eqref{eq:firstdev} holds if the operator $(f''(\ts) \otimes \Id + \Id \otimes f''(\ts) - \gamma f''(\ts) \otimes
f''(\ts) )$ is invertible.
To show this result, we just claim that it is a
symmetric definite positive operator. Indeed, since $\gamma <2L^{-1}$, by \Cref{hyp:strong_convex},
$\Id-(\gamma/2) f''(\ts)$ is symmetric positive definite and is diagonalizable
with the same orthogonal vectors $(\bff_i)_{i \in \zerod}$ as
$f''(\ts)$. If we denote by $(\lambda_i)_{i \in \zerod}$, then we get
that $(f''(\ts) \otimes \Id + \Id\otimes f''(\ts) - \gamma f''(\ts) \otimes
f''(\ts) )  = f''(\ts) \otimes (\Id-\gamma/2f''(\ts)) +  (\Id-\gamma/2f''(\ts))\otimes f''(\ts)$ is also diagonalizable in the orthogonal basis of $\rset^d \otimes \rset^d$, $(\bff_i \otimes \bff_j)_{i,j \in \zerod}$  and  $(\lambda_i(1-\gamma\lambda_j)+\lambda_j(1-\gamma \lambda_i))_{i,j \in \zerod}$ are its eigenvalues.
\end{proof}
\begin{lemma}\label{Lem:auxiliary}
  Assume \Cref{hyp:strong_convex}-\Cref{hyp:regularity}-\Cref{ass:def_noise}-\Cref{ass:lip_noisy_gradient_AS}($6\vee[2(k_{\epsilon}+1)]$)-\Cref{assum:reguarity_noise}. It holds as $\gamma \to 0$, 
\begin{align*}
  \intrd \mcC(\theta) \pigrmd &= \mcC(\ts) + O(\gamma) \eqsp,\\
  \intrd \mcC(\theta) \otimes \{\theta - \ts\} \pigrmd &= \mcC(\ts)\{\tav - \ts\} + O(\gamma)
\end{align*}
where $\mcC$ is given by \eqref{eq:def_cov_matrix}.
\end{lemma}
\begin{proof}
  By a second order Taylor expansion around $\ts$ of $\mcC$ and
  using \Cref{assum:reguarity_noise}, we get for all $x \in \rset^d$ that 
  \begin{equation*}
    \mcC(x) -\mcC(\ts) = \mcC'(\ts)\defEns{x-\ts} + \mcr_1(x) \eqsp,
  \end{equation*}
where $\mcr_1 : \rset^{d} \to \rset^{d}$ satisfies $\sup_{x \in \rset^d}\norm{\mcr_1(x)}/(\norm{x-\ts}^2+\norm{x+\ts}^{k_\epsilon+2}) < \plusinfty$. 
Taking the integral with respect to $\pg$ and using \Cref{preli:theo:statio_general}-\Cref{lem:moment_p} concludes the proof.
\end{proof}

\begin{proof}[Proof of \Cref{theo:statio_general}]
   Let $\gamma \in \ooint{0,2/L}$  and $(\tharg{k})_{k \in \nset}$ given
  by \eqref{eq:def_grad_sto} with $\tharg{0}$ distributed according to
  $\pg$ independent of $(\epsilon_k)_{k \in \nsets}$. 
For conciseness, in the rest of the proof, we skip the explicit dependence in $ \gamma $ in $ \te{i}{\gamma} $: we only denote it $ \theta_i $.

  The proof consists in showing that the residual term in \eqref{eq:firstdev} of
   \Cref{preli:theo:statio_general} is of order $O(\gamma^{2})$ and not only $O(\gamma^{3/2})$.
Note that we have already prove that $\tav - \ts = O(\gamma)$. To find the next term in the development, we develop further each of the terms.  By a fourth order Taylor expansion with integral remainder of $f'$ around $\ts$, and using \Cref{hyp:regularity}, we have 
\begin{align}
\teD{1} -\ts   =&   \teD{0} - \ts - \gamma \big[ f''(\ts) ( \teD{0} - \ts)  + (1/2) f^{(3)} (\ts) ( \teD{0} - \ts)^{\otimes 2} \nonumber\\
&+ (1/6) f^{(4)} (\ts) ( \teD{0} - \ts)^{\otimes 3} + \varepsilon_1(\theta_0) + \mcr_3(\theta) \big]  \label{eq:devordre4_sgd} \eqsp,
\end{align}
where $\mcr_3: \rset^d \to \rset^d$ satisfies $\sup_{x \in \rset^d} \norm{\mcr_3(x)}/\norm{x-\ts}^4 < \plusinfty$. 
Therefore taking the expectation and using \Cref{ass:def_noise}-\Cref{lem:moment_p} we get  
\begin{multline}
\label{devprincipal}
  f''(\ts) (\bar \theta_\gamma- \ts)  =  -  (1/2) f^{(3)}(\ts)  \intrdargD{ ( \theta- \ts)^{\otimes 2}} \\- (1/6) f^{(4)} (\ts)\intrdargD{( \theta - \ts)^{\otimes 3} } +  O(\gamma^{2}) \eqsp . 
\end{multline}
Since $f''(\ts)$ is invertible by \Cref{hyp:strong_convex},  To get the next term in the development, we show that
\begin{enumerate}[label=(\alph*)]
\item \label{item:proof_exp_a}  $ \intrdargD{ ( \theta- \ts)^{\otimes 3}} = \blacksquare \gamma^{2} + o(\gamma^2)$.
\item  \label{item:proof_exp_b} $\intrdargD{ ( \theta- \ts)^{\otimes 2}} = \square \gamma  +\triangle \gamma^{2 }+ o(\gamma^{2})$, for $\square$ given in \eqref{theo:statio_general_eq_2}, proving \eqref{theo:statio_general_eq_2}.
\end{enumerate}

 \ref{item:proof_exp_a}
Denote for $i=0,1$, $\eta_i = \theta_i-\ts$. By \eqref{eq:theo:statio_general_1_0}-\eqref{eq:theo:statio_general_1}, \Cref{lem:moment_p} and \Cref{ass:def_noise}-\Cref{ass:lip_noisy_gradient_AS}(12), we get 
\begin{align*}
  &\PE[\eta_1^{\otimes 3}]  =  \expe{ \left\{  (\Id-  \gamma f''(\ts) ) \eta_0  - \gamma \varepsilon_1(\teD{0}) -\gamma f'''(\ts) \eta_0^{\otimes 2}+ \mcr_1(\theta_0)  \right\} ^{\otimes 3}}\\ 
  &=    \PE \left[\{(\Id-  \gamma f''(\ts) )\eta_0\}^{\otimes 3} + \gamma^2 \{\epsilon_1(\teD{0})\}^{\otimes 2} \otimes \{(\Id-  \gamma f''(\ts) )\eta_0\}\right. \\
  &   +\gamma \{(\Id-  \gamma f''(\ts) )\eta_0\}^{\otimes 2} \otimes \{f'''(\ts) \eta_0^{\otimes 2}\}  \\
  & \qquad \left.+ \gamma \{f'''(\ts) \eta_0^{\otimes 2}\}  \otimes \{(\Id-  \gamma f''(\ts) )\eta_0\}^{\otimes 2} \right] + O(\gamma^3) \\
  &=    \PE \left[\{(\Id-  \gamma f''(\ts) )\eta_0\}^{\otimes 3} + \gamma^2 \{\epsilon_1(\teD{0})\}^{\otimes 2} \otimes \{(\Id-  \gamma f''(\ts) )\eta_0\} \right] + O(\gamma^3) \\
&=\PE \left[\{\eta_0\}^{\otimes 3}\right]+   \PE \left[\gamma \bfB\{\eta_0\}^{\otimes 3} + \gamma^2 \{\epsilon_1(\teD{0})\}^{\otimes 2} \otimes \{(\Id-  \gamma f''(\ts) )\eta_0\} \right] + O(\gamma^3) \eqsp, 
\end{align*}
where $\bfB \in \mathrm{L}(\rset^{d^3} , \rset^{d^3})$ is defined by 
\begin{equation*}
  \bfB = f''(\ts) \otimes \Id \otimes \Id + \Id\otimes f''(\ts)  \otimes \Id +\ \Id \otimes \Id \otimes f''(\ts) \eqsp.
\end{equation*}
Using \Cref{hyp:strong_convex} and the same reasoning as to show that $\bfA$ in \eqref{eq:def_bfA}, is well defined, we get that $\bfB$ is invertible.   Then since  $\eta_0$ and $\eta_1$ has the same distribution  $\pg$, we get 
\begin{multline*}
  \int_{\rset^d} (\theta-\ts)^{\otimes 3} \pg (\rmd \theta) \\= \gamma \bfB^{-1}\parentheseDeux{ \intrdargD{  \{\mcC(\theta)\} \otimes \{(\Id-  \gamma f''(\ts) )(\theta-\ts)\}}} + O(\gamma^2) \eqsp.
\end{multline*}
By \Cref{Lem:auxiliary}, we get 
\begin{equation*}
  \int_{\rset^d} (\theta-\ts)^{\otimes 3} \pg (\rmd \theta) = \gamma \bfB^{-1}\parentheseDeux{   \{\mcC(\ts)\} \otimes \{(\Id-  \gamma f''(\ts) )(\tav-\ts)\}} + O(\gamma^2) \eqsp.
\end{equation*}
Combining this result and \eqref{eq:firstdev} implies \ref{item:proof_exp_a}.

\ref{item:proof_exp_b} First, we have using \eqref{eq:devordre4_sgd}, \Cref{ass:def_noise} and \Cref{lem:moment_p}  that:
\begin{align*}
&\PE [ (\theta_1 -\ts) ^{\otimes  2}]   
  =  \mbe \left[ (\theta_0 -\ts) ^{\otimes  2} - \gamma (\Id\otimes f''(\ts)+ f''(\ts) \otimes \Id )  ( \theta- \ts)^{\otimes 2}  \right. \\
& \qquad   + (\gamma/2) (\theta_0 -\ts) \otimes \{f^{(3)} (\ts) ( \theta_0- \ts)^{\otimes 2}\}   \\
& \qquad  \left. + (\gamma/2)  \{f^{(3)} (\ts) ( \theta_0- \ts)^{\otimes 2} \}\otimes (\theta_0 -\ts) +  \gamma^{2}  \varepsilon_1(\teD{0})^{\otimes 2}(\teD{0}) \right] + O(\gamma^3) \eqsp.
 \end{align*}
 Since $\theta_0$ and $\theta_1$ follow the same distribution $\pg$, it follows that
 \begin{equation}
     \label{momentdordre2}
\begin{aligned}
& \gamma (\Id\otimes f''(\ts)+ f''(\ts) \otimes \Id ) \parentheseDeux{  \intrdargD{  ( \theta- \ts)^{\otimes 2}} } \\
& \qquad = O(\gamma^3)  + \int_{\rset^d} \left[  (\gamma/2)  (\theta -\ts) \otimes\{ f^{(3)} (\ts) ( \theta- \ts)^{\otimes 2}\}  \right.\\
& \qquad  + (\gamma/2) \{f^{(3)} (\ts) ( \theta- \ts)^{\otimes 2} \}\otimes (\theta -\ts)  \left.  + \gamma^{2}  \varepsilon_1(\teD{0})^{\otimes 2}(\theta_0) \right] \pg(\rmd \theta) \eqsp.
\end{aligned}
\end{equation}
Then by linearity of $f'''(\ts)$ and using \ref{item:proof_exp_a} we get  \ref{item:proof_exp_b}.

Finally the proof of \eqref{theo:statio_general_eq_1} follows from
combining the results of \ref{item:proof_exp_a}-\ref{item:proof_exp_b}
in \eqref{devprincipal}.

\end{proof}

\subsection{Proof of \Cref{th:tcl_non_asympt}}
\label{app:convergencetheorem}

\Cref{th:tcl_non_asympt} follows from the following more general result taking $\varphi : \theta \mapsto \theta - \ts$. 
\begin{theorem}
  \label{th:convergence_MC_genfunction}
Let $\varphi: \rset^{d} \to \rset^{q}$ be a Lipschitz function. Assume
\Cref{hyp:strong_convex}-\Cref{hyp:regularity}-\Cref{ass:def_noise}-\Cref{ass:lip_noisy_gradient_AS}(4)
and let $\gamma \in \ooint{0, 1/(2L)}$.  Then setting
$\rho =  (1 - 2\mu  \gamma (1-  \gamma L))^{1/2}$, for any starting point
$\theta_0 \in \rset^d$, $k \in \nsets$
\begin{equation*}
  \expe{ k^{-1} \sum_{i=0}^{k-1} \varphi(\te{i}{\gamma}) }  =  \pg (\varphi) + (1/k) \psi_{\gamma}(\theta_0)  + O(k^{-2})\eqsp,
\end{equation*}
and if $\pg (\varphi)=0$,
 \begin{align*}
\expe{\defEns{ k^{-1} \sum_{i=0}^{k-1} \varphi(\te{i}{\gamma})}^{\otimes 2}} &= \frac{1}{k } \pi_\gamma  \left[  \psi_{\gamma}^{\otimes 2}-(\psi_{\gamma} -\varphi) ^{\otimes 2} \right]
  \\ 
&\hspace{-2em}  -\frac{1}{k^2}   \left[     \pi_\gamma(\varpi_\gamma \varphi^{\top}+\varphi \varpi_\gamma^{\top} )+ \chi^{2}_{\gamma}(\theta_0)-\chi^{1}_{\gamma}(\theta_0)\right] +O(k^{-3}) \eqsp,
\end{align*}
where $\psi_{\gamma}$, $ \varpi_\gamma $, $\chi^{1}_{\gamma}$, $\chi^{2}_{\gamma}$ are solutions of the Poisson equation
\eqref{eq:def_poisson_sol_lip} associated with $\varphi$,  $ \psi_\gamma $, $\psi_{\gamma}^{\otimes 2}$ and $(\psi_{\gamma} -\varphi) ^{\otimes 2}$ respectively. 
\end{theorem}
 
\begin{proof}
  In the proof $C$  will denote generic constants which can change from line to line.
In addition, we skip the dependence on $\gamma$ for $\te{k}{\gamma}$,  simply denoted  $\theta_k$. 
\renewcommand{\te}[2]{\theta_{#1}}
Let $\theta_0 \in \rset^d$.
By \Cref{lem:poisson_exist}, $ \psi_\gamma $ exists and is Lipshitz,
and using \Cref{prop:existence_pi}-\ref{prop:existence_pi_item_2}, $\pg(\psi_{\gamma}) = 0$,
we have that $ \Rg^{k} \psi_{\gamma}(\theta_0) =O(\rho^{k})$, with $\rho:=  (1 - 2\mu  \gamma (1-  \gamma L))^{1/2}$. Therefore, setting
$ \Phi_k ={ k^{-1} \sum_{i=0}^{k-1} \varphi(\te{i}{\gamma})} $,
\begin{align*}
\E [\Phi_k] & =   k^{-1} \sum_{i=0}^{k-1} \expe{ \varphi(\te{i}{\gamma})}=   k^{-1} \sum_{i=0}^{k-1} \Rg^{i}\varphi(\theta_0) \\
&= \pg (\varphi) + k^{-1} \sum_{i=0}^{k-1} (\Rg^{i}\varphi(\theta_0) - \pg(\varphi) ))\\
&= \pg (\varphi) + k^{-1} \psi_{\gamma}(\theta_0) -    \Rg^{k} \psi_{\gamma}(\theta_0)=\pg (\varphi) + k^{-1}  \psi_{\gamma}(\theta_0) + O(\rho^{k}) \eqsp,
\end{align*}
We now consider the Poisson solution associated with
$\varphi \varphi^{\top}$, $\chi^3_{\gamma}$. By
\Cref{lem:poisson_exist_2}, such a function exists and satisfies
$\pg(\chi^3_{\gamma}) = 0$,
$R_{\gamma}^k \chi^3_\gamma (\theta_0) = O(\rho^k)$. Therefore, we
obtain using in addition the Markov property:
\begin{align*}
&\E [ \Phi_k \Phi_k^\top ] =  \frac{1}{k^2} \sum_{i,j=0}^{k-1} \expe{\varphi(\te{i}{\gamma})\varphi(\te{j}{\gamma})^\top }\\
 &=   \frac{1}{k^2} \sum_{i=0}^{k-1}
\bigg(
\expe{\varphi(\te{i}{\gamma}) \varphi(\te{i}{\gamma})^\top}
+  \sum_{j =i+1}^{k-1} \defEns{ \expe{\varphi(\te{i}{\gamma}) \varphi(\te{j}{\gamma})^\top }+ \expe{ \varphi(\te{j}{\gamma}) \varphi(\te{i}{\gamma})^\top}}
\bigg) \\
&= -\frac{1}{k^2} \sum_{i=0}^{k-1}  
  \Rg^{i} (\varphi \varphi^\top) (\theta_0) \\
  & \qquad + \frac{1}{k^2 }\sum_{i=0}^{k-1}
\bigg(
\sum_{j =i+1}^{k-1} \defEns{ \expe{\varphi(\te{i}{\gamma}) \varphi(\te{j}{\gamma})^\top }+ \expe{ \varphi(\te{j}{\gamma}) \varphi(\te{i}{\gamma})^\top}}
\bigg) \\
&= -\frac{1}{k } \pg ( \varphi \varphi^\top)   -\frac{1}{k^2}
\sum_{i=0}^{\infty} \defEns{\Rg^{i}(\varphi \varphi^\top)(\theta_0) - \pg(\varphi \varphi^\top) }+O(\rho^{k})   \\ 
  & \qquad + \frac{1}{k^2 }\sum_{i=0}^{k-1}
\bigg(
\sum_{j =i+1}^{k-1} \defEns{ \expe{\varphi(\te{i}{\gamma}) \varphi(\te{j}{\gamma})^\top }+ \expe{ \varphi(\te{j}{\gamma}) \varphi(\te{i}{\gamma})^\top}}
\bigg) \\
&= -\frac{1}{k } \pg ( \varphi \varphi^\top)   
-\frac{1}{k^2}
\chi^{3}_{\gamma}(\theta_0)+O(\rho^{k})   \\ 
& 
+ \frac{1}{k^2 }\sum_{i=0}^{k-1}
\bigg(
 \sum_{j =0}^{k-1-i} \defEns{ \expe{\varphi(\te{i}{\gamma}) (R_{\gamma}^j\varphi(\te{i}{\gamma}))^\top }+ \expe{ R_{\gamma}^j\varphi(\te{i}{\gamma}) \varphi(\te{i}{\gamma})^\top}}
\bigg)  \eqsp. 
\end{align*}
Thus using that for all $N \in \nset$ and $\theta \in \rset^d$, $\sum_{j=0}^{N} R_{\gamma}^j \varphi(\theta) = \sum_{j=0}^{N} \{R_{\gamma}^j \psi_{\gamma}(\theta)-R_{\gamma}^{j+1} \psi_{\gamma}(\theta)  \} = \psi_{\gamma}(\theta) - R_{\gamma}^{N+1} \psi_{\gamma}(\theta)$, we get 
\begin{align}
  \nonumber
\E [ \Phi_k \Phi_k^\top]&=-\frac{1}{k } \pg ( \varphi \varphi^\top)   
-\frac{1}{k^2}
                          \chi^{3}_{\gamma}  (\theta_0)\\
  \nonumber
  &+ \frac{1}{k^2 }\sum_{i=0}^{k-1}
\defEns{  \Rg^{i} \left[ \varphi \psi_{\gamma}^{\top}  -  \varphi(\Rg^{k-i} \psi_{\gamma}) ^\top\right](\theta_0)  }\\ 
&
  \label{eq:proof_tcl_quanti}
+ \frac{1}{k^2 }\sum_{i=0}^{k-1}
\defEns{\Rg^{i} \left[  \psi_{\gamma} \varphi ^{\top}  -  \Rg^{k-i} \psi_{\gamma} \varphi^\top\right](\theta_0) }  +O(\rho^{k})\eqsp .
\end{align}
Moreover, since $\varphi$ is
Lipschitz and $\Rg^N\psi_{\gamma}$ is $C \rho^N$-Lipschitz and we have 
$\sup_{x \in \rset^d} \{\Rg^N \psi_{\gamma}(x)/\norm{x}\} \leq C \rho^N$
by \Cref{lem:poisson_exist}, we get for all $x,y \in \rset^d$ and $N \in \nset$,
\begin{equation}
  \label{eq:lipproof_tcl}
\norm{\varphi (\Rg^N \psi_{\gamma})^{\top}(x) -   \varphi (\Rg^N \psi_{\gamma})^{\top}(y)} \leq C \rho^N\norm{x-y}(1+\norm{x} + \norm{y}) \eqsp.
\end{equation}
Then, we obtain by \Cref{lem:poisson_exist_2}
\begin{align}
  \nonumber
  \frac{1}{k}\sum_{i=0}^{k-1} \Rg^i[\varphi (\Rg^{k-i} \psi_{\gamma})^{\top}](\theta_0) &=   \frac{1}{k}\sum_{i=0}^{k-1} [\Rg^i-\pg][\varphi (\Rg^{k-i} \psi_{\gamma})^{\top}](\theta_0) \\
    \nonumber
                                                                                        & \qquad   \frac{1}{k}\sum_{i=0}^{k-1} \pg[\varphi (\Rg^{k-1} \psi_{\gamma})^{\top}](\theta_0) \\
     \label{eq:proof_tcl_quanti_2}
  & = (C/k)(1+\norm{\theta_0}) \sum_{i=0}^{k-1} \rho^k + \pg(\varphi \varpi^{\top}_{\gamma})/k + O(k^{-2}) \eqsp,
\end{align}
using $\pi_{\gamma}(\psi_{\gamma}) = 0$, $\sum_{i=0}^{\plusinfty}R_{\gamma}^i \psi_{\gamma}(\theta) = \varpi_{\gamma}(\theta)$, for all $\theta \in \rset^d$, where $\varpi_{\gamma}$ is the Poisson solution associated with $\psi_{\gamma}$.
Similarly, we have
\begin{equation}
       \label{eq:proof_tcl_quanti_3}
  \begin{aligned}
    \frac{1}{k}\sum_{i=0}^{k-1} \Rg^i[\Rg^{k-i} \psi_{\gamma} \varphi ^{\top}](\theta_0)& = \pg( \varpi_{\gamma}\varphi^{\top})/k + O(k^{-2})\\
    \frac{1}{k}\sum_{i=0}^{k-1} \defEns{\Rg^i [\varphi \psig^{\top}](\theta_0)-\pg[\varphi \psig^{\top}]} &= \chi^{4}_\gamma(\theta_0) + O(k^{-2}) \\
     \frac{1}{k}\sum_{i=0}^{k-1} \defEns{ \Rg^i [ \psig \varphi^{\top}](\theta_0)-\pg[ \psig \varphi^{\top}]} &= \chi^{5}_\gamma(\theta_0) + O(k^{-2})
     \eqsp,
  \end{aligned}
\end{equation}
where $\chi^{4}_\gamma$ and $\chi^{5}_\gamma$ are the Poisson solution associated with $\varphi \psig^{\top}$ and $\psig \varphi^{\top}$ respectively. 
Combining \eqref{eq:proof_tcl_quanti_2}-\eqref{eq:proof_tcl_quanti_3} in \eqref{eq:proof_tcl_quanti}, we obtain
\begin{align}
  \nonumber
&\E [ \Phi_k \Phi_k^\top]=\frac{1}{k } [\pg(\varphi \psig^{\top})+ \pg( \psig \varphi^{\top}) - \pg ( \varphi \varphi^\top)]     +O(k^{-3})
\\
   \label{eq:proof_tcl_quanti_4}
   & \qquad - \frac{1}{k^2 }[\pg(\varphi \varpi^{\top}_{\gamma})+ \pg( \varpi_{\gamma}\varphi^{\top}) +   \chi^{3}_{\gamma}  (\theta_0)- \chi^{4}_{\gamma}  (\theta_0) - \chi^{5}_{\gamma}  (\theta_0)]  \eqsp.
\end{align}
First note that 
\begin{equation}
     \label{eq:proof_tcl_quanti_5}
-\varphi \varphi^\top  
+   \varphi \psi_{\gamma}^\top  + \psi_{\gamma}\varphi^\top   =  -(\varphi-\psi_{\gamma}) (\varphi-\psi_{\gamma})^\top  
+   \psi_{\gamma} \psi_{\gamma}^\top \eqsp. 
\end{equation}
In addition, by \Cref{lem:poisson_exist_2} and definition, we have for all $\theta_0$ 
\begin{align*}
  &\chi^{3}_{\gamma}  (\theta_0)- \chi^{4}_{\gamma}  (\theta_0) - \chi^{5}_{\gamma}  (\theta_0) \\
&  =  \sum_{i=1}^{\plusinfty} \defEns{R^i_{\gamma}[\varphi \varphi^\top  
-   \varphi \psi_{\gamma}^\top  - \psi_{\gamma}\varphi^\top](\theta_0) -\pg[\varphi \varphi^\top  
  -   \varphi \psi_{\gamma}^\top  - \psi_{\gamma}\varphi^\top]}\\
  & = \sum_{i=1}^{\plusinfty} \defEns{R^i_{\gamma}[(\varphi-\psi_{\gamma}) (\varphi-\psi_{\gamma})^\top  
-   \psi_{\gamma} \psi_{\gamma}^\top ](\theta_0) -\pg[(\varphi-\psi_{\gamma}) (\varphi-\psi_{\gamma})^\top  
    -   \psi_{\gamma} \psi_{\gamma}^\top ]}\\
  & = \chi^{2}(\theta_0) - \chi^{1}(\theta_0) \eqsp.
\end{align*}
 Combining this result and \eqref{eq:proof_tcl_quanti_5} in \eqref{eq:proof_tcl_quanti_4} concludes the proof.

\end{proof}

\subsection{Proof of~\Cref{cor:quadconv}}
\label{sec:proof-crefc}
\renewcommand{\te}[2]{\theta_{#1}^{(#2)}}
In this section we apply  \Cref{th:tcl_non_asympt} to the case of a quadratic function, more specifically to the LMS algorithm described in~\Cref{ex:iid_observation_learning}, to prove~\Cref{cor:quadconv}.
	Recall that the sequence of iterates can be written, 
	\begin{eqnarray*}
		\te{k}{\gamma} -\ts &=&(\Id- \gamma  \Sigma ) \left( \te{k-1}{\gamma} - \ts \right) + \gamma \varepsilon_k(\te{k-1}{\gamma}) \\
		\varepsilon_k(\te{k-1}{\gamma})  &=& (\Sigma-X_k  X_k^{\top}) (\te{k-1}{\gamma} -\ts ) - ( X_k^{\top}\ts-Y_k ) X_k  = \varrho_k(\te{k-1}{\gamma}) + \xi_k \eqsp.
	\end{eqnarray*}
	%

	First note that with the notations of the text, and with $ \gamma \le 1/\rborne^{2} $, operator $  (\Sigma \otimes \Id + \Id \otimes \Sigma - \gamma \bfT)  $ is a positive operator on the set of symmetric matrices, and is thus invertible.

	We consider the linear function $\varphi$ which is $\varphi(\theta) = \theta - \ts$, thus $  \Phi_k= \bte{k}{\gamma}-\ts$.  	First, by~\Cref{lem:statio_quadractic}, $\pg (\varphi)=0$. 
	We have the following equalities:
	\begin{align}
	\psi_{\gamma}(\theta) &= (\gamma \Sigma)^{-1}(\theta - \ts) \label{eq:psi_id}\\
		\varpi_{\gamma} (\theta) &=  (\gamma \Sigma)^{-2}(\theta - \ts) \nonumber\\
	\psi_{\gamma}(\theta)^{\otimes 2} &=(\gamma \Sigma)^{-1} \varphi (\theta)^{\otimes 2}
	(\gamma \Sigma)^{-1} \nonumber \\
	(\psi_{\gamma} -\varphi)(\theta) ^{\otimes 2} &=  \big[ \Id - ( \gamma \Sigma)^{-1}  \big] 
	\varphi (\theta)^{\otimes 2}
	\big[ \Id - ( \gamma \Sigma)^{-1}  \big], \nonumber\\ 
	\psi_{\gamma}(\theta)^{\otimes 2} -(\psi_{\gamma} -\varphi)(\theta) ^{\otimes 2} & =  - (\Id \otimes \Id - (\gamma \Sigma)^{-1} \otimes \Id - \Id \otimes (\gamma \Sigma)^{-1}  ) ( \varphi (\theta)^{\otimes 2}) \nonumber\\
	& = \gamma^{-1} (\Sigma^{-1} \otimes\Sigma^{-1}) \big[ \Sigma \otimes \Id + \Id \otimes  \Sigma  - \gamma \Sigma \otimes \Sigma  \big]( \varphi (\theta)^{\otimes 2}) \label{eq:psigamma} \eqsp.
	\end{align}  
	Indeed, for \eqref{eq:psi_id} (other equations are basic linear algebra),   starting from  any $\theta_0$: 
	\begin{eqnarray*}
		\psi_{\gamma}(\theta_0)& =& \sum_{i=0}^{\infty} \E (\te{i}{\gamma}) - \ts = \sum_{i=0}^{\infty} (\Id-\gamma \Sigma)^{i}  (\theta_0- \ts)= (\gamma \Sigma)^{-1} (\theta_0- \ts) \eqsp.
	\end{eqnarray*}

	Moreover, the expectation of $\varphi(\theta)^{\otimes 2}$ under the stationary distribution is known  according to~\Cref{lem:statio_quadractic}, 
	\begin{align}
	\intrd \varphi (\theta)^{\otimes 2} \pigrmd
	&=  \gamma \big[ \Sigma \otimes \Id + \Id \otimes  \Sigma - \gamma \Sigma \otimes \Sigma \big]^{-1}  \pi_\gamma(\calC)  \label{eq:exp_phi_g_2}  \\
	&= \gamma [\Sigma \otimes \Id + \Id\otimes \Sigma - \gamma T ] ^{-1}\calC (\ts)\eqsp. \label{eq:exp_phi_g_2bis}
	\end{align}
	Applying \Cref{th:tcl_non_asympt}, we get a bound on $\E \left( (\bte{k}{\gamma}-\ts) (\bte{k}{\gamma}-\ts)^{\top }\right)$, using the notation $ \tilde{=} $ to denote equality up to linearly decaying term  $ O(\rho ^{k}) $:
	\begin{align}
	\label{eq:th5casquad}\PE[\Phi_k\Phi_k^\top]& \tilde{=}    \frac{1}{k } \int_{\rset^{d}} \left[  \psi_{\gamma}(\theta)^{\otimes 2} -(\psi_{\gamma} -\varphi)(\theta) ^{\otimes 2} \right] \rmd\pg (\theta)
	\\ 
	&  +\frac{1}{k^2}  \left[  \chi^{1}_{\gamma}(\theta_0)- \chi^{2}_{\gamma}(\theta_0) \right] - \frac{1}{k^{2}} \int_{\rset^{d}} \left[  \varphi(\theta)\varpi_\gamma(\theta)^{\top}+\varpi_{\gamma}(\theta)\varphi(\theta)^{\top}\right ] \rmd\pg (\theta) \eqsp. \nonumber
 	\end{align}
	
	\paragraph{Term proportional to $1/k$} \ \\
	Thus using Equations~\eqref{eq:psigamma} and \eqref{eq:exp_phi_g_2}:
	\begin{align}
	\label{eq:term1surkquad} \frac{1}{k } \int_{\rset^{d}} \left[  \psi_{\gamma}(\theta)^{\otimes 2} -(\psi_{\gamma} -\varphi)(\theta) ^{\otimes 2} \right] \rmd\pg (\theta)& = k^{-1} (\Sigma^{-1} \otimes\Sigma^{-1}) \pi_{\gamma}(\calC)\\
	& = k^{-1} \Sigma^{-1}  \pi_{\gamma}(\calC)\Sigma^{-1} . \nonumber
	\end{align}
	For the term proportional to $1/k^2$, we first need to compute the function $ \chi^3 $, solution to   the Poisson equation associated with $\theta \mapsto \varphi(\theta)^{\otimes 2}$.
	\paragraph{ Function $\chi^3_\gamma$}\ \\
	Following the proof of \Cref{prop:quadwithT}, we have:
	\begin{align*}
	\expeMarkov{\theta}{(\te{k}{\gamma}-\ts)^{\otimes 2}} &= (\Id -\gamma \Sigma \otimes \Id -\gamma \Id \otimes \Sigma + \gamma^{2} T )\expeMarkov{\theta}{(\te{k-1}{\gamma}-\ts)^{\otimes 2}} +  \E[\xi_{k}^{\otimes 2}].
	\end{align*}
	Thus
	\begin{align*}
	\chi^{3}_\gamma(\theta) &: =\sum_{k=1}^{\infty} \expeMarkov{\theta}{(\te{k}{\gamma}-\ts)^{\otimes 2}} - \pg (\varphi(\theta)^{\otimes 2}) \\
	&= (\gamma \Sigma \otimes \Id +\gamma \Id \otimes \Sigma - \gamma^{2} T )^{-1} \left[  \expeMarkov{\theta}{(\te{0}{\gamma}-\ts)^{\otimes 2}} - \pg (\varphi^{\otimes 2}) ) \right] \\
	\nu_0\left( \chi^{3}_\gamma \right)&: =  (\gamma \Sigma \otimes \Id +\gamma \Id \otimes \Sigma - \gamma^{2} T )^{-1} \left[  (\theta_0-\ts)^{\otimes 2} - \pg (\varphi^{\otimes 2}) ) \right]  \eqsp.
	\end{align*}
	Formally, the simplification comes from the fact that we study an arithmetico-geometric recursion of the form $w_{k+1} = a w_k + b$, $a<1$, and study $\sum_{i=0 }^{\infty} w_k -w_\infty = (1-a)^{-1} (w_0 -w_\infty)$. (note that here we cannot apply the recursion  with $(\Sigma \otimes \Id + \Id \otimes \Sigma - \gamma \Sigma \otimes \Sigma)$ because then ``$b$'' would depend on $k$.)

	\paragraph{Term proportional to $1/k^{2}$}\ \\
	This term is the sum of the following three terms:
	\begin{eqnarray*}
		\chi^1_\gamma (\theta_0) &=&(\gamma \Sigma)^{-1} \chi^3_\gamma(\theta_0) (\gamma \Sigma)^{-1}\\
\chi^2_\gamma(\theta_0)&=& (\Id-(\gamma \Sigma)^{-1}) \chi^3_\gamma(\theta_0)  (\Id-(\gamma \Sigma)^{-1}) \\
		\pi_{\gamma}(\varphi \varpi_\gamma^{\top}+\varpi_\gamma \varphi^{\top}) &=& \gamma^{-2}(\Sigma^{-2}\otimes \Id+ \Id \otimes \Sigma^{-2}) \pi_\gamma(\varphi^{\otimes 2})\eqsp.
	\end{eqnarray*}
	using $\psi_{\gamma} = (\gamma \Sigma)^{-1} \varphi$, and $\Rg{} \psi_{\gamma} = \psi_{\gamma}- \varphi= - (\Id-(\gamma \Sigma)^{-1}) \varphi$. 
	Finally, \begin{eqnarray}
\chi^1_\gamma(\theta_0) + \chi^2_\gamma(\theta_0)  &=& \gamma^{-1} (\Sigma^{-1} \otimes \Sigma^{-1}) (\Sigma \otimes \Id + \Id \otimes \Sigma - \gamma \Sigma \otimes \Sigma) \left( \chi^{3}_\gamma (\theta_0) \right) \nonumber\\
	&=& \gamma^{-2}(\Sigma^{-1} \otimes \Sigma^{-1}) \Omega \left[  (\theta_0-\ts)^{\otimes 2}   - \pg (\varphi^{\otimes 2} ) \right] \eqsp. \label{eq:t1t2}
	\end{eqnarray}
	With $\Omega =  (\Sigma \otimes \Id + \Id \otimes \Sigma - \gamma \Sigma \otimes \Sigma)  (\Sigma \otimes \Id + \Id \otimes \Sigma - \gamma T)^{-1}$.
	
	\paragraph{Conclusion}\ \\
	Combining \eqref{eq:th5casquad}, \eqref{eq:term1surkquad}  and \eqref{eq:t1t2}, we conclude the proof of \Cref{cor:quadconv}.
	\begin{align*}
	\E \bar{\theta}_k - \ts & \tilde{=}   \frac{1}{k\gamma} \Sigma^{-1} ( \theta_0 - \ts)\\
	\expe{\left (\bte{k}{\gamma} -\ts\right )^{\otimes 2}} & \tilde{=}    \frac{1}{k} \Sigma^{-1} \pi_\gamma(\calC) \Sigma^{-1} + \frac{1}{k^2\gamma^2}  \Sigma^{-1}   \Omega  \left[\varphi(\theta_0)^{\otimes 2}   
	-   \pg (\varphi^{\otimes 2} ) \right] \Sigma^{-1} \\
	& -\frac{1}{k^{2}\gamma^{2}}(\Sigma^{-2}\otimes \Id+ \Id \otimes \Sigma^{-2}) \pi_\gamma(\varphi^{\otimes 2})\eqsp .
	\end{align*}
	
\renewcommand{\te}[2]{\theta_{#1}^{(#2)}}



\subsection{Proof of \Cref{THEO:BIAS1}}
\label{sec:proof-crefmc:th}
Before giving the proof of \Cref{THEO:BIAS1}, we need several results
regarding Poisson solutions associated with the gradient flow ODE \eqref{eq:def_generator}. 
\subsubsection{Regularity of the gradient flow and  estimates on Poisson solution}
\label{app:subsec:gradientflow}
Let $\lreg \in \nset^*$ and consider the following assumption.
\begin{assumption}[$\lreg$]
	\label{assum:bounded_derivative}
	$f \in C^\lreg(\rset^d)$ and there exists $M \geq 0$ such that for all $i \in \{2,\ldots,\lreg\}$, $\sup_{\theta \in \rset^d} \norm{f^{(i)}(\theta)} \leq \barL $. 
\end{assumption}

\begin{lemma} 
	\label{lem:flow_properties}
	Assume \Cref{hyp:strong_convex} and
	\Cref{assum:bounded_derivative}($\lreg+1$) for $\lreg \in \nset^*$.
	\begin{enumerate}[label=\alph*)]
		\item \label{item:lem:flow_properties_1} For all $t \geq 0$, $ \varphi_t \in \mrc^{\lreg}(\rset^d, \rset^d)$, where $(\varphi_t)_{t \in \rset_+}$ is the differential flow associated with \eqref{eq:gradientflow}.
		In addition for all $\theta \in \rset$, $t \mapsto  \varphi_t^{(\lreg)}(\theta)$ satisfies the following ordinary differential equation,  
		\begin{equation*}
		\frac{\rmd \varphi_s^{(\lreg)}(\theta)}{\rmd s}\Bigr|_{s=t} = D^{\lreg}\defEns{f'\circ \varphi_t}(\theta) \eqsp, \text{ for all } t \geq 0 \eqsp,
		\end{equation*}
		with $\varphi_0^{'}=\Id$ and $\varphi_0^{(\lreg)}=0$ for $\lreg \geq 2$.
		\item \label{item:lem:flow_properties_2} For all $t \geq 0$ and $\theta \in \rset^d$, $      \normLigne[2]{\varphi_t(\theta)-\ts} \leq \rme^{-2\mu t}\norm[2]{\theta-\ts}$ \eqsp.
		\item \label{item:lem:flow_properties_3} If $\lreg\geq 2$,  for all $t \geq 0$,
		\begin{equation*}
		 \varphi_t'(\ts) = \rme^{- f''(\ts) t}  \eqsp.
		\end{equation*}
		\item \label{item:lem:flow_properties_4} If $\lreg \geq 3$,  for all $t \geq 0$ and $i,j,l\in \{1,\ldots,d\}$,
		\begin{multline*}
                  \ps{\varphi_t{''}(\ts)\defEns{\bff_i\otimes \bff_j}}{\bff_l} 
                  \\
=
                  \begin{cases}
                    \frac{  \rme^{- \lambda_l t}- \rme^{- (\lambda_i+\lambda_j)t}}{\lambda_l - \lambda_i-\lambda_j}  f^{(3)}(\ts) \defEns{\bff_i \otimes \bff_j \otimes \bff_l } & \text{if $\lambda_l \not = \lambda_i + \lambda_j$} \\
                      -t  \rme^{-\lambda_l t} f^{(3)}(\ts) \defEns{\bff_i \otimes \bff_j \otimes \bff_l } & \text{otherwise} \eqsp,  
                  \end{cases}
		\end{multline*}
		where $\{\bff_1,\ldots,\bff_d\}$ and $\{\lambda_1,\ldots,\lambda_d\}$ are the eigenvectors and the eigenvalues of $ f{''}(\ts)$ respectively satisfying for all $i \in \{1,\ldots,d\}$, $ f{''}(\ts) \bff_i = \lambda_i \bff_i$.
	\end{enumerate}
\end{lemma}
\begin{proof}
	\begin{enumerate}[label=\alph*), wide=0pt, labelindent=\parindent]
		\item This is a fundamental result on the regularity of  flows of autonomous differential equations, see \eg~\cite[Theorem 4.1 Chapter V]{Har_1982}
		\item Let $\theta \in \rset^d$. Differentiate $\norm[2]{\varphi_t(\theta)}$ with respect to $t$ and using \Cref{hyp:strong_convex}, that $f$ is at least continuously differentiable and  Gr\"onwall's inequality concludes the proof.
		\item By \ref{item:lem:flow_properties_1} and since $\ts$ is an equilibrium point, for all $x \in
		\rset^d$,  $\xi^{x}_t(\ts)= \varphi_t'(\ts)\defEns{x}$ satisfies the following ordinary differential equation
		\begin{equation}
		\label{eq:stoch_flow_1}
		\dot{\xi^{x}_s}(\ts) = -  f{''}(\varphi_s(\ts))\xi^{x}_s(\ts) \rmd s = -  f{''}(\ts)\xi^{x}_s(\ts) \rmd s \eqsp.
		\end{equation}
		with $\xi^{x}_0(\ts) = x $. The proof then follows from uniqueness of the solution of \eqref{eq:stoch_flow_1}. 
		\item By \ref{item:lem:flow_properties_1}, for all $x_1,x_2 \in \rset^d$,  $\xi^{x_1,x_2}_t(\ts) =  \varphi_t{''}(\ts)\defEns{x_1\otimes x_2}$ satisfies the ordinary stochastic differential equation:
		\begin{multline*}
		\frac{\rmd \xi^{x_1,x_2}_s}{\rmd s}(\ts) = -f^{(3)}(\varphi_s(\ts)) \defEns{\varphi_s{'}(\ts)x_1 \otimes \varphi_s{'}(\ts)x_2 \otimes \bfe_i } \\- 
		f{''}(\varphi_s(\ts))\defEns{\xi^{x_1,x_2}_s \otimes  \bfe_i}  \eqsp.
		\end{multline*}
		By \ref{item:lem:flow_properties_3} and since $\ts$ is an equilibrium point we get that $\xi^{x_1,x_2}_t(\ts)$ satisfies 
		\begin{equation*}
		\frac{\rmd \xi^{x_1,x_2}_s}{\rmd s}(\ts) = -f^{(3)}(\ts) \defEns{\rme^{-f{''}(\ts) t}x_1 \otimes \rme^{-f{''}(\ts) t} x_2 \otimes \bfe_i } - 
		f{''}(\ts)\defEns{\xi^{x_1,x_2}_s\otimes  \bfe_i} \eqsp.
		\end{equation*}
		Therefore we get for all $i,j,l \in \{1,\ldots,d\}$,
		\begin{equation*}
		\frac{ \rmd  \ps{\xi^{\bff_i,\bff_j}_s}{\bff_l}}{\rmd s} = 
		-f^{(3)}(\ts) \defEns{\rme^{-\lambda_i t}\bff_i \otimes \rme^{-\lambda_j t}\bff_j \otimes \bff_l } - 
		\lambda_l \ps{\xi^{\bff_i,\bff_j}_s}{ \bff_l} \eqsp.
		\end{equation*}
		This ordinary differential equation can be solved analytically which finishes the proof.
	\end{enumerate}
\end{proof}
Under \Cref{hyp:strong_convex} and \Cref{assum:bounded_derivative}$(\lreg)$, for any function $g: \rset^{d} \to \rset^{q}$, locally
Lipschitz, denote by $h_g$ the solution of the continuous Poisson
equation defined for all $\theta \in \rset^d$ by
\begin{equation}
\label{eq:def_poisson_solution}
h_g(\theta)=\int_{0}^{\infty} (g(\varphi_s(\theta))- g(\ts)) dt \eqsp.
\end{equation}
Note that $h_g$ is well-defined by \Cref{lem:flow_properties}-\ref{item:lem:flow_properties_2} and since $g$ is assumed to be locally-Lipschitz.
In addition by  \eqref{eq:def_generator}, $h_g$ satisfies 
\begin{equation}
\label{eq:definition_Poisson_deux}
 {  \generator h_g(\theta) = g(\theta) - g(\ts) \eqsp. }
\end{equation}
Define $h_{\Id} : \rset^d \to \rset^d$ for all $x \in \rset^d$ by 
\begin{equation}
h_{\Id}(\theta)=\int_{0}^{\infty} \defEns{\varphi_s(\theta)- \ts} dt \eqsp.
\end{equation}
Note that $h_{\Id}$ is also well-defined by \Cref{lem:flow_properties}-\ref{item:lem:flow_properties_2}.

\begin{lemma}
	\label{lem:lip_g}
	Let
	$g: \rset^d \to \rset$ satisfying \Cref{assum:assum_f_test}($\lreg,\pmom$) for $\lreg,\pmom \in \nset$, $\lreg \geq 1$. 
	Assume  \Cref{hyp:strong_convex} and
	\Cref{assum:bounded_derivative}($\lreg+1$).
	\begin{enumerate}[label=\alph*)]
		\item  Then for all $\theta\in \rset^d$, 
		\begin{equation*}
		\abs{h_g}(\theta) \leq a_g \defEns{(b_g/\mu)\norm{\theta-\ts} + (\pmom \mu)^{-1}\norm[\pmom]{\theta-\ts} } \eqsp.
		\end{equation*}
		\item
		\label{item:third_derivative_poisson}
		If $\lreg \geq 2$, then  $\nabla h_{\Id}(\ts) = (f{''}(\ts))^{-1}$. If $\lreg \geq 3$, then for all $i,j \in \{1,\ldots,d\}$,
		\begin{multline*}
		\hspace{-1.1cm}
                                    \frac{\partial^2 h_{\Id}}{\partial \theta_i \partial \theta_j}(\ts)                  = \sum_{l = 1}^d \Big[ -f^{(3)}(\ts) \defEns{\parentheseDeux{\parenthese{f{''}(\ts) \otimes \Id + \Id \otimes f{''}(\ts) }^{-1}\defEns{\bfe_i \otimes \bfe_j }} \otimes \bfe_i} \\
                   \times (f{''}(\ts))^{-1} \bfe_l \Big]\eqsp.
		\end{multline*} 
	\end{enumerate}
\end{lemma}

\begin{proof}
	\begin{enumerate}[label=\alph*), wide=0pt, labelindent=\parindent]
        \item For all $\theta \in \rset^d$, we have using
         \Cref{lem:lip_g_0} and \eqref{eq:def_poisson_solution}
		\begin{align*}
		\abs{h_g(\theta)} \leq a_g \int_{0}^{\plusinfty} \norm{\varphi_s(\theta)-\ts}\defEns{b_g+ \norm[\pmom]{\varphi_s(\theta)-\ts}} \rmd s \eqsp.
		\end{align*}
		The proof then follows from \Cref{lem:flow_properties}-\ref{item:lem:flow_properties_2}.
		\item The proof is a direct consequence of
		\Cref{lem:flow_properties}-\ref{item:lem:flow_properties_3}-\ref{item:lem:flow_properties_4}
		and \eqref{eq:def_poisson_solution}.
	\end{enumerate}
\end{proof}

\begin{theorem}
  \label{theo:regularity_poisson}
  Let
	$g: \rset^d \to \rset$ satisfying \Cref{assum:assum_f_test}($\lreg,\pmom$) for $\lreg,\pmom \in \nset$, $\lreg \geq 2$.  
	Assume \Cref{hyp:strong_convex}-\Cref{assum:bounded_derivative}($\lreg+1$).
	\begin{enumerate}[label=\alph*)]
		\item \label{item:theo:regularity_poisson_1} For all $i \in \{1,\ldots,\lreg\}$, there exists $C_i \geq 0$  such that for all $\theta \in \rset^d$ and $t \geq 0$,
		\begin{equation*}
		\norm{\varphi_t^{(i)} (\theta)} \leq C_i\rme^{-\mu t} \eqsp. 
		\end{equation*}
		\item  \label{theo:regularity_poisson_b} Furthermore, $h_g \in \mrc^{\lreg}(\rset^d)$ and for all $i \in \{0,\ldots,\lreg \}$, there exists $C_i \geq 0$  such that for all $\theta \in \rset^d$,
		\begin{equation*}
		\norm{h_g^{(i)} (\theta)} \leq C_i \defEns{1+\norm[\pmom]{\theta-\ts}} \eqsp.
		\end{equation*}
	\end{enumerate}
\end{theorem}
\begin{proof}
	\begin{enumerate}[label=\alph*), wide=0pt, labelindent=\parindent]
		\item
		The proof is by induction on $\lreg$. 
		By \Cref{lem:flow_properties}-\ref{item:lem:flow_properties_1}, for all $x \in
		\rset^d$, and $\theta \in \rset^d$, $\xi^{x}_t(\theta)=D \varphi_t(\theta)\defEns{x}$ satisfies
		\begin{equation}
		\frac{\rmd \xi^{x}_s(\theta)}{\rmd s }\Bigr|_{s=t} = -  f{''}(\varphi_t(\theta))\xi^{x}_t(\theta)  \eqsp.
		\end{equation}
		with $\xi^{x}_0(\theta) = x $.  Now  differentiating $s \to \norm[2]{
			\xi^{x}_s(\theta)}$, using \Cref{hyp:strong_convex} and   Gr\"onwall's inequality, we get
		$\norm[2]{\xi^{x}_s(\theta)} \leq \rme^{-2mt}\norm[2]{x}$ which implies the result for $\lreg=2$. 
		
		Let now $\lreg >2$.  Using again \Cref{lem:flow_properties}-\ref{item:lem:flow_properties_1}, Fa\`a  di Bruno's formula
		\cite[Theorem 1]{Lev_2006}  and  since \eqref{eq:gradientflow} can be written on the form 
		\begin{equation*}
		\frac{\rmd \varphi_s(\theta)}{\rmd s }\Bigr|_{s=t} = - \sum_{j=1}^d   f{'}(\varphi_t(\theta)) \defEns{\bfe_j} \bfe_j \eqsp, 
		\end{equation*}
		for all $i \in \{2,\ldots,\lreg\}$, $\theta \in \rset^d$ and $x_1,\cdots,x_i \in \rset^d$,  the function $\xi^{x_1,\cdots,x_i}_t(\theta) = \varphi_t^{(i)}(\theta)\defEns{x_1\otimes \cdots \otimes x_i}$ satisfies the ordinary differential equation:
		\begin{multline}
		\label{eq:stoch_flow_2}
		\frac{\rmd \xi^{x_1,\cdots,x_i}_s(\theta)}{\rmd s}\Bigr|_{s=t} \\= -\sum_{j=1}^d  \sum_{ \Omega \in \mathsf{P}(\{1,\ldots,i \}) } f^{(|\Omega|+1)}(\varphi_t(\theta)) \defEns{\bfe_j \otimes \bigotimes_{l=1}^{i} \bigotimes_{j_1,\cdots,j_l \in \Omega} \xi^{x_{j_1},\cdots,x_{j_{l}}}_t(\theta)} \bfe_j  \eqsp,
		\end{multline}
		where $\mathsf{P}(\{1,\ldots,i \})$ is the set of partitions of
		$\{1,\ldots,i \}$, which does not contain the empty set and
		$|\Omega|$ is the cardinal of $\Omega \in \mathsf{P}(\{1,\ldots,i+1
		\})$.
		We now show  by induction on $i$  that
		for all $i \in \{1,\ldots,\lreg\}$,  there exists a
		universal constant $C_i$ such that for all $t \geq 0$ and $\theta \in \rset^d$,
		\begin{equation}
		\label{eq:stoch_flow_3}
		\sup_{x \in \rset^d} \norm{ \varphi_t^{(i)}(\theta)} \leq C_i  \rme^{-\mu t} \eqsp.
		\end{equation}
		For $i=1$, the result follows from the case $\lreg =1$. Assume that
		the result is true for $\{1,\ldots,i\}$ for $i \in
		\{1,\ldots,\lreg-1\}$. We show the result for $i+1$.  By
		\eqref{eq:stoch_flow_2}, we have for all  $\theta \in \rset^d$ and $x_1,\cdots,x_i \in \rset^d$, 
		\begin{multline*}
\frac{\rmd \norm[2]{\xi^{x_1,\cdots,x_{i+1}}_s(\theta)}}{\rmd s}\Bigr|_{s=t}    \\
=		- \sum_{ \Omega \in \mathsf{P}(\{1,\ldots,i+1 \}) } f^{(|\Omega|+1)}(\varphi_t(\theta)) \defEns{\xi^{x_1,\cdots,x_{i+1}}_t(\theta) \otimes  \bigotimes_{l=1}^{i+1} \bigotimes_{j_1,\ldots,j_l \in \Omega} \xi^{x_{j_1},\cdots,x_{j_{l}}}_t(\theta)}    \eqsp.
		\end{multline*}
		Isolating the term corresponding to $\Omega = \{ \{1,\ldots,i+1\} \}$ in
		the sum above and using  Young's
		inequality, \Cref{hyp:strong_convex}, Gr\"onwall's inequality and the
		induction hypothesis, we get that there exists a universal constant
		$C_{i+1}$ such that for all $t \geq 0$ and $x \in\rset^d$
		\eqref{eq:stoch_flow_3} holds for $i+1$. 
		\item 
		The proof is a consequence of \ref{item:theo:regularity_poisson_1}, \eqref{eq:def_poisson_solution}, \Cref{assum:assum_f_test}($\lreg,\pmom$) and Lebesgue's dominated convergence theorem.
	\end{enumerate}
\end{proof}


\subsubsection{Proof of Theorem~\ref{THEO:BIAS1}}   
We preface the proof of the Theorem by two fundamental first estimates. 
\begin{theorem}
	\label{theo:bias_0}
        Let $g: \rset^d \to \rset$ satisfying
        \Cref{assum:assum_f_test}($3,\pmom$) for $\pmom \in
        \nset$. Assume
        \Cref{hyp:strong_convex}-\Cref{hyp:regularity}-\Cref{ass:def_noise}-\Cref{assum:reguarity_noise}.
        Furthermore, suppose that there exists $q \in \nset$ and $C \geq 0$ such that for all $\theta \in \rset^d$,
        \begin{equation*}
	\label{theo:bias_0_eq_1}
          \expe{\norm{\epsilon_1(\theta)}^{\pmom+3}} \leq C(1+\norm[q]{\theta-\ts}) \eqsp,
        \end{equation*}
and        \Cref{ass:lip_noisy_gradient_AS}($2\tilde{p}$) holds for $\tilde{p} = 
        \pmom+3+q \vee k_{\varepsilon}$. Let $C_{\tildep}$
        be the numerical constant given by \Cref{lem:moment_p} associated with $\tilde{p}$.
        \begin{enumerate}[label=(\alph*)]
        \item 	\label{theo:bias_0_item_1}         For all $\gamma \in \ooint{0,1/(LC_{\ptilde})}$,
        $k \in \nset^*$, and starting point  $\theta_0 \in \rset^d$,
	\begin{multline*}
          { \expeMarkov{}{k^{-1} \sum_{i=1}^k
              \defEns{g(\tharg{i})-g(\ts)}}=
            \frac{h_g(\theta_0)-\expeMarkov{}{h_g(\tharg{k+1})}}{k
              \gamma} } \\ {+(\gamma/2) \int_{\rset^d}
            h_g''(\tildetheta) \expe{\defEns{
                \epsilon_1(\tildetheta)}^{\otimes 2}} \rmd \pi_\gamma
            (\tildetheta) - (\gamma/k) \tilde{A}_1(\theta_0,k) - \gamma^2
            \tilde{A}_2(\theta_0,k) \eqsp,}
	\end{multline*}
where $\tharg{k}$ is the Markov chain starting from $\theta_0$,  defined by the recursion \eqref{eq:def_grad_sto}, and
	\begin{align}
	\label{theo:bias_0_eq_1}
       \sup_{i \in \nsets}   \tilde{A}_1(\theta_0,i)  &\leq C \defEns{1+\norm[\ptilde]{\theta_0-\ts}} \eqsp, \, \\
          	\label{theo:bias_0_eq_2}
          \tilde{A}_2(\theta_0,k)&  \leq C \defEns{1+\norm[\ptilde]{\theta_0-\ts}/k} \eqsp,
	\end{align}
	for some constant $C \geq 0$ independent of $\gamma$ and $k$.
      \item \label{theo:bias_0_item_2} For all $\gamma \in \ooint{0,1/(LC_{\ptilde})}$,
	\begin{equation*}
	\abs{\int_{\rset^d} g(\tildetheta) \pi_{\gamma}(\rmd \tildetheta) -g(\ts) + (\gamma/2) \int_{\rset^d} h_g''(\tildetheta) \expe{\defEns{ \epsilon(\tildetheta)}^{\otimes 2}} \rmd \pi_\gamma (\tildetheta) } \leq C \gamma^2 \eqsp.
	\end{equation*}
        \end{enumerate}
      \end{theorem}

      \begin{proof}
        \begin{enumerate}[wide, labelwidth=!, labelindent=0pt,label=(\alph*)]
        \item 	Let $k \in \nset^*$, $\gamma >0$ and $\theta \in \rset^d$. Consider the sequence $(\tharg{k})_{k \geq 0}$ defined by the stochastic gradient recursion \eqref{eq:def_grad_sto} and starting at $\theta$. 
	\Cref{theo:regularity_poisson}-\ref{theo:regularity_poisson_b} shows that $h_g \in \mrc^3(\rset^d)$.
	Therefore using \eqref{eq:def_grad_sto} and the Taylor expansion
	formula, we have for all $i \in \{1,\ldots,k\}$
	\begin{multline*}
	h_g(\tharg{i+1}) =     h_g(\tharg{i}) + \gamma h_g{'}(\tharg{i}) \defEns{-f'(\tharg{i}) + \epsilon_{i+1}(\tharg{i})} \\+(\gamma^2/2) h_g''(\tharg{i}) \defEns{-f'(\tharg{i}) + \epsilon_{i+1}(\tharg{i})}^{\otimes 2}
	\\
	+ (\gamma^3/(3!)) h_g^{(3)}(\tharg{i}+ s^{(\gamma)}_i\Delta\tharg{i+1}  ) \defEns{-f'(\tharg{i}) + \epsilon_{i+1}(\tharg{i})}^{\otimes 3} \eqsp,
	\end{multline*}
	where $s^{(\gamma)}_i \in \ccint{0,1}$ and $\Delta\tharg{i+1} = \tharg{i+1}-\tharg{i}$.
	Therefore by \eqref{eq:definition_Poisson_deux}, we get 
	\begin{multline*}
	 {k^{-1} \sum_{i=1}^k \defEns{g(\tharg{i})-g(\ts)}= \frac{h_g(\theta)-h_g(\tharg{k+1})}{k \gamma} 
		+k^{-1} \sum_{i=1}^k h_g{'}(\tharg{i-1}) \epsilon_{i+1}(\tharg{i}) }\\
	 { +(\gamma/(2k)) \sum_{i=1}^k h_g''(\tharg{i}) \defEns{-f'(\tharg{i}) + \epsilon_{i+1}(\tharg{i})}^{\otimes 2} }\\
	 {+(\gamma^2/(3!k))  \sum_{i=1}^k h_g^{(3)}(\tharg{i}+ s^{(\gamma)}_i\Delta\tharg{i+1}  ) \defEns{-f'(\tharg{i}) + \epsilon_{i+1}(\tharg{i})}^{\otimes 3} \eqsp.}
	\end{multline*}
	Taking the expectation and using \Cref{ass:def_noise}, we have
	\begin{multline*}
	 {  \expeMarkov{}{k^{-1} \sum_{i=1}^k \defEns{g(\tharg{i})-g(\ts)}} = \frac{\expeMarkov{}{h_g(\theta)-h_g(\tharg{k+1})}}{k \gamma} } \\
	 { +(\gamma/2) \int_{\rset^d} h_g''(\tildetheta) \expe{\defEns{ \epsilon_1(\tildetheta)}^{\otimes 2}} \rmd \pi_\gamma (\tildetheta)-
		(\gamma/(2k))	\tilde{B}_1 + (\gamma^2/(3!k)) \tilde{B}_2 \eqsp,}
	\end{multline*}
	where
	\begin{align*}
          \tilde{B}_1(\theta_0,k)
&          =  
 \expeMarkov{}{\sum_{i=1}^k \parenthese{h_g''(\ts) \defEns{ \epsilon_{1}(\ts)}^{\otimes 2} -  h_g''(\tharg{i}) \defEns{-f'(\tharg{i}) + \epsilon_{i+1}(\tharg{i})}^{\otimes 2}}} \\
	\tilde{B}_2 (\theta_0,k) &=   \expeMarkov{}{ \sum_{i=1}^k h_g^{(3)}(\tharg{i}+ s^{(\gamma)}_i\Delta\tharg{i+1}  ) \defEns{-f'(\tharg{i}) + \epsilon_{i+1}(\tharg{i})}^{\otimes 3}} \eqsp.
	\end{align*}
        Then it remains to show that \eqref{theo:bias_0_eq_1} and
        \eqref{theo:bias_0_eq_2} holds. By
        \Cref{hyp:regularity}, \Cref{THEO:BIAS1}-\ref{theo:regularity_poisson_b}
        and \Cref{assum:reguarity_noise}, there exists $C \geq 0$ such
        that we have that for all $\theta \in \rset^d$,
        \begin{align*}
         \norm{H'(\theta)}& \leq  C_1(1+\norm[k_{\epsilon} + \pmom+2]{\theta - \ts})  \eqsp,
        \end{align*}
        where
        $H : \theta \mapsto h_g''(\theta)
        \expeLigne{\{-f'(\theta)+\varepsilon_1(\theta)\}^{\otimes
            2}}$.  Therefore \eqref{theo:bias_0_eq_1} follows from \Cref{ass:def_noise},
        \Cref{lem:lip_g_0} and
        \Cref{theo:convergence_loc_lip_wasserstein}. Finally by
        \Cref{theo:regularity_poisson}-\ref{theo:regularity_poisson_b} and Jensen inequality,
        there exists $C \geq 0$ such that for all
        $i \in \{1,\ldots,k\}$, almost surely,
        \begin{multline*}
          h_g^{(3)}(\tharg{i}+ s^{(\gamma)}_i\Delta\tharg{i+1}  ) \defEns{-f'(\tharg{i}) + \epsilon_{i+1}(\tharg{i})}^{\otimes 3}\\
 \leq C\parenthese{1+\norm[p_2]{\tharg{i}}
+ \norm[p_2]{\epsilon_{i+1}(\tharg{i})}}\parenthese{\norm[3]{f'(\tharg{i})} + \norm[3]{\epsilon_{i+1}(\tharg{i})}}\eqsp.
        \end{multline*}
        The proof of \eqref{theo:bias_0_eq_2} then follows from~\Cref{hyp:regularity}, \Cref{ass:def_noise}, \eqref{theo:bias_0_eq_1} and \Cref{lem:moment_p}. 
      \item 	This result is a direct consequence of \Cref{theo:convergence_loc_lip_wasserstein} and \ref{theo:bias_0_item_1}.
      \end{enumerate}
    \end{proof}

\begin{proof}[Proof of \Cref{THEO:BIAS1}]
  Under the stated assumptions, the functions 
  $\psi : \theta \mapsto h_g''(\theta) \expeLigne{\defEns{
      \epsilon(\theta)}^{\otimes 2}}$ and $g$ satisfy the conditions of
  \Cref{theo:bias_0}. The proof then follows from combining
  \Cref{theo:bias_0}-\ref{theo:bias_0_item_2} applied to $\psi$
  and \Cref{theo:bias_0} applied to $g$.
\end{proof}



\section{Acknowledgments}
The authors would like to thank \'Eric Moulines and Arnak Dalalyan for helpful discussions. We acknowledge support from the chaire Economie des nouvelles donnees with the data science joint research initiative with the fonds AXA pour la recherche, and the Initiative de Recherche ``Machine Learning for Large-Scale Insurance'' from the Institut Louis Bachelier.

\bibliographystyle{plain}
\bibliography{biblio_all}

\end{document}